\documentclass{article}

\usepackage[main,final,nonatbib]{neurips_2026}
\usepackage[numbers]{natbib}
\usepackage{algorithm}
\usepackage{algorithmic}
\usepackage[utf8]{inputenc} %
\usepackage[T1]{fontenc}    %
\usepackage{hyperref}       %
\usepackage{url}            %
\usepackage{booktabs}       %
\usepackage{amsfonts}       %
\usepackage{nicefrac}       %
\usepackage{microtype}      %
\usepackage{xcolor}         %

\usepackage{amsmath,amsfonts,bm,amsthm}

\newcommand{\piego}{\pi^\text{ego}}

\newcommand{\Pitrain}{\Pi^\text{train}}

\newcommand{\Pieval}{\Pi^\text{eval}}

\newcommand{\psieval}{\psi^\text{eval}}

\newcommand{\BR}{\text{BR}}
\newcommand{\pxp}{p_\text{XP}}
\newcommand{\psp}{p_\text{SP}}
\newcommand{\pmp}{p_\text{MP}}

\usepackage{thm-restate}
\usepackage{theoremref} %

\theoremstyle{plain}

\newtheorem{definition}{Definition}

\newtheorem{prop}{Proposition}

\def\eqref#1{equation~\ref{#1}}

\def\1{\bm{1}}

\DeclareMathAlphabet{\mathsfit}{\encodingdefault}{\sfdefault}{m}{sl}
\SetMathAlphabet{\mathsfit}{bold}{\encodingdefault}{\sfdefault}{bx}{n}

\newcommand{\E}{\mathbb{E}}

\newcommand{\R}{\mathbb{R}}

\DeclareMathOperator*{\argmax}{arg\,max}

\newcommand{\ours}{\textsc{ROTATE}}

\newcommand{\eqskips}[2]{\setlength{\abovedisplayskip}{#1}\setlength{\abovedisplayshortskip}{#1}\setlength{\belowdisplayskip}{#2}\setlength{\belowdisplayshortskip}{#2}}

\usepackage{algorithm}
\usepackage{graphicx}
\usepackage{array}
\usepackage{subcaption}
\usepackage{enumitem}
\usepackage{enumitem}
\usepackage{wrapfig}
\usepackage{siunitx}
\usepackage{setspace}
\usepackage{cleveref}
\usepackage{mathtools}
\crefname{algorithm}{Algorithm}{Algorithms}
\crefname{ALC@unique}{Line}{Lines}
\crefname{table}{Table}{Tables}
\crefname{line}{Line}{Lines}
\makeatletter
\def\theHALC@unique{\arabic{ALC@unique}}
\renewcommand{\theALC@unique}{\arabic{ALC@line}}
\makeatother

\definecolor{asparagus}{rgb}{0.53, 0.66, 0.42}
\definecolor{bittersweet}{rgb}{1.0, 0.44, 0.37}
\definecolor{ao(english)}{rgb}{0.0, 0.5, 0.0}
\definecolor{mydarkblue}{rgb}{0,0.08,0.45}
\definecolor{blue(ncs)}{rgb}{0.0, 0.53, 0.74}
\definecolor{celestialblue}{rgb}{0.29, 0.59, 0.82}
\definecolor{earthyellow}{rgb}{0.88, 0.66, 0.37}
\definecolor{lightgray}{rgb}{0.83, 0.83, 0.83}
\definecolor{brickred}{rgb}{0.8, 0.25, 0.33}
\definecolor{myblue}{rgb}{0.345, 0.545, 0.902}
\ifdefined\nohyperref\else\ifdefined\hypersetup
  \hypersetup{ %
    pdftitle={},
    pdfsubject={},
    pdfkeywords={},
    pdfborder=0 0 0,
    pdfpagemode=UseNone,
    colorlinks=true,
    linkcolor=mydarkblue,
    citecolor=ao(english),
    filecolor=mydarkblue,
    urlcolor=magenta,
    }
  \fi
\fi

\title{\textsc{ROTATE}: Regret-driven Open-ended Training for Ad Hoc Teamwork}

\author{%
  Caroline Wang\textsuperscript{1}\thanks{Equal contribution.\quad\textsuperscript{2}Appendix~\ref{sec:related_work} reviews prior work on AHT, teammate generation, and open-ended learning.} \quad
  Arrasy Rahman\textsuperscript{1}\footnotemark[1] \quad
  Benjamin Nativi \quad
  Johnny Liu\textsuperscript{1} \quad
  Jiaxun Cui\textsuperscript{1} \\
  \textbf{Yoonchang Sung\textsuperscript{2} \quad
  Peter Stone\textsuperscript{1,3}} \\
  \textsuperscript{1}The University of Texas at Austin \enspace
  \textsuperscript{2}Nanyang Technological University \enspace
  \textsuperscript{3}Sony AI \\
  \texttt{\{caroline.l.wang, jyliu, cuijiaxun\}@utexas.edu}, \texttt{yoonchang.sung@ntu.edu.sg}, \\
  \texttt{\{arrasy, pstone\}@cs.utexas.edu}, \texttt{benjamin.nativi@outlook.com}
}

\begin{document}

\maketitle

\begin{abstract}
    Learning to collaborate with previously unseen partners 
    is a fundamental generalization challenge, known as Ad Hoc Teamwork (AHT). 
    Existing methods often adopt a two-stage pipeline: first, a fixed population of \textit{teammates} is generated, and second, an \textit{AHT agent} is trained to collaborate with them. 
    As a consequence, training teammates are generated ignoring whether they are informative for the AHT agent to learn from. Similarly, AHT agents are typically trained under the assumption that the training teammate set is uncontrollable, though its composition strongly influences 
    generalization.
    This paper presents a unified framework for AHT by reformulating the problem as an open-ended learning process between an AHT agent and an adversarial teammate generator. 
    We introduce \ours{}, a regret-driven, open-ended training algorithm that alternates between improving the AHT agent and generating teammates that probe its collaboration deficiencies. 
    Experiments across Overcooked and Level-Based Foraging tasks demonstrate that \ours{} substantially outperforms baselines on unseen teammates, establishing a new standard for robust, generalizable teamwork.
\end{abstract}

\section{Introduction}
\label{sec:intro}
As AI agents are deployed in diverse applications, it is increasingly crucial that they can collaborate effectively with previously unseen AI agents and humans. While methods for training teams of agents have been explored in cooperative multi-agent reinforcement learning (CMARL)~\citep{foerster2018counterfactual, rashid2020monotonic}, prior work highlighted that CMARL agents fail to perform optimally when collaborating with unfamiliar teammates~\citep{vezhnevets2020options, rahman_towards_2021}. 
Dealing with previously unseen teammates requires adaptive AI agents that efficiently approximate the optimal strategy for collaborating with diverse teammates. The training of such adaptive agents has been explored within ad hoc teamwork (AHT)~\citep{bowling2005coordination, stone2010ad, mirsky2022survey} and zero-shot coordination (ZSC)~\citep{hu2020other,cui2021k, lupu2021trajectory}. 

Most work has decomposed AHT learning into two stages~\citep{mirsky2022survey}, consisting of first creating a fixed set of teammates, and then training an AHT agent using reinforcement learning (RL), based on interactions with teammates sampled from the set. 
Methods for AHT agent learning typically rely on human-designed or pretrained teammates~\citep{papoudakis2020liam, zintgraf2021deep, rahman_towards_2021} and therefore struggle to handle novel behaviors outside the predefined set of teammates~\citep{strouse_fcp_2022,carroll_utility_2019}.
Recent work enhances the generalization capabilities of AHT agent learning methods by substituting the predefined set of teammates with a generated collection of diverse teammates~\citep{lupu2021trajectory,  rahman2024minimum}, which are trained to maximize different notions of diversity. 
One such diversity notion is \textit{adversarial diversity}~\citep{rahman2023generating, charakorn2023generating}, which seeks to generate a set of teams that cooperate well with other team members, but not across different teams.%
\stepcounter{footnote}\textsuperscript{\thefootnote}

\begin{wrapfigure}[15]{r}{0.4\textwidth}
    \vspace{-5pt}
    \centering
    \includegraphics[width=\linewidth, trim=284pt 0 0 0, clip]{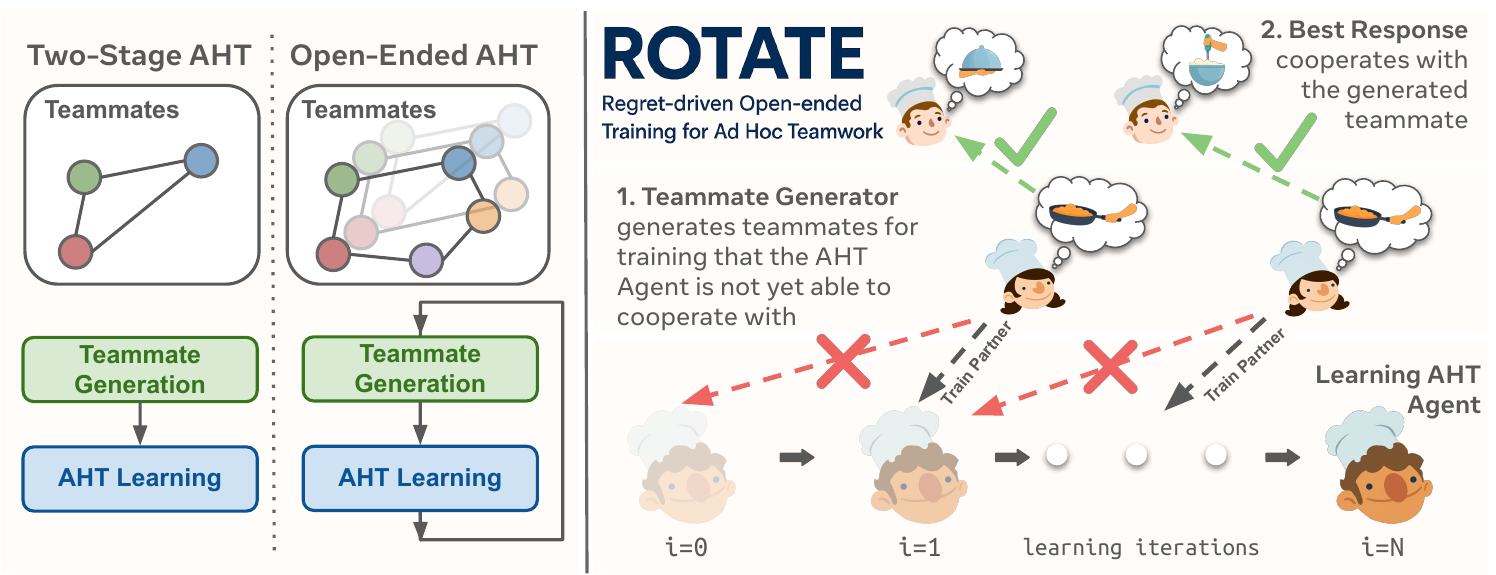}
    \caption{\small\textbf{\ours{} Overview.} \ours{} is an open-ended AHT learning algorithm that improves the AHT agent by iteratively generating diverse teammates with whom the AHT agent struggles yet can still learn to collaborate.}
    \label{fig:method}
\end{wrapfigure}

This paper addresses two issues that cause current methods to fail to learn policies that effectively collaborate with some teammates. 
First, two-stage AHT methods~\citep{papoudakis2020liam, zintgraf2021deep, rahman_towards_2021, rahman2023generating} learn from interacting with teammates from a small fixed training set. Even when the training set is diverse, the AHT agent remains incapable of collaborating effectively with some teammates sampled from the vast space of possible strategies, specifically those with significantly different behavior from the policies in the training set~\citep{vezhnevets2020options, rahman_towards_2021}. 
Second, other works generate diverse teammate behaviors iteratively~\citep{yuan2023learning}, but maximize adversarial diversity~\citep{charakorn2023generating, rahman2023generating}, which yields teammates whose return-diminishing tendencies in certain states make it challenging for a randomly initialized RL-based AHT agent to learn to collaborate with them effectively~\citep{cui2023adversarial, sarkar2023diverse,lauffer2025rpg}.

We address the above issues by proposing an open-ended learning framework that continually generates diverse yet cooperative and competent teammates. We formulate our learning objective by observing that maximizing the expected returns of an AHT agent on a known set of teammates is equivalent to minimizing its expected \textbf{\textit{cooperative regret}}: the utility gap between the best response to a given teammate, and the AHT agent's performance with that teammate. 
We take inspiration from unsupervised environment design (UED) methods~\citep{wang2020enhanced, dennis_emergent_2020, jiang2021replay, rutherford2024no} and train an AHT agent to minimize its regret against generated teammates designed to maximize the AHT agent's cooperative regret.
Unlike UED methods that only maximize regret from starting states, we generate teammates by maximizing regret in all encountered states while imposing competence constraints. This design aims to avoid generating non-cooperative and incompetent teammate policies during training, which are unlikely to be encountered during evaluation. 
Building on these foundations, we propose a practical algorithm, \ours{} (Fig. \ref{fig:method}), which optimizes a cooperative regret-based minimax objective while maintaining a population of all previously designed teammates. 

This paper makes four main contributions.
First, we reformulate AHT as a regret-driven game between a teammate generator and the AHT agent, motivating an open-ended AHT learning procedure that continually generates new teammates.
Second, we characterize \textit{sabotage} as a property of an objective that induces \textit{incompetent} behavior from teammates, and propose a competence-bounded per-state regret objective that provably bounds the competence of generated teammates on states encountered during training.
Third, we propose a novel algorithm, \ours{}, that instantiates the proposed open-ended AHT framework.
Fourth, extensive empirical evaluations demonstrate that \ours{} significantly improves return against unseen teammates compared to representative baselines from AHT and open-ended learning on Overcooked and Level-Based Foraging.

\vspace{-5pt}
\section{Problem Formulation}
\label{sec:problem_formulation}
\vspace{-5pt}
This section provides a brief formalization of the AHT problem explored in existing literature~\cite{mirsky2022survey}. We start by formalizing the interaction between the learning agent and its teammates. We then detail the objectives of existing AHT learning methods.

The interaction between agents in an AHT setting is formalized as a decentralized Markov decision process (Dec-MDP)~\citep{bernstein2002complexity}. 
A Dec-MDP is characterized by a 7-tuple, $\langle N,S,\{\mathcal{A}^{i}\}_{i=1}^{|N|}, P, p_0, R, \gamma \rangle
$, where $N$, $S$, and $\gamma$ respectively denote the index set of agents within an interaction, the state space, and a discount rate, $0 \leq \gamma \leq 1$. 
Every interaction between agents begins from a state sampled from the initial state distribution, $s_{0}\sim p_{0}$. 
At timestep $t$, each agent, $i\in{N}$, jointly executes an action selected from its action space, $a^{i}_{t} \in \mathcal{A}^{i}$, based on the observed state, $s_{t}$, and its policy, $\pi^{i}(s_{t})$. We assume that teammates choose their actions only based on the current state. Meanwhile, the AHT agent, also referred to as the \textit{ego agent}, selects actions based on its state-action history, which is necessary to distinguish between different types of teammates effectively. Denoting the set of all probability distributions over a set $X$ as $\Delta(X)$, the execution of the joint action, $a_{t} = (a^{1}_{t}, \dots, a_{t}^{|N|})$, results in agents observing a new state, $s_{t+1}$, sampled according to the environment transition function, $P:S\times\mathcal{A}^{1}\times\dots\times\mathcal{A}^{|N|}\mapsto\Delta{(S)}$, and a common scalar reward, $r_{t}$, based on the reward function, $R:S\times\mathcal{A}^{1}\times\dots\times\mathcal{A}^{|N|}\mapsto\mathbb{R}$.

Within the Dec-MDP framework, AHT methods aim to train ego agents to achieve maximal returns when collaborating with a set of \textit{unknown yet competent} teammate policies, denoted as $\Pieval$~\citep{stone2010ad}. 
We designate the ego agent as agent $i \in N$, and the remaining $|N|-1$ agents as its \textit{teammates}. Let $\pi^\text{ego}$ refer to the ego agent's policy, and $\pi^{-i}$ denote its $|N|-1$ teammates' joint policy. We denote the return of the team formed by the ego agent and the teammates, starting from state $s_0$, by:
{\eqskips{3pt}{1pt}
\begin{equation}
 \label{def:return_obj}
    V(s_0 | \pi^{-i}, \piego) = \mathbb{E}_{a^{\text{ego}}_{t} \sim \pi^{\text{ego}},\, a^{-i}_{t}\sim \pi^{-i}}\left[\sum\nolimits_{t=0}^{\infty} \gamma^{t}R(s_{t}, a_{t}) \,\middle|\, s_{0}\right].
\end{equation}}%

Given a teammate policy $\pi^{-i}$, $\text{BR}(\pi^{-i})$ is a best response policy to $\pi^{-i}$ if and only if the team policy formed by $\pi^{-i}$ and $\BR(\pi^{-i})$ achieves maximal return, i.e., $\BR(\pi^{-i}) \in \argmax_{\pi} \mathbb{E}_{s_0 \sim p_0} [V(s_0 | \pi^{-i}, \pi)]$.
We refer to interactions between a teammate $\pi^{-i}$ and the ego agent as \textit{cross-play} (\textbf{XP}), and to interactions between $\pi^{-i}$ and its best response $\BR(\pi^{-i})$ as \textit{self-play} (\textbf{SP}).
We define the set of $p_0$-competent teammate policies as the set of teammates that achieve a team return of at least $K$ with their best response, $\Pi^{-i}_K(p_0) = \{ \pi^{-i} : \E_{s_0\sim p_0}[V(s_0 | \pi^{-i}, \BR(\pi^{-i}))] \geq K \}$.
In the AHT problem, evaluation teammates are assumed to be competent, i.e. $\Pieval \subseteq \Pi^{-i}_K(p_0)$.

Let $\psieval(\Pieval)$ denote the probability distribution used during evaluation to sample teammates from $\Pieval$. 
The \textbf{AHT objective} is formalized as finding an ego policy $\piego$ that maximizes the expected return when collaborating with joint teammate policies sampled from $\psieval(\Pieval)$:
{\eqskips{3pt}{5pt}
\begin{equation}
\label{eq:ObjPOAHT}
\max_{\piego} \; V(\psieval, \Pieval, \piego) = \max_{\piego} \;
\mathbb{E}_{\pi^{-i}\sim\psieval(\Pieval),\, s_0 \sim p_0} 
\left[ V\!\left(s_0 \mid \pi^{-i}, \piego \right) \right].
\end{equation}}%
An optimal $\piego$ that maximizes Eq.~\ref{eq:ObjPOAHT} adaptively approximates the \textit{best response policy} performance when collaborating with $\pi^{-i}\in\Pi^{\text{eval}}$.

Since $\Pi^{\text{eval}}$ is unknown, AHT algorithms~\citep{mirsky2022survey} learn by interacting with policies from the training set, $\Pitrain$, which are learned or manually designed by leveraging an expert's domain knowledge about the characteristics of $\Pieval$.
After forming $\Pitrain$, current AHT algorithms use RL to discover an ego agent policy.
While the precise training objective varies with the AHT algorithm, methods commonly maximize the expected return during interactions with teammate policies sampled uniformly from $\Pitrain$:
{\eqskips{3pt}{1pt}
\begin{equation}
\label{def:LearningObjPOAHT}
\pi^{*,\text{ego}}(\Pi^{\mathrm{train}}) \in
\underset{\pi^{\text{ego}}}{\mathrm{argmax}}\;
\mathbb{E}_{\pi^{-i}\sim\mathcal{U}(\Pi^{\text{train}}),\, s_0 \sim p_0} \left[ V\!\left(s_0 \mid \pi^{-i}, \pi^{\text{ego}} \right) \right].
\end{equation}}%
Naturally, $\pi^{*, \text{ego}}(\Pitrain)$ may be suboptimal with respect to $\Pieval$ and $\psieval$, if there is distribution shift between the training and evaluation objectives.
\vspace{-5pt}
\section{Ad Hoc Teamwork as an Open-Ended Learning Problem}
\label{sec:oeaht}
\vspace{-5pt}
In this section, we reformulate the AHT objective (Eq.~\ref{eq:ObjPOAHT}) as a minimax regret game between a teammate generator and an ego agent learner, suggesting an open-ended learning procedure to address AHT problems.
We show that for a fixed set of teammates $\Pieval$ and sampling distribution $\psieval$ over $\Pieval$, maximizing the return of the ego agent is equivalent to minimizing its cooperative regret.
In the absence of knowledge about $\Pieval$ and $\psieval$, we argue that minimizing the \textit{worst-case} cooperative regret of the ego agent with respect to regret-maximizing, competent teammates leads to ego agents that cooperate well with unknown, competent teammates. 
Based on this argument, we propose a novel \textit{minimax regret} objective (Eq.~\ref{eq:minimax_regret_game}).
Finally, we present an algorithmic framework for optimizing the minimax regret objective in an iterative fashion (Alg.~\ref{alg:OpenEndedFramework}).

\begin{definition}[Cooperative Regret]\label{def:coop_regret}
The \textbf{cooperative regret} of an ego policy $\piego$ when interacting with a joint teammate policy $\pi^{-i}$ from a starting state $s_0$ is:
\begin{equation*}
    \text{CR}(\piego, \pi^{-i}, s_0) = V\left(s_0 | \pi^{-i}, \text{ BR}(\pi^{-i})\right) - V\left(s_0 |\pi^{-i}, \piego \right).
\end{equation*}
\end{definition}
Observe that any ego policy that \textit{maximizes} the AHT objective (Eq.~\ref{eq:ObjPOAHT}) must also \textit{minimize} the expected regret over joint teammate policies sampled from $\psieval(\Pieval)$:
{\eqskips{3pt}{1pt}
\begin{equation}
    \label{eq:expected_regret}
    \text{CR}(\psi^{\text{eval}}, \Pi^{\text{eval}}, \piego) = \mathbb{E}_{\pi^{-i}\sim{\psi^{\text{eval}}(\Pi^{\text{eval}})}, s_0 \sim p_{0}} \left[\text{CR}(\piego, \pi^{-i}, s_0)\right],
\end{equation}}%
which results from $V\left(s_0 | \pi^{-i}, \text{ BR}(\pi^{-i})\right)$ being independent of $\piego$ for any $\pi^{-i}$ and $s_0$.

The above equivalence suggests training ego policies to minimize the expected cooperative regret, $\text{CR}(\psieval, \Pieval, \piego)$. 
However, as $\Pieval$ and $\psieval$ are unknown, we propose optimizing $\piego$ to minimize the \textit{worst-case regret} that could be induced by any teammate policy within the set of $p_0$-competent teammates, $\pi^{-i} \in \Pi^{-i}_K(p_0)$:
{\eqskips{3pt}{0pt}
\begin{equation}
\label{eq:minimax_regret_game}
    \underset{\piego}{\min\ } \underset{\pi^{-i}\in\Pi^{-i}_K(p_0)}{\max\ } 
     \mathbb{E}_{s_0 \sim p_{0}} \left[\text{CR} (\piego, \pi^{-i}, s_{0})\right].
\end{equation}}%

The minimax regret game in Eq.~\ref{eq:minimax_regret_game} has two properties that make it a suitable training objective for AHT.
First, for the ego agent, finding $\piego$ that achieves zero worst-case regret is equivalent to finding an ego agent that achieves the best-response return with any joint teammate policy $\pi^{-i}$. If such a $\piego$ exists, then this AHT agent maximizes Eq.~\ref{eq:ObjPOAHT} for any $\psi^{\text{eval}}$ and $\Pi^{\text{eval}}$.
However, its existence is not guaranteed.\footnote{Consider a one-step task in which two agents simultaneously retrieve one of two objects and succeed only if the objects differ. Against a teammate that always retrieves the first object and another that always retrieves the second, any ego policy has positive regret with at least one of the two teammates.}
In practice, we are therefore content with \textit{minimizing} the worst-case regret.
Second, for the teammate, maximizing the regret of the ego agent encourages discovering $\pi^{-i}$ for which the ego agent policy has substantial room for improvement, inducing a curriculum for the ego agent learner.
In UED, similar minimax regret formulations have successfully induced curricula to improve the ability of RL agents to generalize to unseen environment variations~\citep{dennis_emergent_2020,wang2020enhanced}.
Note that regret maximization does not require access to an exact best response, so $\text{BR}(\pi^{-i})$ may be trained with RL, rather than assumed to be available.\footnote{Define the regret as $V(s_0|\pi^{-i},\pi^\text{tm}) - V(s_0|\pi^{-i},\piego)$, where $(\pi^{-i},\pi^\text{tm})$ jointly maximize regret while $\piego$ minimizes it. To maximize regret, $\pi^\text{tm}$ can only act on the first term to maximize the value with $\pi^{-i}$, recovering our original minimax regret formulation (cf. Thm.~2 of \citet{dennis_emergent_2020}). Under this view, $\pi^\text{tm}$ need not be the exact best response for the objective to be well posed.}

In Eq.~\ref{eq:minimax_regret_game}, constraining regret maximization to $p_0$-competent teammates is especially important, both to align the optimization procedure to the AHT problem definition, and because unconstrained regret maximization may lead to incompetent teammates that perform poorly against any $\pi^{\text{ego}}$. 
Such teammates are undesirable because they introduce unnecessary learning challenges for RL-based AHT learning algorithms, and are unlikely to be encountered in coordination scenarios. 
We expand upon this discussion in Sec.~\ref{sec:method_teamgen}, where we propose a practical regret-maximization objective.

\vspace{-6pt}
\begin{algorithm}[H]
   \caption{Open-Ended Ad Hoc Teamwork Framework}
   \label{alg:OpenEndedFramework}
    \begin{algorithmic}[1]
       \STATE {\bfseries Input:} Environment $\text{Env}$, iterations $T^{\text{iter}}$, initial ego parameters $\theta^{\text{ego}}$
       \STATE Initialize teammate policy buffer $\text{B}_{\pi} \leftarrow \emptyset$
       \FOR{$j=1$ {\bfseries to} $T^{\text{iter}}$}
          \STATE $\text{B}^{\text{new}}_{\pi} \leftarrow \text{TeammateGenerator}(\text{Env}, \theta^{\text{ego}}, \text{B}_{\pi})$ \textcolor{blue}{// Train teammates to maximize regret}
          \STATE $\theta^{\text{ego}} \leftarrow \text{EgoUpdate}(\text{Env}, \theta^{\text{ego}}, \text{B}^{\text{new}}_{\pi})$ \textcolor{blue}{// Train ego agents to minimize regret}
          \STATE $\text{B}_{\pi} \leftarrow \text{B}^{\text{new}}_{\pi}$
       \ENDFOR
       \STATE {\bfseries Output:} Final ego policy $\theta^{\text{ego}}$
    \end{algorithmic}
\end{algorithm}
\vspace{-12pt}
Alg.~\ref{alg:OpenEndedFramework} outlines our general framework for training an ego agent to minimize the worst-case cooperative regret induced by any teammate $\pi^{-i} \in \Pi^{-i}_K(p_0)$.
Like coordinate ascent algorithms~\citep{d1959convex}, the framework alternates between optimizing for $\pi^{-i}$ and $\piego$ for $T^\text{iter}$ iterations, while holding the other fixed.
We call a phase in which $\piego$ is fixed and $\pi^{-i}$ is updated to maximize the ego agent's regret the \textbf{teammate generation phase} (Line 4).
Meanwhile, assuming that $\pi^{-i}$ is fixed, the \textbf{ego agent update phase} (Line 5) updates $\piego$ to minimize regret.

Sec.~\ref{sec:method_teamgen} \& \ref{sec:method} describe \ours, our practical instantiation of Alg. \ref{alg:OpenEndedFramework} as an open-ended learning approach for AHT. Sec.~\ref{sec:method_teamgen} focuses on generating regret-maximizing teammate policies that remain competent. Meanwhile, Sec.~\ref{sec:method} covers the practical experience collection process for \ours's teammate generation and ego agent update phases.
\ours{} is an open-ended learning procedure according to the definition of~\citet{hughes_position_2024} because it continually generates \textit{novel} yet \textit{learnable} artifacts (i.e., teammates) for an observer (i.e., ego agent), as discussed in App.~\ref{app:supp_disc:formalize_oel}.

\newcommand{\trajobjbody}{\underset{\pi^{-i}}{\max\ } 
    \mathbb{E}_{s_0 \sim p_{0}} \left[\text{CR} (\piego, \pi^{-i}, s_{0}) + \lambda V\left(s_0|\pi^{-i},\BR(\pi^{-i})\right)\right]}
\newcommand{\stateobjbody}{J(\pi^{-i},\piego) &= 2\E _{s_0\sim \mathcal{U}(\pxp,\text{ } \psp)}\left[\text{CR}\left( \piego, \pi^{-i}, s_0 \right)\right] \\ &+ \lambda_1\E _{s_0\sim\pxp}\left[V\left(s_0|\pi^{-i}, \BR(\pi^{-i})\right) \right] \\ &+ \lambda_2\E _{s_0\sim\psp}\left[V\left(s_0|\pi^{-i}, \BR(\pi^{-i})\right) \right]}
\vspace{-5pt}
\section{Generating Regret-Maximizing Yet Competent Teammates}
\label{sec:method_teamgen}
\vspace{-5pt}
Given a fixed $\piego$, we wish to generate teammate policies that maximize the cooperative regret of $\piego$. 
However, naive regret maximization may lead to incompetent teammates that act adversarially towards the ego agent, introducing learning challenges. 
We propose a novel regret-based objective that addresses the problem in two steps. 
First, teammate competence is bounded on the initial state distribution. 
Second, because a teammate that is competent from $p_0$ may still be incompetent at the states into which it steers the ego agent, we define \textit{sabotage} as a property of a teammate generation objective and show that bounding competence from $p_0$ does not remove the incentive to sabotage. Optimizing competence-bounded regret from all states visited during training then yields the \ours{} teammate generation objective, which bounds teammate competence on those states.
The presented theoretical results are proven in App.~\ref{app:theory:proofs}.

\subsection{Ensuring Teammate Competence}
\label{sec:bounding_competence}
\vspace{-5pt}
The most direct instantiation of the regret maximization component of Eq.~\ref{eq:minimax_regret_game} as an RL objective is the following \textbf{per-trajectory regret} objective: 
{\eqskips{3pt}{1pt}
\begin{equation}
    \label{eq:traj_regret_obj}
    \underset{\pi^{-i}}{\max\ } 
    \mathbb{E}_{\textcolor{blue}{s_0 \sim p_{0}}} \left[\text{CR} (\piego, \pi^{-i}, s_{0})\right].
\end{equation}}%
Eq.~\ref{eq:traj_regret_obj} is the regret over trajectories starting from the initial state distribution. 
The per-trajectory regret objective corresponds directly to the adversarial diversity objectives previously proposed in the teammate generation~\citep{rahman2023generating,charakorn2023generating} and UED literature~\citep{wang2020enhanced, dennis_emergent_2020}.

However, maximizing per-trajectory regret alone does not ensure that the generated teammate is competent. 
In App.~\ref{app:theory:examples}, a matrix game example shows that a regret-maximizing teammate can minimize returns against any $\piego$.
Prior AHT work has observed that adversarial diversity objectives frequently yield teammates that minimize their partners' returns~\citep{cui2023adversarial, sarkar2023diverse, charakorn2024sabotage, lauffer2025rpg}, and has proposed modifications to the objective to counteract this behavior.
Because $\Pieval$ is assumed to consist of $p_0$-competent teammates, training against incompetent teammates exposes the ego agent to teammates unlikely to be encountered during evaluation.
Furthermore, because an incompetent teammate yields low returns regardless of the ego agent's actions, training against such a teammate provides little signal about the coordination behaviors that competent teammates reward, so the ego agent may fail to cooperate well even with the teammates that it does encounter at evaluation time.

To ensure that maximizing regret leads to $p_0$-competent teammates, we augment per-trajectory regret with the $\lambda$-weighted return of the teammate and its best response, yielding the \textbf{competence-bounded per-trajectory regret} objective:
{\eqskips{3pt}{1pt}
\begin{equation}
    \label{eq:traj_regret_obj_lambda}
    \trajobjbody.
\end{equation}}%
The following theorem characterizes the competence that each value of $\lambda$ enforces on any $\pi^{-i}$ that maximizes Eq.~\ref{eq:traj_regret_obj_lambda}.\footnote{Eq.~\ref{eq:traj_regret_obj_lambda} is the Lagrangian of the problem of maximizing per-trajectory regret subject to the constraint $\pi^{-i} \in \Pi^{-i}_K(p_0)$, with $\lambda$ held fixed. Primal-dual methods for single-agent constrained RL rely on the strong duality of constrained MDPs~\citep{paternain2019constrained}, which relies on the linearity of the objective in the occupancy measure induced by the learning agent. Here, the objective involves the BR value $\max_{\pi'} V(s_0|\pi^{-i},\pi')$, which is not linear in the occupancy measure of $\pi^{-i}$. Thus, strong duality does not follow from~\citet{paternain2019constrained}. Theorem~\ref{thm:traj_competence} instead directly characterizes the competence attained with a fixed $\lambda$.}
Choosing $\lambda$ greater than the provided constant yields teammates that are $p_0$-competent in the same sense assumed of $\Pieval$.

\begin{restatable}[Per-Trajectory Competence Lower Bound]{theorem}{trajcompetencetheorem}
\label{thm:traj_competence}
Let $\lambda \in \mathbb{R}^{> 0}$, and let $R^{\max}_{p_0} = \max_{\pi'} \E_{s_0 \sim p_0}\left[V(s_0 | \pi', \BR(\pi'))\right]$ and $R^{\min}_{p_0} = \min_{\pi'} \E_{s_0 \sim p_0}\left[V(s_0 | \pi', \piego)\right]$.
If $\pi^{-i}$ maximizes Eq.~\ref{eq:traj_regret_obj_lambda}, then
$\E_{s_0\sim p_0}\left[V\left(s_0|\pi^{-i},\BR(\pi^{-i})\right)\right] \geq \frac{\lambda}{\lambda+1}R^{\max}_{p_0} + \frac{1}{\lambda+1}R^{\min}_{p_0}.$
Consequently, for any $K \in [R^{\min}_{p_0}, R^{\max}_{p_0})$, choosing $\lambda \geq \frac{K - R^{\min}_{p_0}}{R^{\max}_{p_0} - K}$ guarantees $\pi^{-i} \in \Pi^{-i}_K(p_0)$.
\end{restatable}

\subsection{Ensuring Teammate Competence on a Per-State Basis}
\label{sec:per_state_regret}

\vspace{-5pt}

An objective incentivizes \textbf{sabotage} if a teammate can increase the objective by becoming \textit{incompetent} starting from states visited during XP interactions between $\pi^{-i}$ and $\piego$, which lowers the return that \textit{any} partner can obtain from those states.
Competence-bounded per-trajectory regret alone (Eq.~\ref{eq:traj_regret_obj_lambda}) does not mitigate sabotage because $p_0$-competence constrains only the return that $\pi^{-i}$ achieves with its BR from $p_0$. 
Thus, incompetence at states visited only with $\piego$ raises regret without lowering $p_0$-competence.
For example, a teammate that acts competently only at states reached with its BR and incompetently everywhere else is $p_0$-competent and achieves high per-trajectory regret, yet obtains low return with any $\piego$ that ever strays from the BR's states.

Teammates generated by an objective that incentivizes sabotage are undesirable for two reasons.
First, such teammates are not expected to be encountered in AHT, as the teammates in $\Pieval$ are assumed to act in good faith towards the shared task. 
Second, they introduce learning challenges (see Sec.~\ref{sec:bounding_competence}).
This problem is not encountered by typical UED approaches~\cite{dennis_emergent_2020, jiang2021prioritized, rutherford2024no}, which generate only an environment's initial state under fixed transition dynamics.

\begin{wrapfigure}[15]{r}{0.5\textwidth} 
    \vspace{-10pt}
    \centering
    \includegraphics[width=\linewidth]{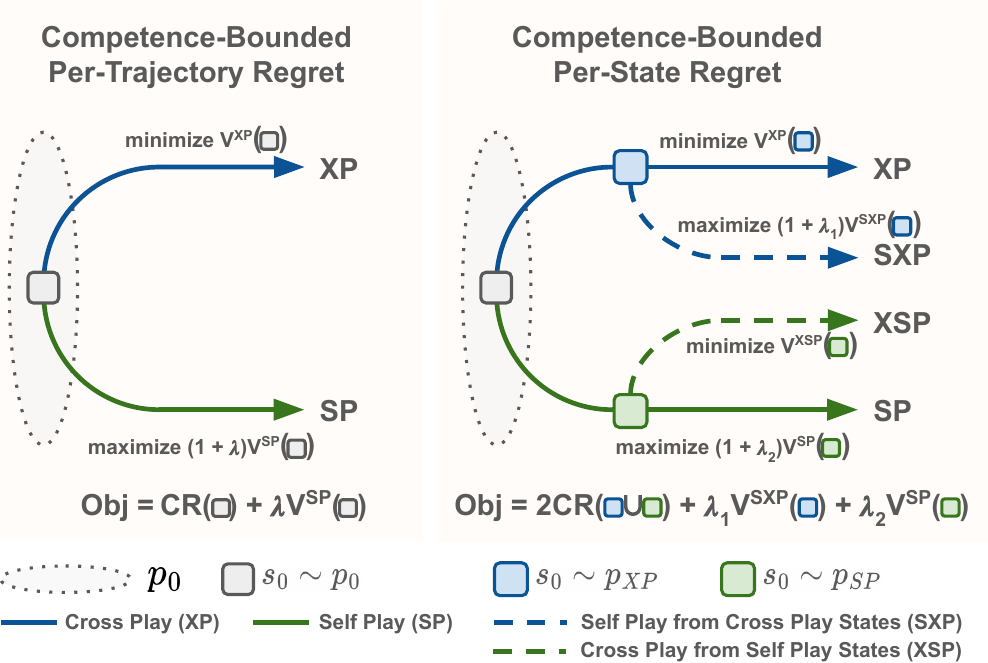}
    \caption{\small Teammate policy optimization objectives.}
    \label{fig:regret}
\end{wrapfigure}
Sabotage is defined as a property of an objective rather than of a policy, as sabotage suggests deliberate intent, which is encoded by the objective.
For example, the same incompetent behavior may arise from two objectives that represent very different incentives.
Consider a robot chef that cooks competently in a single-width corridor kitchen but performs a safety stop for the remainder of the episode if a partner draws too close. 
The robot is $p_0$-competent, since its BR keeps clear and never triggers the stop, but per-state incompetent once the stop triggers, since no partner can complete a dish from a blocked kitchen. 
The robot could learn to stop as a side effect of a safety-prioritizing objective, or because halting on states reached only with $\piego$ raises the competence-bounded per-trajectory regret of $\piego$ at no cost to $p_0$-competence. 
Characterizing sabotage as a behavioral property fails to capture the difference between these two learning scenarios.

Prior work characterizes sabotage as return-minimizing behavior of a teammate when interacting with partners other than its BR~\citep{cui2023adversarial, sarkar2023diverse, charakorn2024sabotage}.
Our definition differs in two respects.
First, sabotage is defined as a property of an objective rather than of a policy, as argued above.
Second, sabotage is defined in terms of per-state incompetence rather than return minimization, so that the return is evaluated with respect to the \textit{best possible partner} from the affected states, rather than with the non-BR partner that a teammate is currently interacting with.\footnote{Note that the behaviors identified as sabotage in prior work, such as deliberately playing unplayable cards in Hanabi~\citep{cui2023adversarial}, lower the best possible return from the affected states and are therefore also per-state incompetent.}
In fact, given that the competence constraints on teammate behavior are active, it is \textit{desirable} for the teammate to minimize the return of $\piego$ because it indicates that the teammate has learned a convention that the ego agent does not know.
The self-sabotage definition of \citet{lauffer2025rpg} likewise defines sabotage as a property of an objective, but constrains the teammate differently. 
It requires the teammate to be a BR to some partner from $p_0$ only, whereas per-state competence requires that the teammate can achieve return above a threshold with some partner from every state the ego agent encounters.
Neither condition implies the other, since a teammate may best-respond to a partner from $p_0$ while sabotaging states that partner never visits, and a competent teammate need not be optimal for any partner.

We mitigate the incentive to sabotage by training $\pi^{-i}$ to maximize regret \textit{while remaining competent} starting from the distribution of \textit{all} states visited by the teammate during training.
In particular, regret estimation involves the SP and XP state visitation distributions.
Let $d(\pi^1, \pi^2; p)$ denote the state visitation distribution when $\pi^1$ and $\pi^2$ interact based on a starting state distribution $p$. The SP and XP state visitation distributions are denoted by the shorthand $\psp(\pi^{-i}) := d(\pi^{-i}, \text{BR}(\pi^{-i}); p_0)$ and $\pxp(\pi^{-i}) := d(\pi^{-i}, \pi^{\text{ego}}; p_0)$.
We write $\psp$ and $\pxp$ for $\psp(\pi^{-i})$ and $\pxp(\pi^{-i})$ when the teammate is clear from context.
Based on $\psp$ and $\pxp$, the \textbf{per-state regret} of the ego agent is defined as the regret $\text{CR}(\piego, \pi^{-i}, s_0)$ with $s_0$ drawn from $\mathcal{U}(\pxp, \psp)$, i.e., from $\pxp$ or $\psp$ with equal probability.

\begin{wrapfigure}{r}{0.6\textwidth}
    \vspace{-14pt}
    \refstepcounter{equation}\label{eq:constrained_state_based_regret_obj}
    \noindent
     \fbox{%
     \begin{minipage}{\dimexpr\linewidth-2\fboxsep-2\fboxrule\relax}
     \vspace{-6pt}
     \begin{align*}
     \stateobjbody. \tag{\theequation}
     \end{align*}
     \end{minipage}%
     }
    \vspace{-8pt}
\end{wrapfigure}
Analogous to Sec.~\ref{sec:bounding_competence}, per-state regret is augmented with the $\lambda$-weighted return of the teammate and its BR to obtain the final \ours{} teammate generation objective, which we refer to as \textbf{competence-bounded per-state regret}. 
Here, $\BR(\pi^{-i})$ is the best response to $\pi^{-i}$ at every visited state rather than only from $p_0$, so that $V(s_0|\pi^{-i}, \BR(\pi^{-i}))$ measures the best return that some partner can achieve with $\pi^{-i}$ from arbitrary start state distributions.

Optimizing competence-bounded per-state regret (Eq.~\ref{eq:constrained_state_based_regret_obj}) offers two intuitive benefits. 
First, it yields a $\pi^{-i}$ with which $\piego$ can improve its action selection across a broader range of states. 
Second, maximizing regret from SP states disincentivizes signals for unconditional cooperation (or \textit{handshake conventions} \citep{sarkar2023diverse}) by requiring $\pi^{-i}$ to decrease the ego agent's return starting from any SP state. 

The following theorem bounds the competence of any $\pi^{-i}$ that maximizes Eq.~\ref{eq:constrained_state_based_regret_obj}.

\begin{restatable}[Per-State Competence Lower Bound]{theorem}{competencetheorem}
\label{thm:competence2}
Let $\lambda_1, \lambda_2 \in \mathbb{R}^{\geq 0}$ with $\Lambda := \lambda_1 + \lambda_2 > 0$. For any teammate $\pi'$, let $\pxp(\pi')$ and $\psp(\pi')$ denote the XP and SP state visitation distributions induced by $\pi'$, and define the mixtures $U(\pi') := \mathcal{U}(\pxp(\pi'), \psp(\pi'))$, $\mu(\pi') := \tfrac{\lambda_1}{\Lambda}\pxp(\pi') + \tfrac{\lambda_2}{\Lambda}\psp(\pi')$, and $\nu(\pi') := \tfrac{\lambda_1+1}{\Lambda+2}\pxp(\pi') + \tfrac{\lambda_2+1}{\Lambda+2}\psp(\pi')$.
Let $R^{\max}_{\mu} = \max_{\pi'} \mathbb{E}_{s_0 \sim \mu(\pi')}\left[V(s_0 | \pi', \BR(\pi'))\right]$ and $R^{\min}_{U} = \min_{\pi'}\E_{s_0\sim U(\pi')}\left[V\left(s_0|\pi',\piego\right)\right]$.
If $\pi^{-i}$ maximizes Eq.~\ref{eq:constrained_state_based_regret_obj}, then
\begin{equation*}
\E_{s_0\sim \nu(\pi^{-i})}\left[V\left(s_0|\pi^{-i},\BR(\pi^{-i})\right)\right]
\geq \frac{\Lambda}{\Lambda+2}R^{\max}_{\mu} + \frac{2}{\Lambda+2}R^{\min}_{U}.
\end{equation*}
\end{restatable}

$R^{\min}_{U}$ and $R^{\max}_{\mu}$ are teammate-independent, fixed quantities.
Theorem~\ref{thm:competence2} guarantees a minimum level of competence for any teammate that maximizes the $\ours$ objective in Eq.~\ref{eq:constrained_state_based_regret_obj}. 
As the total competence weight increases, if the ratio between $\lambda_1$ and $\lambda_2$ is fixed, the bound increases monotonically from $R^{\min}_{U}$ towards $R^{\max}_{\mu}$. 
Thus, the competence weights control the strength of the guarantee.
Unlike Theorem~\ref{thm:traj_competence}, Theorem~\ref{thm:competence2} certifies competence with respect to $\nu(\pi^{-i})$, the states encountered during ego training, rather than to $p_0$. 
Thus, compared to Eq.~\ref{eq:traj_regret_obj_lambda}, Eq.~\ref{eq:constrained_state_based_regret_obj} trades alignment with the definition of $\Pieval$~\citep{stone2010ad} for cooperativeness on the states over which the ego agent learns.

\vspace{-5pt}
\section{Practical Algorithm: \ours{}}
\label{sec:method}
\vspace{-5pt}
This section presents \ours{}, our practical algorithm that instantiates Alg.~\ref{alg:OpenEndedFramework} by specifying the \texttt{TeammateGenerator} and \texttt{EgoUpdate} procedures.
In \ours, all policies are estimated using Proximal Policy Optimization (PPO)~\citep{Schulman2017ProximalPO}.
See App.~\ref{app:algorithm_pseudocode} for a detailed description and pseudocode.
\paragraph{\ours{} Teammate Generator}
\vspace{-10pt}
Optimizing Eq.~\ref{eq:constrained_state_based_regret_obj} requires a different data collection process compared to the usual RL methods.
While maximizing $V(s_0 | \pi^{-i}, \BR(\pi^{-i}))$ over $\psp$ is straightforward, maximizing the other terms from Eq.~\ref{eq:constrained_state_based_regret_obj} requires collecting experiences using either environment resetting or policy switching. 
Taking the maximization of $V(s_0 | \pi^{-i}, \BR(\pi^{-i}))$ from $s_0\sim\pxp$ as an example, if an environment supports resetting to arbitrary states, then states encountered during XP interaction can be stored and used as the initial state for SP interactions. 
We call such data ``self-play from cross-play'' (SXP) data.
Otherwise, we may sample a random timestep $t$, run XP interaction until timestep $t$, and then switch to SP interaction~\citep{sarkar2023diverse}. 
Only data gathered after timestep $t$ is used to optimize the expected value of $V(s_0 | \pi^{-i}, \BR(\pi^{-i}))$ over $\pxp$. 
The same SXP data is used to train $\BR(\pi^{-i})$, so that it is a best response to $\pi^{-i}$ at every state the teammate is observed in, as required in Sec.~\ref{sec:per_state_regret}.
Both data collection schemes assume it is possible for agent policies to act starting from arbitrary points within an interaction. While the assumption is straightforwardly satisfied by fully observable settings with Markov states, it may not be in partially observable settings, in which teammates typically require memory. \ours{} extends to this setting by conditioning non-recurrent agent policies on a shared recurrent belief state (App.~\ref{app:supp:partial_obs}).
\vspace{-10pt}
\paragraph{\ours{} Ego Agent Update}
At each iteration, \ours{} creates a teammate that highlights the cooperative weaknesses of the previous iteration's ego agent by maximizing the ego agent's per-state regret. 
To allow the ego agent to improve its robustness over time and reduce the possibility that it forgets how to cope with previously generated teammates, \ours{} maintains a \textit{population buffer} of generated teammates. 
At each iteration, the ego agent is trained with teammates sampled uniformly from the population buffer.
While prior AHT methods that train against fixed teammate populations have proposed worst-case teammate sampling strategies~\citep{erlebach2024raccoon,villin2025minimaxapproachadhoc}, worst-case sampling risks concentrating ego agent training on teammates with irreducible regret (e.g., due to convention conflict), causing ego learning to stagnate~\citep{beukman2024refining}. 
Uniform sampling does not remove adversarial pressure from ego training, as the teammate generator can regenerate any convention in the population buffer that the ego agent is unable to coordinate with. 
Combining the \ours{} teammate generator with adversarial buffer curation is left for future work (Sec.~\ref{sec:discussion}).

\vspace{-5pt}
\section{Experimental Results}
\label{sec:exps}
\vspace{-10pt}
This section presents the empirical evaluation of \ours{} compared to baseline methods, as well as several ablations. Code and supplemental results are in the Appendix.
The main research questions are: 
\begin{itemize}[leftmargin=*, itemsep=0pt, topsep=0pt, partopsep=0pt, parsep=0pt]
    \item \textbf{RQ1}: Does \ours{} generalize better to unseen teammates than AHT and UED baselines? \textcolor{blue}{(Yes.)}
    \item \textbf{RQ2}: Does the ROTATE teammate generation objective mitigate the emergence of teammates that are difficult to learn with? \textcolor{blue}{(Yes.)}
    \item \textbf{RQ3}: What is the influence of the competence weights, $\lambda_1, \lambda_2$, on generalization performance?
\end{itemize}

\vspace{-5pt}
\subsection{Experimental Setup}
\label{sec:exp:setup}
This section describes the tasks, baselines, evaluation set construction, and evaluation metric.

\textbf{Tasks.} 
\ours{} is evaluated on a didactic matrix game and six benchmark tasks. %
The benchmark tasks are Level-Based Foraging (LBF)~\citep{albrecht_game-theoretic_2013} and the five classic layouts from the Overcooked suite~\citep{carroll_utility_2019}: Cramped Room (CR), Asymmetric Advantages (AA), Counter Circuit (CC), Coordination Ring (CoR), and Forced Coordination (FC). 
All tasks support multiple conventions, and are commonly used in the AHT literature~\citep{albrecht_game-theoretic_2013,christianos2020shared, papoudakis2020liam}.
In LBF, two agents cooperate to collect randomly placed apples in a gridworld. In Overcooked, two agents collaborate in kitchen gridworlds to prepare dishes.

\textbf{Baselines.} 
As our method is most closely related to methods from UED and teammate generation, we compare against two UED methods adapted for AHT (PAIRED~\citep{dennis_emergent_2020}, Minimax Return~\citep{morimoto_robust_2005}), three AHT algorithms focused on teammate generation (Fictitious Co-Play~\citep{strouse_fcp_2022}, BRDiv~\citep{rahman2023generating}, CoMeDi~\citep{sarkar2023diverse}), and an open-ended AHT method based on an empirical game-theoretic framework (COLE~\citep{li_cole_2023}).
Curator-based UED methods such as PLR~\citep{jiang2021replay, jiang2021prioritized} are excluded, as they are orthogonal to ours~\citep{erlebach2024raccoon, villin2025minimaxapproachadhoc, chaudhary2025improvinghumanaicoordinationadversarial}. 
Similarly, we do not compare against AHT algorithms for ego learning~\citep{ALBRECHT201866}.
Fuller descriptions of each baseline, including how the UED baselines are adapted for AHT, are provided in App. \ref{app:baselines}. 
For fair comparison, all methods received manual and automated hyperparameter tuning and were trained for the same number of environment interactions (App.~\ref{app:hyperparams}).

\textbf{Construction of $\Pieval$.}  
We wish to evaluate all methods on as diverse a set of evaluation teammates as feasible, while ensuring that each teammate is competent. 
To achieve this goal, for each task, we construct 9 to 13 evaluation teammates using three strategies: independent PPO (IPPO)~\citep{yu_surprising_2022} with varied seeds and reward shaping, BRDiv, and manually programmed heuristic agents.
The evaluation set sizes and sources are comparable to those of prior AHT work~\citep{wang2024nagent, rahman2024minimum, charakorn2023generating}. 
Candidate teammates that added no diversity, as measured by the cross-play matrix among evaluation teammates, were pruned (\cref{fig:heldout_xp_matrices}).
App.~\ref{app:pieval_details} fully describes the generation procedure and the final teammates in $\Pieval$.

\textbf{Evaluation Metrics.}
Ego agent policies are evaluated with each teammate in $\Pieval$ for 64 evaluation episodes, where the episode return is normalized with a zero lower bound and an estimated best-response upper bound for each teammate.
The upper bounds are obtained by training best-response policies via RL, then revising each bound upward whenever any evaluated partner policy exceeds it.
Performance of a method on $\Pieval$ is primarily reported as the normalized mean return with bootstrapped 95\% confidence intervals across six trials for the core comparison (Fig.~\ref{fig:rotate_main_results}~(left)) and three trials for all other experiments, computed via the \texttt{rliable} library~\citep{agarwal_deep_2021}.
The normalized return metric is a member of the BRProx metric family recommended by \citet{wang_zsc-eval_2024}.
Following \citet{agarwal_deep_2021}, 95\% bootstrapped confidence interval estimates are used to summarize comparisons rather than significance tests. Because changing the evaluation teammate changes the transition and reward dynamics faced by the ego agent, each evaluation teammate is treated as a distinct evaluation task, and the bootstrap resamples training seeds within each evaluation task.
Details about the normalization procedure and specific bounds for each teammate are reported in App. \ref{app:lower_upper_return_bounds}.
\vspace{-10pt}
\subsection{Results}
\label{sec:exp:results}
\vspace{-5pt}
This section addresses the research questions introduced at the beginning of Sec.~\ref{sec:exps}. 
Extensive supplemental analysis is provided in the appendix, most of which is highlighted within this section.
Among the analyses that are not highlighted, App.~\ref{app:supp:teammate_gen_ablations} considers ablations of \ours{} in which the teammate generation objective is replaced by competence-bounded per-trajectory regret or a mixed-play variant, while App.~\ref{app:supp:ind_ego_agent_pop} ablates the population buffer and analyzes the independent utility of the \ours{} population. %
Finally, App.~\ref{app:supp:learning_curves} provides learning curves for all variants of \ours{}.

\begin{figure*}
    \centering
    \begin{minipage}[t]{0.66\linewidth}
        \centering
        \includegraphics[width=\linewidth]{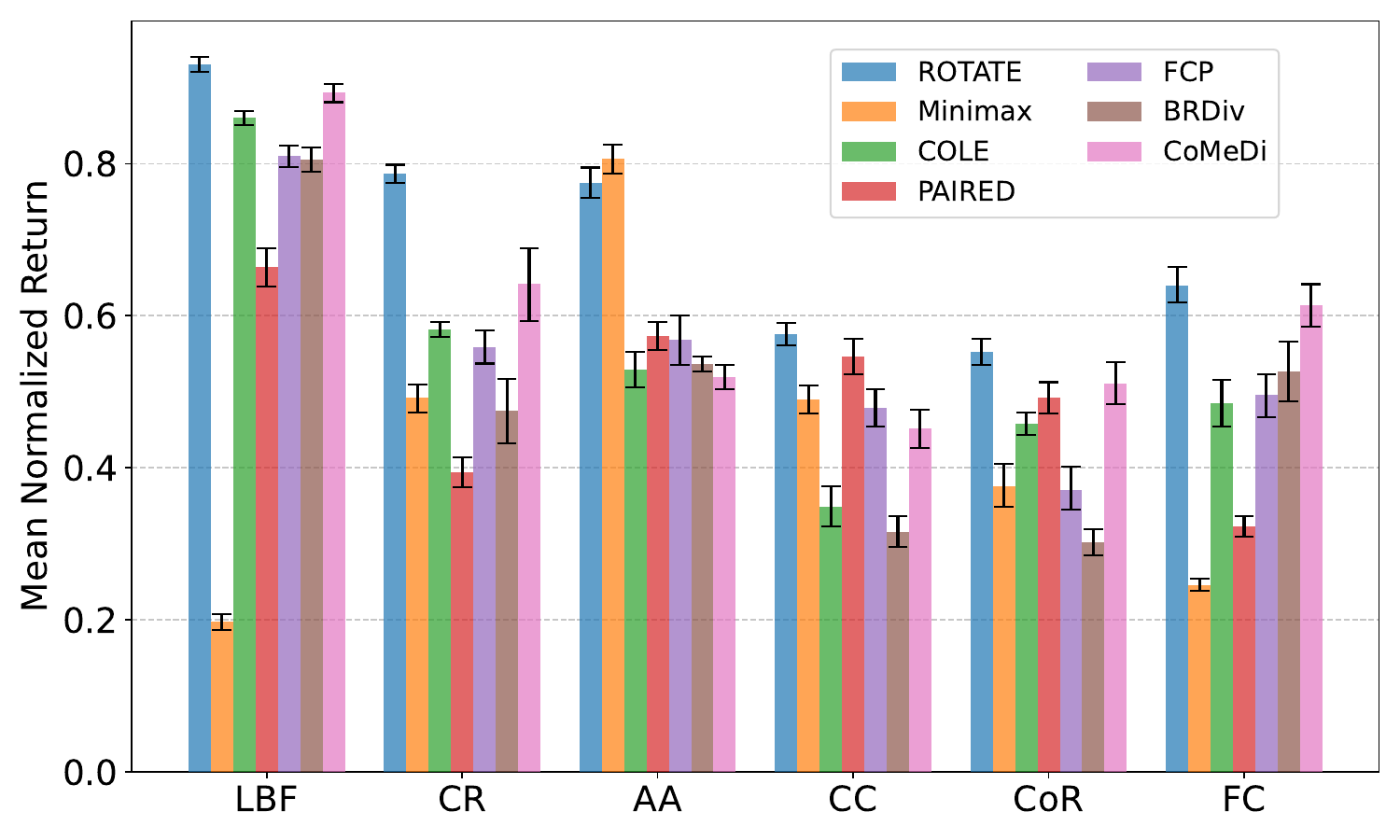}
    \end{minipage}
    \hfil
    \begin{minipage}[t]{0.235\linewidth}
    \centering
        \includegraphics[width=\linewidth]{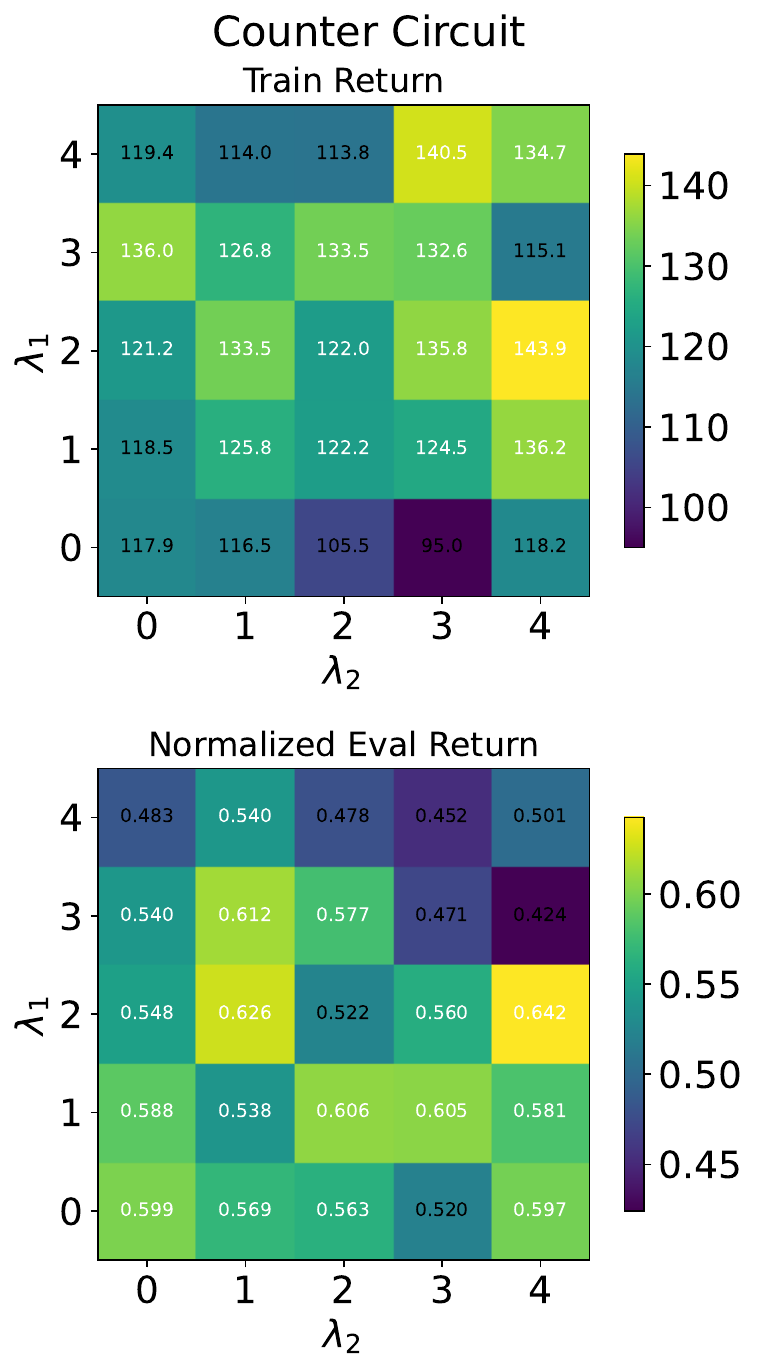}
    \end{minipage}
    \caption{\small \textbf{(Left)} \ours{} achieves the highest normalized evaluation return on five of six tasks compared to baselines. \textbf{(Right)} Higher competence weights ($\lambda_1$, $\lambda_2$) lead to higher training returns, reflecting the increased cooperativity of training teammates. However, intermediate competence weights correspond to the highest evaluation returns, demonstrating that the most cooperative teammates are not necessarily the best training partners for learning generalizable cooperative behaviors.}
    \label{fig:rotate_main_results}
    \vspace{-12pt}
\end{figure*}
\vspace{-.25cm}
\paragraph{RQ1: Does \ours{} generalize better to unseen teammates than baselines? \textcolor{blue}{(Yes.)}}
Fig.~\ref{fig:rotate_main_results}~(left) compares the normalized mean returns for \ours{} and baseline methods on $\Pieval$ across the six tasks.
\ours{} attains the highest mean normalized return on five of the six tasks, with the sole exception of AA, where Minimax Return performs best. 

CoMeDi and FCP are particularly strong baselines.
We attribute CoMeDi's strong performance to the resemblance of its mixed-play objective to our per-state regret objective, which we discuss in App. \ref{app:supp_disc:comedi_mp_discussion}.
Meanwhile, FCP tends to perform especially well with the IPPO policies in $\Pieval$, likely because the IPPO evaluation teammates are in-distribution relative to the teammates constructed by FCP.
Likewise, on CoR and FC, where CoMeDi performs similarly to \ours{}, CoMeDi's returns are driven by high performance with the IPPO evaluation teammates.

Minimax Return's strong performance in AA can be attributed to AA's separated kitchen halves, which give each agent all resources needed for individual task completion. Thus, fully independent agents may achieve high returns, although coordination yields still higher returns (App.~\ref{app:comment_on_aa}).
By contrast, on LBF and FC, where coordination is crucial to obtain positive returns, Minimax Return is the worst-performing baseline.

BRDiv and PAIRED exhibit comparatively worse performance, which may be partially attributed to their teammate generation objectives that resemble per-trajectory regret. 
Furthermore, PAIRED's update structure involves a lockstep training process for the teammate generator, best response, and ego agent, in contrast to \ours{}, which trains a new teammate-BR pair for each iteration. 
This synchronized training may hinder the emergence of diverse conventions crucial for AHT.

Supplemental analyses in the appendix further support the main comparison.
App.~\ref{app:supp:core_views} shows that \ours{} attains the highest interquartile mean, median, and probability of improvement on the five tasks identified above, even under unnormalized or worst-case returns. The per-teammate breakdown in App.~Fig.~\ref{fig:all_radar_charts_combined} further shows that \ours{} has more robust and well-rounded cooperative capabilities compared to baselines, improving more evenly across evaluation teammates.
App.~\ref{app:supp:human_proxy_eval}--\ref{app:supp:n_agent} extend the evaluation to a human proxy study on Overcooked and to $N$-agent and partially observable scenarios, in which \ours{} again matches or outperforms baselines.
App.~\ref{app:supp:teammate_inspection} finds interpretable conventions, no handshake sabotage, largely competent teammates, and reasonable induced regret in the \ours{} populations on LBF and CC. App.~\ref{app:supp:br_quality} finds that the co-trained BRs show some approximation error but no systematic under- or over-estimation.

\vspace{-6pt}
\paragraph{RQ2: Does the ROTATE teammate generation objective mitigate the emergence of teammates that are difficult to learn with? \textcolor{blue}{(Yes.)}} 

\begin{wrapfigure}[23]{r}{0.37\textwidth}
    \vspace{-12pt}
    \centering
    \resizebox{0.55\linewidth}{!}{
    \begin{tabular}{llll}
    & \textit{H} & \textit{T} & \textit{S} \\ \cline{2-4} 
    \multicolumn{1}{l|}{\textit{H}} & \multicolumn{1}{l|}{1}  & \multicolumn{1}{l|}{0}  & \multicolumn{1}{l|}{-1} \\ \cline{2-4} 
    \multicolumn{1}{l|}{\textit{T}} & \multicolumn{1}{l|}{0}  & \multicolumn{1}{l|}{1}  & \multicolumn{1}{l|}{-1} \\ \cline{2-4} 
    \multicolumn{1}{l|}{\textit{S}} & \multicolumn{1}{l|}{-1} & \multicolumn{1}{l|}{-1} & \multicolumn{1}{l|}{-1} \\ \cline{2-4} 
    \end{tabular}
    }\par\vspace{6pt}
    \includegraphics[width=\linewidth]{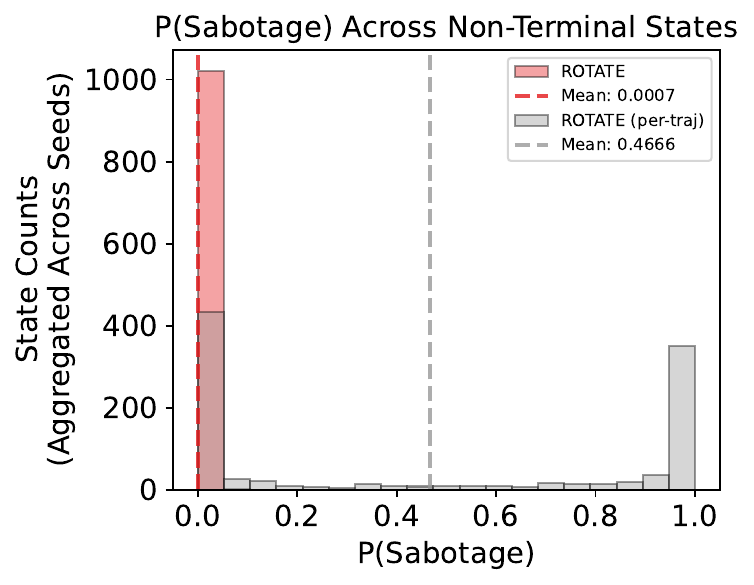}
    \captionof{figure}{\small \textbf{(Top)} Payoff matrix for the sabotage game. \textbf{(Bottom)} Probability of the sabotage action at all non-terminal states in the sabotage game for \ours{} teammates vs those trained with per-trajectory regret. Results are aggregated across three trials.} 
    \label{fig:sabotage_example_histogram}
\end{wrapfigure}

We design a \textit{sabotage game} to investigate whether the proposed objective leads to teammate policies that are easier to learn to interact with compared to the per-trajectory regret objective.
The sabotage game is a fully cooperative, iterated matrix game lasting five timesteps, with payoff matrix in Fig.~\ref{fig:sabotage_example_histogram}~(top). 
Each agent observes a game state consisting of the complete history of joint actions, and chooses among three actions: H, T, and S(abotage). 
H and T lead to two cooperative outcomes, while S leads to a reward of $-1$ and immediate episode termination if selected by either agent. 
Thus, choosing sabotage corresponds to a grim trigger strategy, in which an agent acts to minimize the team's payoffs regardless of the other agent's actions.

We compare \ours{} with a variant in which teammates optimize per-trajectory regret.
To measure the extent to which the learned teammate policies engage in sabotage, the probability of the sabotage action is measured at each of the game's 341 non-terminal states for the last generated teammate policy. 
Fig.~\ref{fig:sabotage_example_histogram}~(bottom) shows that \ours{} generates a teammate policy with a near-zero probability of taking the sabotage action at all non-terminal states, while the per-trajectory regret objective yields P(S) near $1.0$ in over a third of states. Learning useful collaboration strategies with these grim trigger teammates is difficult, since the ego agent rarely receives positive rewards. 
In App.~\ref{app:sabotage_game_extended}, competence weights are swept for both per-state and per-trajectory regret variants on the sabotage game, showing that applying competence bounds to per-state regret mitigates the emergence of sabotage strategies more effectively than applying the same bound to per-trajectory regret.

\vspace{-10pt}
\paragraph{RQ3: What is the influence of the competence weights, $\lambda_1, \lambda_2$, on generalization performance?}
Sec.~\ref{sec:per_state_regret} shows theoretically that maximizing the sum of per-state regret and $V(s_0 | \pi^{-i}, \BR(\pi^{-i}))$ provides a lower bound on the competence of the teammate, where $\lambda_1, \lambda_2$ are the weights of the return from XP and SP states, respectively.
Fig.~\ref{fig:rotate_main_results}~(right) compares the training and evaluation returns of \ours{} as $\lambda_1$ and $\lambda_2$ are varied on CC. 
The training returns are the mean returns of the final \ours{} ego agent when paired with the population of generated teammates, while the evaluation returns are its mean returns with the evaluation teammates. 
Results for all tasks are presented in App.~\ref{app:supp:comp_weights_gen}. 
As $\lambda_1$ and $\lambda_2$ are increased, the training returns increase, which reflects the increased cooperativeness of the generated teammates.
On the other hand, the influence of the competence weights on the evaluation returns---which reflect the ability of the ego agent to generalize its cooperative behavior---is task-dependent. 
On CC, intermediate values of $\lambda_1$ and $\lambda_2$ lead to the best evaluation returns, demonstrating that teammates that are overly cooperative with the ego agent are not ideal for learning generalizable cooperative behavior.
However, on LBF, increased competence weights improve both training and evaluation returns.

\vspace{-10pt}
\section{Discussion and Conclusion}
\label{sec:discussion}
\vspace{-10pt}
This paper reformulates AHT as an open-ended learning problem and introduces ROTATE, a regret-driven algorithm. 
ROTATE iteratively alternates between improving an AHT agent and generating challenging yet cooperative teammates. The teammates are generated by optimizing a regret-based objective designed to discover competent teammates that exploit cooperative vulnerabilities yet remain learnable.
To this end, sabotage is characterized as a property of an objective that induces incompetent teammates, and the proposed objective is shown to bound teammate competence on the states encountered during AHT agent training.
Experiments on an illustrative matrix game demonstrate that the proposed objective mitigates sabotage.  
Extensive evaluations across six benchmark tasks demonstrate that ROTATE significantly improves AHT agent generalization to previously unseen teammates, outperforming baselines from both AHT and UED.

The current work has three limitations that suggest future work. 
First, the theoretical results show that mixing regret and return maximization bounds the competence of generated teammates, but do not prove that bounding competence on teammate-ego interactions is beneficial for ego agent learning.
Second, each open-ended iteration trains the teammate and best response from scratch, which is computationally expensive and may bias regret estimates due to approximation error, motivating amortized training schemes.
Third, this work focuses on the teammate generation phase of open-ended AHT. Future work might explore both ego agent training methods that better handle the nonstationarity induced by open-ended teammate generation, e.g., via continual learning techniques, and alternative methods to sample teammates from the population buffer.
Compared to uniform sampling, a worst-case sampler over the buffer~\citep{buening2023minimax, villin2025minimaxapproachadhoc} could improve worst-case robustness, but coupling it with the already-adversarial \ours{} teammate generator ties the ego agent's training distribution to its own current weaknesses, risking strategic cycling and catastrophic forgetting. 
Finally, previously generated teammates and ego agents could be treated as an empirical meta-game, using meta-strategy solvers to compute both agents' training distributions~\citep{lanctot2017unifiedgametheoretic}.

\begin{ack}
This work has taken place in the Learning Agents Research
Group (LARG) at UT Austin.  LARG research is supported in part by NSF
(FAIN-2019844, NRT-2125858, STC-2535195), ONR (N00014-24-1-2550), ARO
(W911NF-17-2-0181, W911NF-23-2-0004, W911NF-25-1-0065), Lockheed
Martin, Lyda Hill, and Good Systems.  Peter Stone serves as the Chief
Scientist of Sony AI and receives financial compensation for that
role.  The terms of this arrangement have been reviewed and approved
by the University of Texas at Austin in accordance with its policy on
objectivity in research.
\end{ack}

\bibliography{main}
\bibliographystyle{plainnat}
\appendix
\clearpage
\onecolumn
\appendix
\section*{Appendix}
\label{app}
Please find the code for this paper at \url{https://github.com/carolinewang01/rotate/}.

\section{Related Work}
\label{sec:related_work}

\textbf{Training AHT Agents.} 
Most AHT methods follow a two-stage design process, in which a fixed training set of teammate policies is generated, followed by AHT training. Given teammates from the training set, AHT methods~\citep{mirsky2022survey} train an ego agent to model teammates~\citep{ALBRECHT201866} by first modeling important characteristics such as goals, beliefs, policies, and then estimating the best-response policy to these teammates based on the inferred characteristics. Recent AHT methods~\citep{rahman_towards_2021,papoudakis2020liam, zintgraf2021deep, wang2024nagent} use neural networks trained using reinforcement learning~\citep{Schulman2017ProximalPO, mnih2016asynchronous}. To further improve AHT training, several approaches learn a distribution for sampling teammate policies during training based on maximizing the worst-case returns~\citep{villin2025minimaxapproachadhoc} or regret~\citep{erlebach2024raccoon, chaudhary2025improvinghumanaicoordinationadversarial} of trained agents. While few, exceptions to the two-stage process include methods designed for continual AHT~\citep{nekoei_continuous_2021,nekoei_fewshot_2023,yuan2023learning}, methods that co-evolve populations of ego agents and teammates~\citep{xue_maze_2025, yuan2023learning}, self-play based methods, which do not explicitly optimize for diversity~\citep{yan2023an,cornelisse2024humancompat}, and empirical game theoretic methods that optimize for cooperative diversity as a heuristic to induce generalization to unseen teammates~\citep{li_cole_2023}.

\textbf{Teammate Generation for AHT \& ZSC.} Recent work removes the need to predefine teammate policy sets by generating diverse teammates during or before agent training. Other-Play~\citep{hu2020other} creates symmetry-equivalent teammates while training the agent policy, while E3T~\citep{yan2023an} mixes the agent’s current policy with a random policy to encourage diversity. FCP~\citep{strouse_fcp_2022} trains teammates via repeated CMARL runs with different seeds, later improved by methods maximizing information-theoretic diversity objectives such as Jensen-Shannon divergence~\citep{lupu2021trajectory}, mutual information~\citep{lucas2022any}, and entropy~\citep{ijcai2021p66, zhao2023maximum}. 
More recent approaches~\citep{ charakorn2023generating, rahman2024minimum, yuan2023learning} generate teammates that require distinct best-response strategies by maximizing adversarial diversity metrics, similar to \ours{}'s cooperative regret. Unlike \ours{}, these methods (i) maximize regret between generated teammates rather than with the trained agent, (ii) fix the teammate set prior to training, and (iii) evaluate regret only at the initial state. 
Prior work characterizes such sabotage as a behavior of the trained teammate, which identifies that it is interacting in cross-play and then minimizes the team return~\citep{cui2023adversarial, sarkar2023diverse, charakorn2024sabotage}. 
By contrast, \ours{} adopts a more systematic training objective that we demonstrate leads to performance gains.
\citet{lauffer2025rpg} instead define \textit{self-sabotage} as irrational behavior induced by an adversarial optimization problem, and remedy it by constraining the teammate to be a best response to some partner. \ours{} shares this objective-level view but requires competence from the states the ego agent encounters rather than optimality for some partner from the initial state (Sec.~\ref{sec:per_state_regret}).

\textbf{Open-Ended Learning (OEL).} OEL~\citep{langdon2005pfeiffer, taylor2019evolutionary} studies algorithms that continually generate novel tasks to train generally capable agents~\citep{hughes_position_2024,baker_emergent_2019}. Many OEL approaches in RL take the form of unsupervised environment design (UED)~\citep{dennis_emergent_2020}, which improves generalization by designing or sampling new environments with varied initial states. Some methods directly train neural networks to propose environments that induce high regret in the agent~\citep{dennis_emergent_2020}, while others selectively sample curated tasks generated by procedural generators based on criteria such as expected return~\citep{wang2020enhanced}, TD-error~\citep{jiang2021prioritized}, regret~\citep{jiang2021replay}, or learnability~\citep{rutherford2024no}. In competitive MARL, OEL often produces new opponents through self-play~\citep{silver2016mastering,lin_tizero_2023}. For AHT, MACOP~\citep{yuan2023learning} generates novel teammates via an adversarial diversity objective optimized with evolutionary methods and similar to the objectives studied by~\citet{charakorn2023generating} and \citet{rahman2023generating}. Applying this objective only to the initial state can yield teammates that diminish returns in some states, introducing learning challenges for an AHT agent~\citep{charakorn2024sabotage,sarkar2023diverse,lauffer2025rpg}. Prior work characterizes such sabotage as a behavior of the trained teammate, which identifies that it is interacting in cross-play and then minimizes the team return~\citep{cui2023adversarial, sarkar2023diverse, charakorn2024sabotage}. \citet{lauffer2025rpg} instead define \textit{self-sabotage} as irrational behavior induced by an adversarial optimization problem, and remedy it by constraining the teammate to be a best response to some partner. \ours{} shares this objective-level view but requires competence from the states the ego agent encounters rather than optimality for some partner from the initial state (Sec.~\ref{sec:per_state_regret}). 
By contrast, \ours{} adopts a more systematic training objective that we demonstrate leads to performance gains.

\section{Theory}
\label{app:theory}
\subsection{Illustrative Example}
\label{app:theory:examples}
\begin{wraptable}[9]{r}{0.3\textwidth} 
    \centering
    \caption{\small Simple matrix game where maximizing regret minimizes return.}
    \label{tab:max-regret-min-return}
    \scalebox{1.04}{
    \begin{tabular}{llll}
    & \textit{X} & \textit{Y} \\ \cline{2-3} 
    \multicolumn{1}{l|}{\textit{A}} & \multicolumn{1}{l|}{5}  & \multicolumn{1}{l|}{1} \\ \cline{2-3} 
    \multicolumn{1}{l|}{\textit{B}} & \multicolumn{1}{l|}{5}  & \multicolumn{1}{l|}{0} \\ \cline{2-3} 
    \end{tabular}
    }
\end{wraptable}

In this section, we present a simple matrix game to illustrate that a teammate that maximizes regret is not guaranteed to be competent. In fact, it can even be the case that a teammate that maximizes regret also \textit{minimizes} return with all ego agents.

For the payoff matrix in Table~\ref{tab:max-regret-min-return}, the row player is the ego agent and the column player is the teammate agent, so the ego agent has actions $\{A,B\}$ and the teammate has actions $\{X, Y\}$.

If the teammate maximizes cooperative regret
\begin{equation*}
    CR(\piego, \pi^{-i}) = V(\pi^{-i}, BR(\pi^{-i})) - V(\pi^{-i}, \piego),\hspace{1.5in}
\end{equation*}
then the teammate will actually minimize return with all ego agents.

To see this, suppose $\piego$ plays $A$ with probability $p$ and $B$ with probability $1-p$. If the teammate plays $X$, then
\begin{equation}
CR(\piego, X) = V(X, BR(X)) - V(X, \piego) = 5 - (p*5 + (1-p)*5) = 5-5 = 0,
\end{equation}
whereas if the teammate plays $Y$, then
\begin{equation}
CR(\piego, Y) = V(Y, BR(Y)) - V(Y, \piego) = 1 - (p*1 + (1-p)*0) = 1-p.
\end{equation}
Thus $CR(\piego, X) \leq CR(\piego, Y)$ for all $\piego$, so the teammate that plays $Y$ maximizes regret with all ego agents. However, playing $Y$ also clearly minimizes return with all ego agents.

\subsection{Proofs}
\label{app:theory:proofs}

This section proves the two competence bounds stated in Section~\ref{sec:method_teamgen}. Theorem~\ref{thm:traj_competence} bounds the competence of a teammate maximizing competence-bounded per-trajectory regret (Eq.~\ref{eq:traj_regret_obj_lambda}), and Theorem~\ref{thm:competence2} bounds the competence of a teammate maximizing the \ours{} objective (Eq.~\ref{eq:constrained_state_based_regret_obj}). Each theorem is preceded by a proposition relating the regret and return components of its objective.

Throughout, $\piego$ is fixed. For any teammate policy $\pi'$, we write $V^{\text{SP}}_{\pi'}(s_0) := V\left(s_0|\pi', \BR(\pi')\right)$ for the return of $\pi'$ with its best response and $V^{\text{XP}}_{\pi'}(s_0) := V\left(s_0|\pi', \piego\right)$ for its return with the ego agent. As mentioned in Sec.~\ref{sec:per_state_regret}, $\BR(\pi')$ denotes the best response to $\pi'$ at every state, so $V^{\text{SP}}_{\pi'}(s_0)$ is the best return that some partner can achieve with $\pi'$ from $s_0$, whether or not $s_0$ is reachable under self-play from $p_0$.

\subsubsection{Competence-Bounded Per-Trajectory Regret}
\label{app:theory:proofs:traj}

Recall the competence-bounded per-trajectory objective of Sec.~\ref{sec:bounding_competence},
\begin{equation}
\trajobjbody. \tag{\ref{eq:traj_regret_obj_lambda}}
\end{equation}
\begin{prop}\label{prop:traj}
Let $\lambda \in \R^{> 0}$. If the teammate $\pi^{-i}$ maximizes Eq.~\ref{eq:traj_regret_obj_lambda}, then
\begin{equation*}
\E_{s_0\sim p_0}\left[CR\left(\piego, \pi^{-i}, s_0\right)\right] \geq \lambda\left(R^{\max}_{p_0} - \E_{s_0\sim p_0}\left[V^{\text{SP}}_{\pi^{-i}}(s_0)\right]\right),
\end{equation*}
where $R^{\max}_{p_0} = \max_{\pi'} \E_{s_0 \sim p_0}\left[V^{\text{SP}}_{\pi'}(s_0)\right]$.
\end{prop}

\begin{proof}
Because cooperative regret is always non-negative, we get
\begin{align*}
\max_{\pi'}\E_{s_0\sim p_0}\left[CR\left(\piego, \pi', s_0\right) + \lambda V^{\text{SP}}_{\pi'}(s_0)\right]
&\geq \max_{\pi'}\E_{s_0\sim p_0}\left[\lambda V^{\text{SP}}_{\pi'}(s_0)\right] = \lambda R^{\max}_{p_0}.
\end{align*}
Because $\pi^{-i}$ maximizes the left-hand side,
\begin{equation*}
\E_{s_0\sim p_0}\left[CR\left(\piego, \pi^{-i}, s_0\right)\right] + \lambda\E_{s_0\sim p_0}\left[V^{\text{SP}}_{\pi^{-i}}(s_0)\right] \geq \lambda R^{\max}_{p_0},
\end{equation*}
and rearranging shows that the proposition holds.
\end{proof}

This proposition states that cooperative regret, which is the gap between the SP value and the XP value, must be larger than $\lambda$ times the gap between the best possible value and the SP value. For larger $\lambda$, in order for the inequality to hold, either the cooperative regret has to be larger or the teammate must be more competent (so as to achieve returns closer to $R^\text{max}_{p_0}$). We leverage this proposition along with the recognition that cooperative regret may be bounded above to prove the following theorem:

\trajcompetencetheorem*

\begin{proof}
By Proposition~\ref{prop:traj} and the definition of cooperative regret,
\begin{align*}
\E_{s_0\sim p_0}\left[V^{\text{SP}}_{\pi^{-i}}(s_0)\right]
&\geq R^{\max}_{p_0} - \frac{1}{\lambda}\E_{s_0\sim p_0}\left[V^{\text{SP}}_{\pi^{-i}}(s_0) - V^{\text{XP}}_{\pi^{-i}}(s_0)\right] \\
&\geq R^{\max}_{p_0} - \frac{1}{\lambda}\E_{s_0\sim p_0}\left[V^{\text{SP}}_{\pi^{-i}}(s_0)\right] + \frac{1}{\lambda}R^{\min}_{p_0},
\end{align*}
where the second inequality holds because $\pi^{-i}$ is one of the teammates over which the minimum defining $R^{\min}_{p_0}$ is taken.
Collecting terms gives $\frac{\lambda+1}{\lambda}\E_{s_0\sim p_0}\left[V^{\text{SP}}_{\pi^{-i}}(s_0)\right] \geq R^{\max}_{p_0} + \frac{1}{\lambda}R^{\min}_{p_0}$, and multiplying by $\frac{\lambda}{\lambda+1}$ yields the stated bound.

For the second claim, rewrite the right-hand side of the bound as
\begin{equation*}
\frac{\lambda}{\lambda+1}R^{\max}_{p_0} + \frac{1}{\lambda+1}R^{\min}_{p_0} = R^{\min}_{p_0} + \frac{\lambda}{\lambda+1}\left(R^{\max}_{p_0} - R^{\min}_{p_0}\right).
\end{equation*}
Because $K < R^{\max}_{p_0}$, this quantity is at least $K$ if and only if
\begin{align*}
\frac{\lambda}{\lambda+1} &\geq \frac{K - R^{\min}_{p_0}}{R^{\max}_{p_0} - R^{\min}_{p_0}} \\
\iff \lambda\left(R^{\max}_{p_0} - R^{\min}_{p_0}\right) &\geq (\lambda+1)\left(K - R^{\min}_{p_0}\right) \\
\iff \lambda\left(R^{\max}_{p_0} - K\right) &\geq K - R^{\min}_{p_0},
\end{align*}
and dividing by $R^{\max}_{p_0} - K > 0$ gives $\lambda \geq \frac{K - R^{\min}_{p_0}}{R^{\max}_{p_0} - K}$. Therefore, any such $\lambda$ yields $\E_{s_0\sim p_0}\left[V^{\text{SP}}_{\pi^{-i}}(s_0)\right] \geq K$, so $\pi^{-i} \in \Pi^{-i}_K(p_0)$.
\end{proof}

\subsubsection{Competence-Bounded Per-State Regret}
\label{app:theory:proofs:state}

The $\ours$ teammate generation objective (Sec.~\ref{sec:per_state_regret}, Eq. \ref{eq:constrained_state_based_regret_obj}) has two components, and is restated below for convenience. One component is the cooperative regret maximization over $U(\pi^{-i})$. The other component is the BR-teammate return maximization over $\pxp(\pi^{-i})$ and $\psp(\pi^{-i})$.

\begin{align*}
\stateobjbody. \tag{\ref{eq:constrained_state_based_regret_obj}}
\end{align*}

Following the same strategy as for the per-trajectory regret results, we first give a proposition stating the relationship between these two components of the teammate generation objective to prove the theorem in Section~\ref{sec:per_state_regret}.

For any teammate $\pi'$, the XP and SP state visitation distributions are $\pxp(\pi') := d(\pi', \piego; p_0)$ and $\psp(\pi') := d(\pi', \BR(\pi'); p_0)$. Let $\Lambda := \lambda_1 + \lambda_2 > 0$, and define the three mixtures
\begin{align*}
U(\pi') &:= \tfrac{1}{2}\pxp(\pi') + \tfrac{1}{2}\psp(\pi'), \\
\mu(\pi') &:= \tfrac{\lambda_1}{\Lambda}\pxp(\pi') + \tfrac{\lambda_2}{\Lambda}\psp(\pi'), \\
\nu(\pi') &:= \tfrac{\lambda_1+1}{\Lambda+2}\pxp(\pi') + \tfrac{\lambda_2+1}{\Lambda+2}\psp(\pi'),
\end{align*}
so that $U(\pi^{-i})$ is the distribution over which regret is maximized in Eq.~\ref{eq:constrained_state_based_regret_obj}, $\mu(\pi')$ is the $\lambda$-weighted mixture obtained by combining the two return terms of Eq.~\ref{eq:constrained_state_based_regret_obj} into a single expectation, and $\nu(\pi')$ is the mixture appearing in Theorem~\ref{thm:competence2}.

\begin{prop}\label{prop:lambda-1-2}
Let $\lambda_1,\lambda_2 \in \R^{\geq 0}$ with $\Lambda > 0$. If the teammate $\pi^{-i}$ maximizes Eq.~\ref{eq:constrained_state_based_regret_obj}, then 
\begin{equation*}
\E _{s_0\sim U(\pi^{-i})}\left[CR\left(\piego,\pi^{-i}, s_0\right)\right] \geq \frac{\Lambda}{2}\left(R^{\max}_{\mu} - \E_{s_0 \sim \mu(\pi^{-i})} \left[V^{\text{SP}}_{\pi^{-i}}(s_0)\right]\right),
\end{equation*}
where 
\begin{equation*}
R^{\max}_{\mu} = \max_{\pi'} \mathbb{E}_{s_0 \sim \mu(\pi')}\left[V^{\text{SP}}_{\pi'}(s_0)\right]
\end{equation*}
denotes the maximum $\mu$-weighted return that any teammate achieves with its best response on the states that teammate induces.
\end{prop}

\begin{proof}
First, for any teammate $\pi'$ and function $F(s_0)$, 
\begin{align*}
\MoveEqLeft \lambda_1\E _{s_0\sim\pxp(\pi')}\left[F(s_0) \right] + \lambda_2\E _{s_0\sim\psp(\pi')}\left[F(s_0) \right] \\
&= \Lambda\left(\frac{\lambda_1}{\Lambda}\E _{s_0\sim\pxp(\pi')}\left[F(s_0) \right] + \frac{\lambda_2}{\Lambda}\E _{s_0\sim\psp(\pi')}\left[F(s_0) \right]\right) \\
&= \Lambda \, \E_{s_0 \sim \mu(\pi')} \left[F(s_0)\right].
\end{align*}

Because cooperative regret is always non-negative, we may obtain
\begin{align*}
\MoveEqLeft \max_{\pi'}\left[2\E _{s_0\sim U(\pi')}\left[CR\left( \piego, \pi', s_0 \right)\right] + \lambda_1\E _{s_0\sim\pxp(\pi')}\left[V^{\text{SP}}_{\pi'}(s_0) \right] + \lambda_2\E _{s_0\sim\psp(\pi')}\left[V^{\text{SP}}_{\pi'}(s_0) \right]\right] \\
&\geq \max_{\pi'}\left[\lambda_1\E _{s_0\sim\pxp(\pi')}\left[V^{\text{SP}}_{\pi'}(s_0) \right] + \lambda_2\E _{s_0\sim\psp(\pi')}\left[V^{\text{SP}}_{\pi'}(s_0) \right]\right]\\
&= \max_{\pi'}\left[\Lambda \, \E_{s_0 \sim \mu(\pi')} \left[V^{\text{SP}}_{\pi'}(s_0)\right]\right]\\
&= \Lambda R^{\max}_{\mu},
\end{align*}
where the last line is the definition of $R^{\max}_{\mu}$.

Because $\pi^{-i}$ maximizes this equation, we get
\begin{align*}
\MoveEqLeft 2\E _{s_0\sim U(\pi^{-i})}\left[CR\left( \piego, \pi^{-i}, s_0 \right)\right]\\
&\geq \Lambda R^{\max}_{\mu} - \lambda_1\E_{s_0\sim\pxp(\pi^{-i})}\left[V^{\text{SP}}_{\pi^{-i}}(s_0) \right] - \lambda_2\E_{s_0\sim\psp(\pi^{-i})}\left[V^{\text{SP}}_{\pi^{-i}}(s_0) \right]\\
&=\Lambda\left(R^{\max}_{\mu} - \E_{s_0 \sim \mu(\pi^{-i})} \left[V^{\text{SP}}_{\pi^{-i}}(s_0)\right]\right).
\end{align*}
Dividing by 2 shows that the proposition holds.
\end{proof}

This proposition states that, for any teammate that maximizes the ROTATE teammate generation objective, it must be true that the cooperative regret on $U(\pi^{-i})$ is at least as large as $\frac{\Lambda}{2}$ times the gap between the overall best possible $\mu$-weighted return $R^{\max}_{\mu}$, and the best possible return with $\pi^{-i}$ on the state distribution $\mu(\pi^{-i})$.
The following theorem relies on the proposition to show that the guaranteed minimum competence of the teammate increases as $\lambda_1$ and $\lambda_2$ increase.

\competencetheorem*

\begin{proof}

By Proposition \ref{prop:lambda-1-2},
\begin{align*}
\MoveEqLeft \E_{s_0\sim \mu(\pi^{-i})}\left[V^{\text{SP}}_{\pi^{-i}}(s_0)\right]\\
&\geq R^{\max}_{\mu} - \frac{2}{\Lambda}\E_{s_0\sim U(\pi^{-i})}\left[CR\left(\piego, \pi^{-i}, s_0\right)\right] \\
&= R^{\max}_{\mu} - \frac{2}{\Lambda}\E_{s_0\sim U(\pi^{-i})}\left[V^{\text{SP}}_{\pi^{-i}}(s_0) - V^{\text{XP}}_{\pi^{-i}}(s_0)\right] \\
&= R^{\max}_{\mu} - \frac{2}{\Lambda}\E_{s_0\sim U(\pi^{-i})}\left[V^{\text{SP}}_{\pi^{-i}}(s_0)\right] + \frac{2}{\Lambda}\E_{s_0\sim U(\pi^{-i})}\left[V^{\text{XP}}_{\pi^{-i}}(s_0)\right]\\
&\geq R^{\max}_{\mu} - \frac{2}{\Lambda}\E_{s_0\sim U(\pi^{-i})}\left[V^{\text{SP}}_{\pi^{-i}}(s_0)\right] + \frac{2}{\Lambda}R^{\min}_{U},
\end{align*}
where the last inequality holds because $\pi^{-i}$ is one of the teammates over which the minimum defining $R^{\min}_{U}$ is taken.

This implies that:
\begin{align*}
 \E_{s_0\sim \mu(\pi^{-i})}\left[V^{\text{SP}}_{\pi^{-i}}(s_0)\right] + \frac{2}{\Lambda}\E_{s_0\sim U(\pi^{-i})}\left[V^{\text{SP}}_{\pi^{-i}}(s_0)\right]
 \geq R^{\max}_{\mu} + \frac{2}{\Lambda}R^{\min}_{U}.
\end{align*}

Multiply both sides with $\dfrac{\Lambda}{\Lambda+2}$ to obtain:
\begin{align*}
\MoveEqLeft \dfrac{\Lambda}{\Lambda+2}\E_{s_0\sim \mu(\pi^{-i})}\left[V^{\text{SP}}_{\pi^{-i}}(s_0)\right] + \frac{2}{\Lambda+2}\E_{s_0\sim U(\pi^{-i})}\left[V^{\text{SP}}_{\pi^{-i}}(s_0)\right]\\
&\geq \dfrac{\Lambda}{\Lambda+2}R^{\max}_{\mu} + \frac{2}{\Lambda+2}R^{\min}_{U}.
\end{align*}

Then 
\begin{align*}
\MoveEqLeft \E_{s_0\sim \frac{\lambda_1}{\Lambda+2}\pxp(\pi^{-i}) + \frac{\lambda_2}{\Lambda+2}\psp(\pi^{-i}) + \frac{2}{\Lambda+2}U(\pi^{-i})}\left[V^{\text{SP}}_{\pi^{-i}}(s_0)\right]\\
&\geq \dfrac{\Lambda}{\Lambda+2}R^{\max}_{\mu} + \frac{2}{\Lambda+2}R^{\min}_{U}.
\end{align*}

Combining the distributions, we can rewrite the above inequality as
\begin{align*}
\E_{s_0\sim \nu(\pi^{-i})}\left[V^{\text{SP}}_{\pi^{-i}}(s_0)\right]
\geq \dfrac{\Lambda}{\Lambda+2}R^{\max}_{\mu} + \frac{2}{\Lambda+2}R^{\min}_{U}.
\end{align*}
\end{proof}

Theorem~\ref{thm:competence2} certifies competence with respect to $\nu(\pi^{-i})$, which places weight $\frac{\lambda_1+1}{\Lambda+2}$ on the XP state distribution $\pxp(\pi^{-i})$. By contrast, Theorem~\ref{thm:traj_competence} certifies competence only with respect to $p_0$, so a teammate maximizing Eq.~\ref{eq:traj_regret_obj_lambda} can become incompetent at states visited only with $\piego$ without weakening its guarantee, which is the incentive to sabotage described in Sec.~\ref{sec:per_state_regret}. Under the \ours{} objective, becoming incompetent at such states lowers $V^{\text{SP}}_{\pi^{-i}}$ on the XP state distribution, and is therefore penalized by the return terms of Eq.~\ref{eq:constrained_state_based_regret_obj}.

\section{Supplemental Results}
\label{app:supp_results}

This section provides additional experiments examining the \ours{} teammate generation objective and population buffer in further detail. It also provides supplemental material supporting the main results. Specifically:

\begin{itemize}
    \item We present additional analyses of the main result: unnormalized returns, per-task interval estimates, probability of improvement, worst-case returns, and a breakdown by individual evaluation teammate (App.~\ref{app:supp:core_views}).
    \item We provide a human proxy evaluation of \ours{} and all baselines on the Overcooked tasks (App.~\ref{app:supp:human_proxy_eval}).
    \item We provide evaluations of \ours{} when extended to a partially observable variant of LBF (App.~\ref{app:supp:partial_obs}) and an N-agent variant of LBF (App.~\ref{app:supp:n_agent}).
    \item We analyze the impact of competence weights on sabotage rates for per-state and per-trajectory regret (App.~\ref{app:sabotage_game_extended}).
    \item We analyze the generalization performance of \ours{} as competence weights are varied (App.~\ref{app:supp:comp_weights_gen}).
    \item We compare \ours{} to variants that use trajectory regret modulated by return, and CoMeDi-style mixed-play return maximization (App.~\ref{app:supp:teammate_gen_ablations}).
    \item We ablate the population buffer and examine whether the population generated by \ours{} is useful for training an independent ego agent (App.~\ref{app:supp:ind_ego_agent_pop}).
    \item We analyze the teammates generated by \ours{} for their conventions, presence of handshake conventions, competence, and the regret they induce on the ego agent (App.~\ref{app:supp:teammate_inspection}). We also assess the quality of the co-trained best responses (App.~\ref{app:supp:br_quality}).
    \item We present the learning curves for all variants of \ours{} tested in this paper (App.~\ref{app:supp:learning_curves}).
\end{itemize}

\subsection{Additional Analyses of the Main Result}
\label{app:supp:core_views}

This subsection presents five additional views of Fig.~\ref{fig:rotate_main_results}~(left), which compares the mean normalized evaluation return of \ours{} to all baselines on all tasks. Here, the same comparison is performed with unnormalized return units, aggregated by various summary statistics (IQM, median, probability of improvement), by worst case returns, and broken down by individual evaluation teammate.

\begin{figure}[H] %
    \centering
    \begin{subfigure}[h]{0.48\linewidth}
        \centering
        \includegraphics[width=\linewidth]{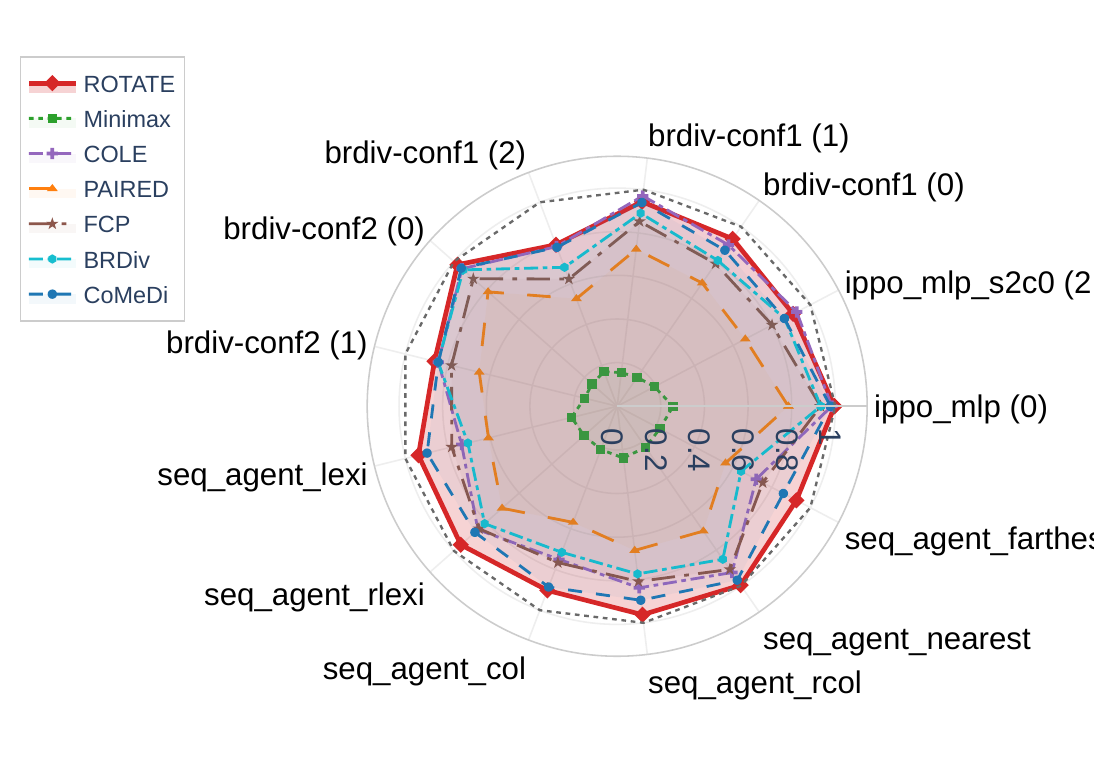}
        \caption{LBF.}
        \label{fig:lbf_radar_sub}
    \end{subfigure}
    \hfill %
    \begin{subfigure}[h]{0.48\linewidth}
        \centering
        \includegraphics[width=\linewidth]{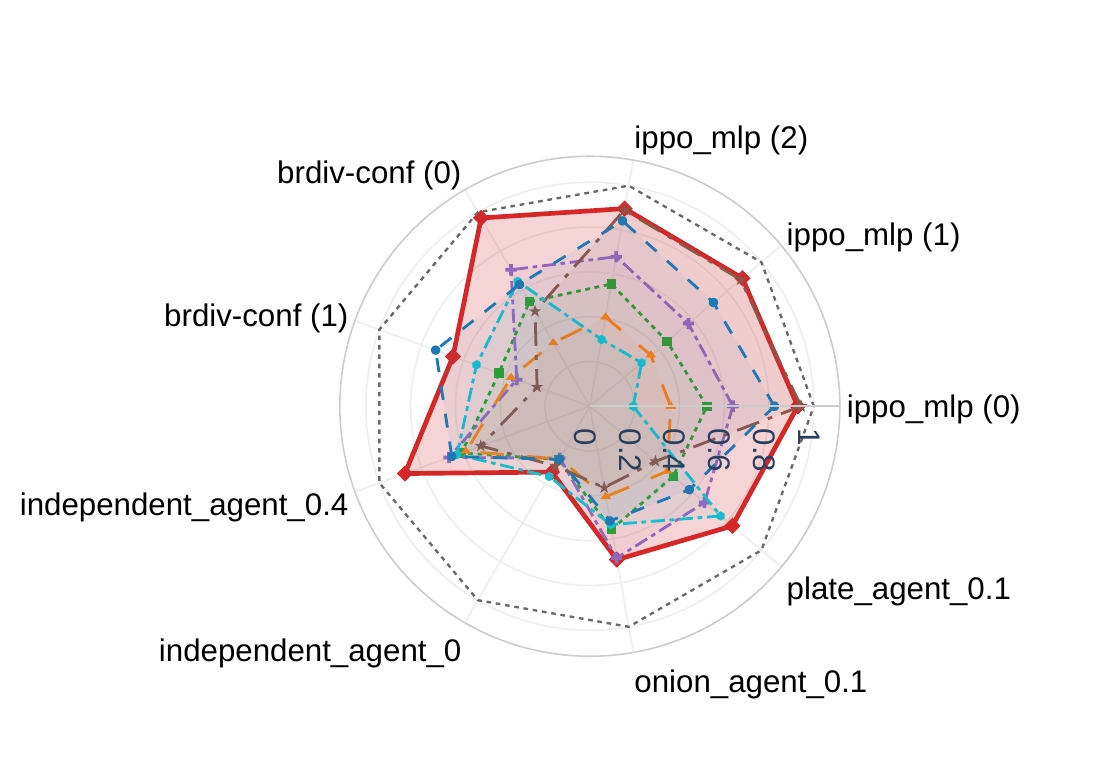}
        \caption{Cramped Room.}
        \label{fig:cramped_room_radar_sub}
    \end{subfigure}

    \begin{subfigure}[h]{0.48\linewidth}
        \centering
        \includegraphics[width=\linewidth]{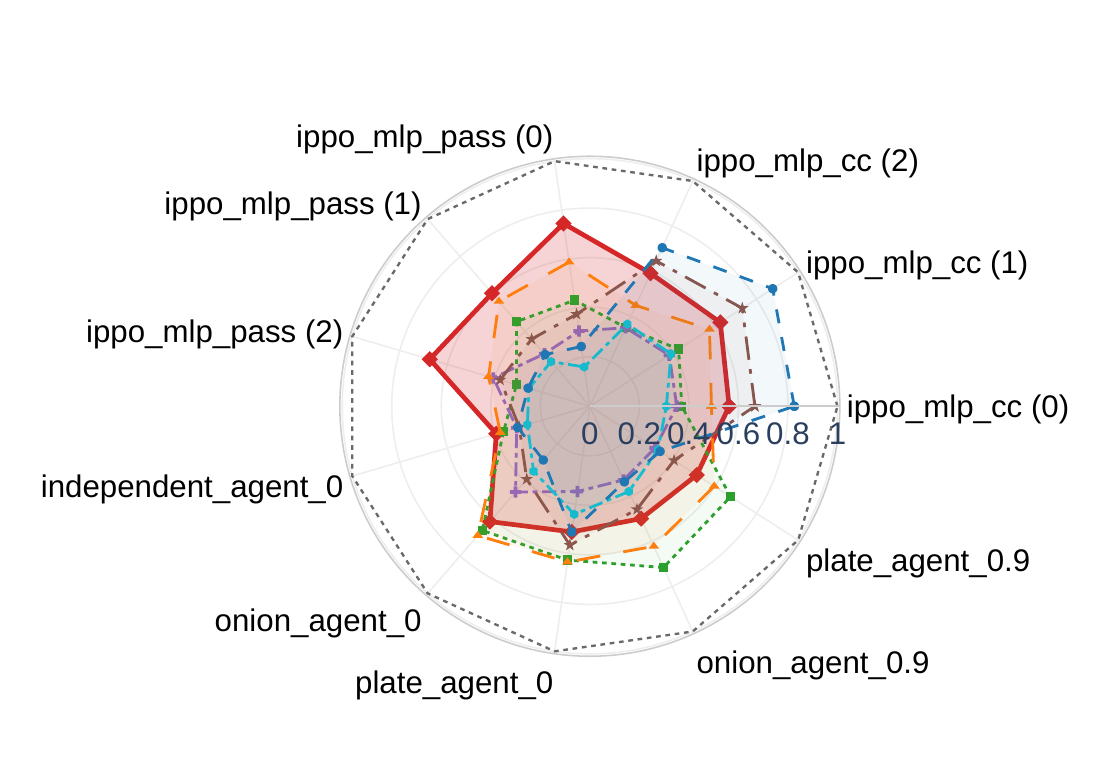}
        \caption{Counter Circuit.}
        \label{fig:counter_circuit_radar_sub}
    \end{subfigure}
    \hfill
    \begin{subfigure}[h]{0.48\linewidth}
        \centering
        \includegraphics[width=\linewidth]{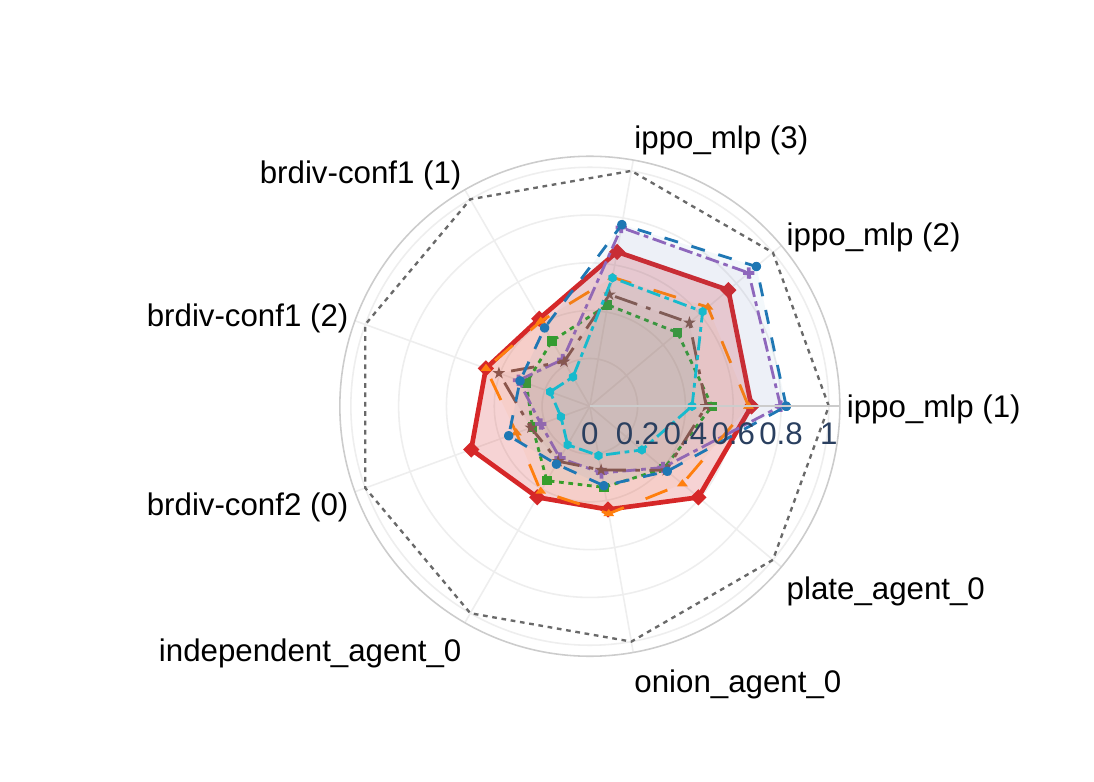}
        \caption{Coordination Ring.}
        \label{fig:coord_ring_radar_sub}
    \end{subfigure}

    \begin{subfigure}[h]{0.48\linewidth}
        \centering
        \includegraphics[width=\linewidth]{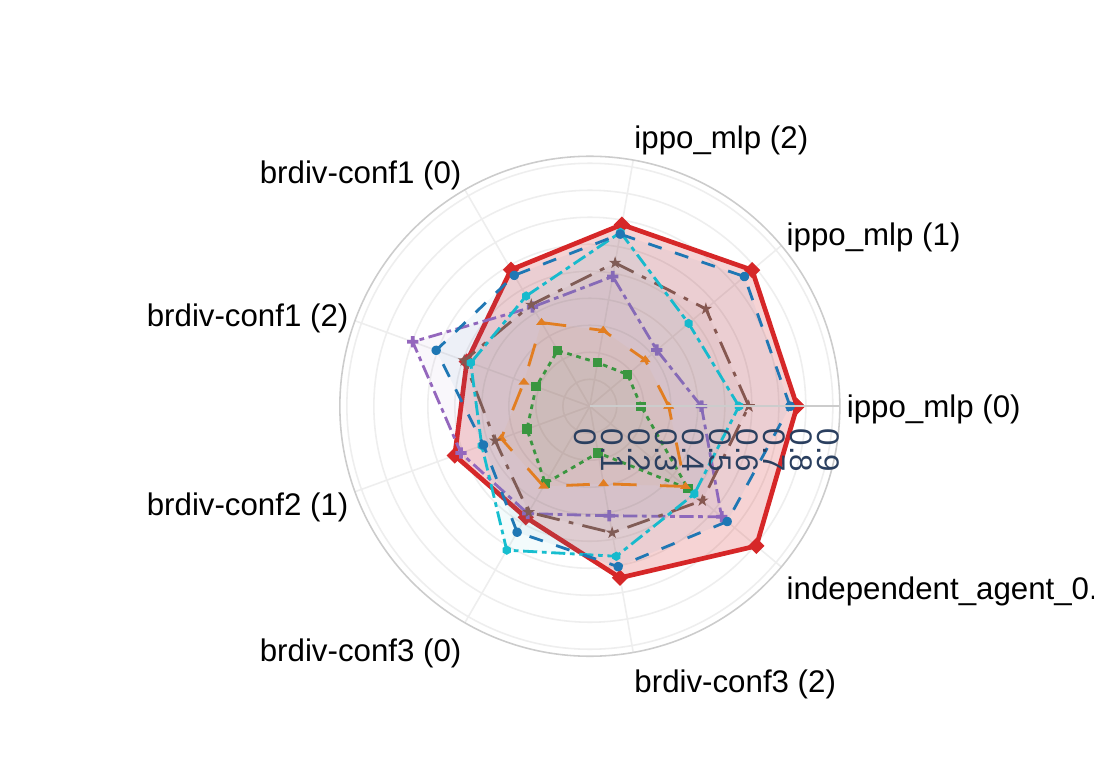}
        \caption{Forced Coordination.} %
        \label{fig:forced_coord_radar_sub}
    \end{subfigure}
    \hfill
    \begin{subfigure}[h]{0.48\linewidth}
        \centering
        \includegraphics[width=\linewidth]{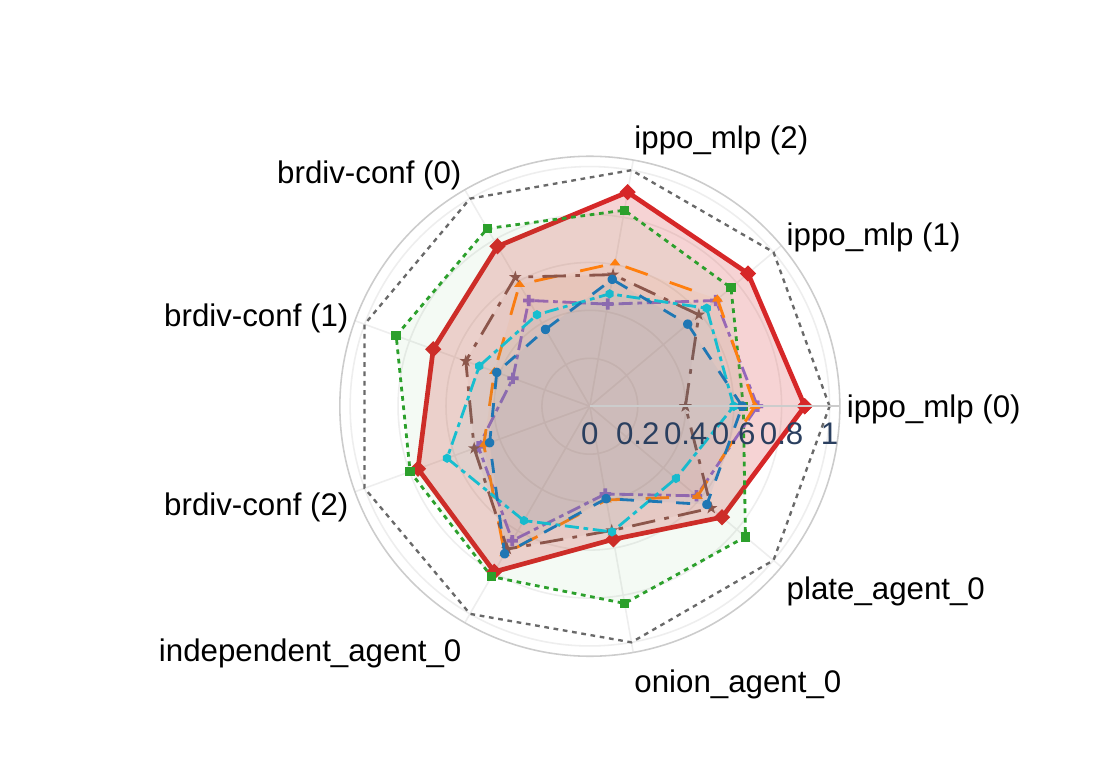}
        \caption{Asymmetric Advantages.} %
        \label{fig:asymm_adv_radar_sub}
    \end{subfigure}
    \caption{Normalized mean returns of \ours{} and all baselines across all tasks, broken down by evaluation teammate in $\Pieval$. Legend shown for LBF applies for all plots.}
    \label{fig:all_radar_charts_combined}
\end{figure}

\paragraph{Unnormalized returns.}
\label{app:supp:unnormalized}
Table~\ref{tab:unnormalized} reports the unnormalized evaluation returns corresponding to Fig.~\ref{fig:rotate_main_results}~(left).
Under unnormalized evaluation returns, \ours{} remains the best method on five of six tasks with the exception of AA, where Minimax Return attains a higher return. 
The per-teammate upper bounds used for normalization are reported in \cref{tab:normalization_bounds}, and per-teammate performance is shown in Fig.~\ref{fig:all_radar_charts_combined}.

\begin{table}[h]
    \renewcommand{\arraystretch}{1.3}
    \captionsetup{skip=10pt}
    \centering
    \small
    \caption{Unnormalized mean evaluation returns with 95\% stratified bootstrapped CI's. For LBF, the percent eaten is shown, while for Overcooked, the base return is shown.}
    \label{tab:unnormalized}
    \resizebox{\linewidth}{!}{%
    \begin{tabular}{lcccccc}
    \hline
        ~ & LBF & CR & AA & CC & CoR & FC \\ \hline
        \ours{} & \textbf{89.5} [88.9, 90.1] & \textbf{175.1} [173.6, 176.7] & 251.7 [248.2, 255.1] & \textbf{83.2} [82.2, 84.1] & \textbf{115.5} [114.2, 116.8] & \textbf{107.0} [105.5, 108.5] \\
        Minimax Return & 19.0 [18.4, 19.7] & 107.9 [106.7, 109.1] & \textbf{258.6} [256.6, 260.5] & 64.6 [63.6, 65.6] & 77.9 [76.6, 79.2] & 36.3 [35.5, 37.0] \\
        COLE & 82.5 [81.7, 83.4] & 127.1 [125.6, 128.6] & 172.7 [169.4, 176.0] & 50.0 [48.9, 51.1] & 101.6 [100.5, 102.8] & 77.7 [76.1, 79.3] \\
        PAIRED & 63.8 [62.8, 64.8] & 85.8 [84.6, 87.1] & 186.2 [183.1, 189.3] & 75.3 [74.3, 76.2] & 101.6 [100.4, 102.9] & 50.5 [49.6, 51.4] \\
        FCP & 77.9 [77.0, 78.8] & 129.0 [127.4, 130.5] & 181.6 [178.9, 184.2] & 72.1 [70.8, 73.4] & 78.3 [77.0, 79.5] & 83.1 [81.5, 84.8] \\
        BRDiv & 77.2 [76.2, 78.1] & 100.3 [98.3, 102.5] & 175.0 [172.5, 177.7] & 44.8 [43.8, 45.7] & 66.8 [65.8, 67.9] & 87.2 [85.4, 89.0] \\
        CoMeDi & 85.8 [85.1, 86.6] & 145.2 [143.1, 147.4] & 168.3 [164.9, 171.7] & 71.7 [70.5, 72.8] & 111.6 [110.1, 113.1] & 103.5 [101.8, 105.2] \\ \hline
    \end{tabular}
    }
\end{table}

\begin{figure}[H]
    \centering
    \includegraphics[width=0.85\linewidth]{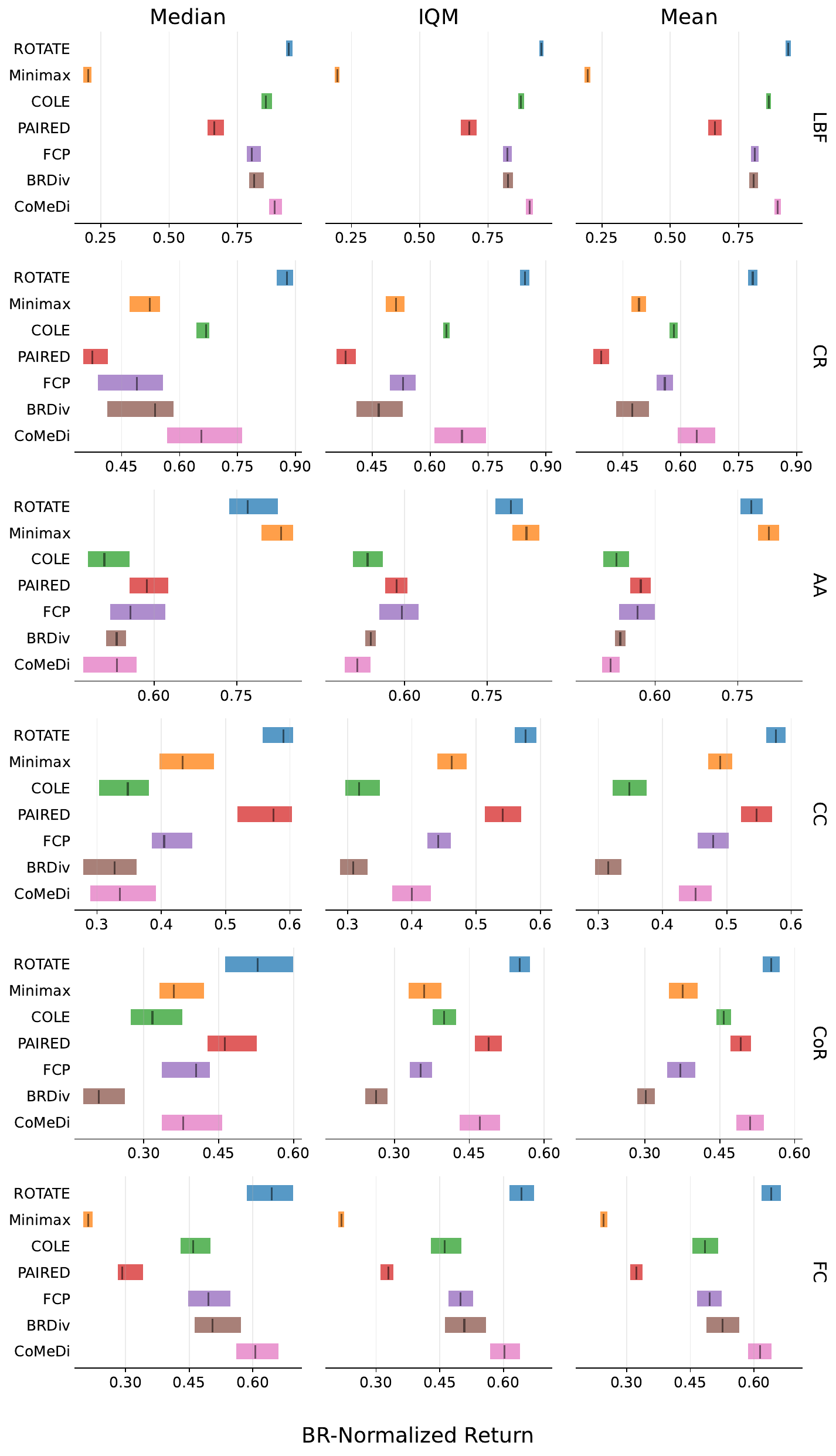}
    \caption{Median, IQM and mean normalized return per task, with 95\% stratified bootstrap CI's over six trials.}
    \label{fig:per_task_interval_estimates}
\end{figure}

\paragraph{Additional statistics.} 
 The main paper reports mean normalized returns with 95\% stratified bootstrapped CI's. Following the recommendations of \citet{agarwal_deep_2021}, Fig.~\ref{fig:per_task_interval_estimates}  reports the median, interquartile mean (IQM), and mean normalized return per task with 95\% stratified bootstrap CI's. 
 \ours{} attains the highest value of all three statistics on five of the six tasks, with the sole exception of AA. The one near-tie is the CC median, on which \ours{} and PAIRED are indistinguishable (0.635 each).

\begin{figure}[H]
    \centering
    \includegraphics[width=\linewidth]{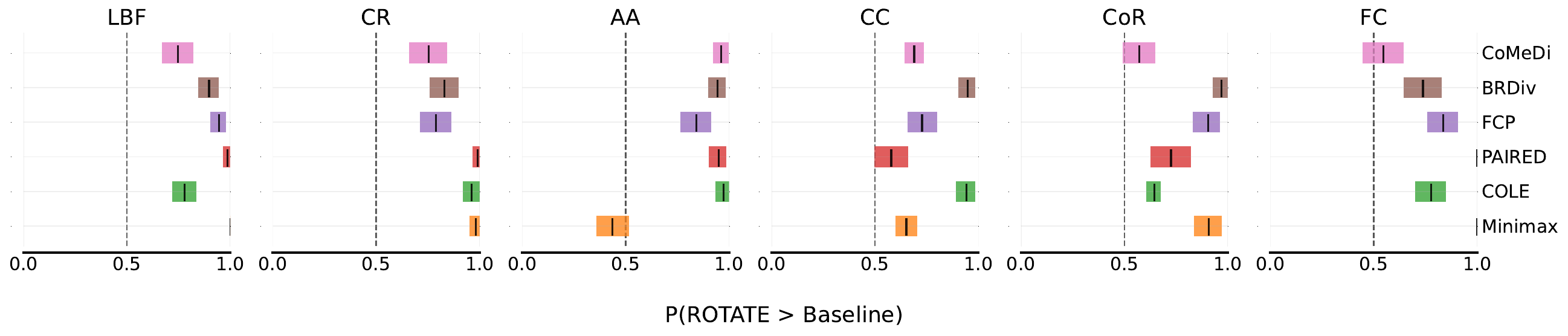}
    \caption{Probability that \ours{} improves over each baseline for each task, with 95\% stratified bootstrap confidence intervals over six trials. The dashed line marks $P = 0.5$.}
    \label{fig:per_task_prob_improvement}
\end{figure}

Fig.~\ref{fig:per_task_prob_improvement} also reports the probability that, for each task and baseline, a randomly selected \ours{} run outperforms a randomly selected baseline run on a randomly selected evaluation teammate. 
\ours{} is favored ($P > 0.5$) in 35 of the 36 (task, baseline) pairs; the exception is Minimax Return on AA ($0.44$ [$0.36$, $0.52$]). 
Of the 35 favorable comparisons, all but three have confidence intervals that exclude $0.5$: PAIRED on CC, CoMeDi on CoR, and CoMeDi on FC.

\paragraph{Worst-case returns.} Because the goal of \ours{} is robustness to $\Pieval$, all methods are also compared on the \textit{minimum} normalized return achieved against any member of $\Pieval$. Table~\ref{tab:worstcase} shows that \ours{} attains the highest worst-case return on five of six tasks with the sole exception of AA, where \ours{} is the second-best method. 
This result indicates that \ours{} demonstrates better-rounded cooperation with all evaluation teammates by avoiding performance collapse against any particular teammate.

\begin{table}[h]
    \renewcommand{\arraystretch}{1.3}
    \captionsetup{skip=10pt}
    \centering
    \small
    \caption{Worst-case (minimum) normalized return achieved by a method with any evaluation teammate. 95\% stratified bootstrap CI's over six trials displayed.}
    \label{tab:worstcase}
    \resizebox{\linewidth}{!}{%
    \begin{tabular}{lcccccc}
    \hline
        ~ & LBF & CR & AA & CC & CoR & FC \\ \hline
        \ours{} & \textbf{0.79} [0.71, 0.87] & \textbf{0.34} [0.32, 0.36] & 0.56 [0.51, 0.63] & \textbf{0.39} [0.36, 0.43] & \textbf{0.42} [0.39, 0.45] & \textbf{0.47} [0.44, 0.51] \\
        Minimax Return & 0.15 [0.11, 0.20] & 0.28 [0.25, 0.30] & \textbf{0.64} [0.57, 0.70] & 0.31 [0.27, 0.35] & 0.26 [0.22, 0.30] & 0.16 [0.14, 0.19] \\
        COLE & 0.72 [0.66, 0.77] & 0.27 [0.26, 0.28] & 0.34 [0.28, 0.40] & 0.28 [0.25, 0.32] & 0.22 [0.19, 0.25] & 0.32 [0.23, 0.44] \\
        PAIRED & 0.53 [0.45, 0.64] & 0.27 [0.24, 0.30] & 0.40 [0.34, 0.46] & 0.38 [0.33, 0.43] & 0.33 [0.26, 0.39] & 0.26 [0.21, 0.31] \\
        FCP & 0.62 [0.55, 0.68] & 0.25 [0.19, 0.31] & 0.40 [0.30, 0.51] & 0.30 [0.28, 0.32] & 0.22 [0.19, 0.24] & 0.37 [0.34, 0.39] \\
        BRDiv & 0.64 [0.59, 0.70] & 0.19 [0.13, 0.27] & 0.44 [0.41, 0.47] & 0.16 [0.14, 0.18] & 0.13 [0.10, 0.16] & 0.43 [0.36, 0.51] \\
        CoMeDi & 0.78 [0.73, 0.83] & 0.27 [0.22, 0.35] & 0.37 [0.35, 0.39] & 0.24 [0.18, 0.32] & 0.28 [0.25, 0.33] & 0.42 [0.40, 0.44] \\
    \hline
    \end{tabular}%
    }
\end{table}

\paragraph{Per evaluation teammate performance breakdown.} We break down the performance of \ours{} and all baseline methods by individual evaluation teammate policies as radar charts in Fig. \ref{fig:all_radar_charts_combined}. 
The radar charts show that \ours{} achieves higher performance across a larger number and variety of evaluation teammates than baselines. 
The best baseline, CoMeDi, achieves unusually high returns with the IPPO teammates on CC and CoR, and with BRDiv teammates on CR and FC, which explains CoMeDi's high mean return on these tasks.
The radar charts also show that the second-best baseline, FCP, is strong specifically against IPPO teammates and relatively weaker on heuristics and BRDiv teammates, especially in CR and CC. As mentioned in the main paper, we attribute FCP's relative strength on IPPO evaluation teammates to the fact that the IPPO evaluation teammates are closer to the training teammate distribution constructed by FCP. While FCP is not especially strong against the ``IPPO pass" agents in CC, these agents were trained via reward shaping to solve the task by passing onions across the counter rather than navigating around the counter, which is the policy found by IPPO without reward shaping (denoted as ``IPPO CC" in the figures).
COLE is also strong against IPPO teammates on CR, CC, CoR, perhaps because it relies on a pure return-maximization objective to generate novel teammates. 
Finally, on AA, Minimax Return is strong against the role-based heuristic teammates and somewhat adversarial BRDiv teammates, while being weaker against the cooperative IPPO teammates. 

\subsection{Human Proxy Evaluations}
\label{app:supp:human_proxy_eval}

The results in the main paper show that ROTATE is better able to generalize to unseen partners compared to baseline methods. 
Here, we evaluate the ability of ROTATE to generalize to \textit{human} partners, compared to baselines. 
The evaluation is conducted with human proxy teammates generated by behavior cloning on the human gameplay dataset published by \citet{carroll_utility_2019}. 
Returns are normalized by the self-play performance of the human proxy teammates.
Table~\ref{tab:human_proxy} shows that ROTATE matches or exceeds the strongest baseline on all five Overcooked layouts. 
No baseline attains a significantly higher return than ROTATE on any layout, and \ours{} holds the highest mean return on CoR and FC. 
\ours{} is also the most consistent across layouts, with the other strong baselines collapsing in performance on at least one layout. 
CoMeDi collapses to 0.43 on AA, Minimax Return to 0.25 on FC, and PAIRED to 0.32 on FC, while ROTATE does not fall below 0.61. 
This pattern mirrors the main result with the default evaluation teammates, in which ROTATE's advantage similarly comes from consistently avoiding cooperative failure with particular teammates.

\begin{table}[h]
    \renewcommand{\arraystretch}{1.3} %
    \captionsetup{skip=10pt} %
    \centering
    \caption{Normalized evaluation return with the human proxy teammate (95\% stratified bootstrap CI's over six trials shown).}
    \label{tab:human_proxy}
    \resizebox{\linewidth}{!}{%
    \begin{tabular}{llllll}
    \hline
        ~ & CR & AA & CC & CoR & FC \\ \hline
        ROTATE & 0.84 [0.82, 0.87] & 0.70 [0.68, 0.71] & 0.87 [0.83, 0.92] & \textbf{0.68} [0.66, 0.73] & \textbf{0.61} [0.58, 0.65] \\
        CoMeDi & \textbf{0.86} [0.86, 0.86] & 0.43 [0.42, 0.44] & \textbf{0.89} [0.82, 0.96] & 0.65 [0.63, 0.68] & 0.57 [0.45, 0.68] \\
        Minimax Return & 0.53 [0.39, 0.64] & \textbf{0.73} [0.67, 0.76] & 0.80 [0.75, 0.84] & 0.40 [0.32, 0.53] & 0.25 [0.22, 0.30] \\
        PAIRED & 0.49 [0.43, 0.58] & 0.61 [0.59, 0.62] & \textbf{0.89} [0.78, 0.98] & 0.50 [0.45, 0.56] & 0.32 [0.30, 0.33] \\
        FCP & 0.24 [0.16, 0.32] & 0.62 [0.55, 0.66] & 0.66 [0.62, 0.72] & 0.43 [0.37, 0.52] & 0.55 [0.51, 0.57] \\
        BRDiv & 0.63 [0.62, 0.64] & 0.57 [0.52, 0.61] & 0.62 [0.47, 0.70] & 0.59 [0.55, 0.61] & 0.50 [0.42, 0.58] \\
        COLE & 0.69 [0.66, 0.71] & 0.45 [0.42, 0.46] & 0.57 [0.55, 0.61] & 0.67 [0.66, 0.67] & 0.55 [0.53, 0.57] \\ \hline
    \end{tabular}%
    }
\end{table}

\subsection{Extension to a Partially Observable Scenario}
\label{app:supp:partial_obs}

We consider a partially observable LBF variant where agents have a field of view limited to radius $2$ in the $7\times7$ grid. A naive extension of \ours{} that simply uses recurrent networks for teammates and BR policies leads to a problem in SXP and XSP trajectory collection. Specifically, under recurrent policies, the BR's hidden state at the switching point is not meaningful to the ego agent.

Instead, we train a recurrent belief model inspired by Off-Belief Learning ~\citep{hu2021off}. The belief model is an LSTM encoder-decoder that takes partial observations as input and predicts the full environment state. 
The use of a belief model allows all agents to condition on a shared state estimate, sidestepping the hidden-state incompatibility. The belief model is trained via supervised learning on rollouts collected from a population of self-play IPPO teams, achieving prediction accuracies of ${\sim}42\%$ and ${\sim}36\%$ for hidden foods and teammate position, respectively (in contrast to ${\sim}9\%$ accuracy for a random model).

\begin{table}[h]
    \renewcommand{\arraystretch}{1.3} %
    \captionsetup{skip=10pt} %
    \centering
    \caption{\ours{} outperforms CoMeDi on partially observable LBF. Mean returns and bootstrapped 95\% CI's are shown.}
    \label{tab:partially_obs_lbf}
    \begin{tabular}{llllll}
    \hline
        Method                & PO LBF \\ \hline
        \ours{} + belief model & \textbf{0.89} [0.87, 0.90] \\
        \ours{} (recurrent)    & 0.75 [0.72, 0.77] \\
        CoMeDi (recurrent)    & 0.70 [0.68, 0.72] \\ \hline
    \end{tabular}
\end{table}

Table~\ref{tab:partially_obs_lbf} shows that \ours{} achieves higher returns on partially observable LBF compared to CoMeDi. Even the naive recurrent extension of \ours{} already outperforms CoMeDi under partial observability. The belief-model variant improves further.

\subsection{Extension to N-Agent Scenario}
\label{app:supp:n_agent}

We consider an N-agent variant of LBF with 2 ego agents and 2 teammates. The extension of ROTATE and CoMeDi is performed using the most straightforward strategy of parameter sharing~\citep{christianos2021scalingmultiagent}, in which the agent ID is appended to the observation to permit behavior differentiation between agents. The ego agents are considered one subteam, and the teammates another.
Table~\ref{tab:n_agent_lbf} shows that \ours{} achieves higher returns on N-agent LBF compared to CoMeDi.

\begin{table}[!h]
    \renewcommand{\arraystretch}{1.3} %
    \captionsetup{skip=10pt} %
    \centering
    \caption{\ours{} outperforms CoMeDi on N-agent LBF. Mean returns and bootstrapped 95\% CI's are shown.}
    \label{tab:n_agent_lbf}
    \begin{tabular}{llllll}
    \hline
        Method  & N-agent LBF \\ \hline
        \ours{} & \textbf{0.887} [0.871, 0.902] \\
        CoMeDi  & 0.672 [0.651, 0.692] \\ \hline
    \end{tabular}
\end{table}

\subsection{Impact of Competence Weights on Sabotage Rates for Per-State and Per-Trajectory Regret}
\label{app:sabotage_game_extended}

This section analyzes the influence of competence weights on the rate of sabotage in the sabotage game specified in Fig.~\ref{fig:sabotage_example_histogram}~(top) for \ours{} with both per-state regret and per-trajectory regret.
We show that as the competence weight increases, the probability of the sabotage action decreases for both per-state and per-trajectory regret, with a much stronger effect observed for per-state regret. 
Further, the competence-bounded per-trajectory regret objective of Sec.~\ref{sec:bounding_competence} (Eq.~\ref{eq:traj_regret_obj_lambda}) does not fully reduce the probability of the sabotage action.

Recall that Eq.~\ref{eq:traj_regret_obj_lambda} (competence-bounded per-trajectory regret) differs from the full ROTATE teammate generation objective only in the distributions over which the regret and teammate-BR return are maximized, and that its competence guarantee (Theorem~\ref{thm:traj_competence}) is with respect to the initial state distribution $p_0$.

\begin{figure}[h]
    \centering
    \includegraphics[width=0.75\textwidth]{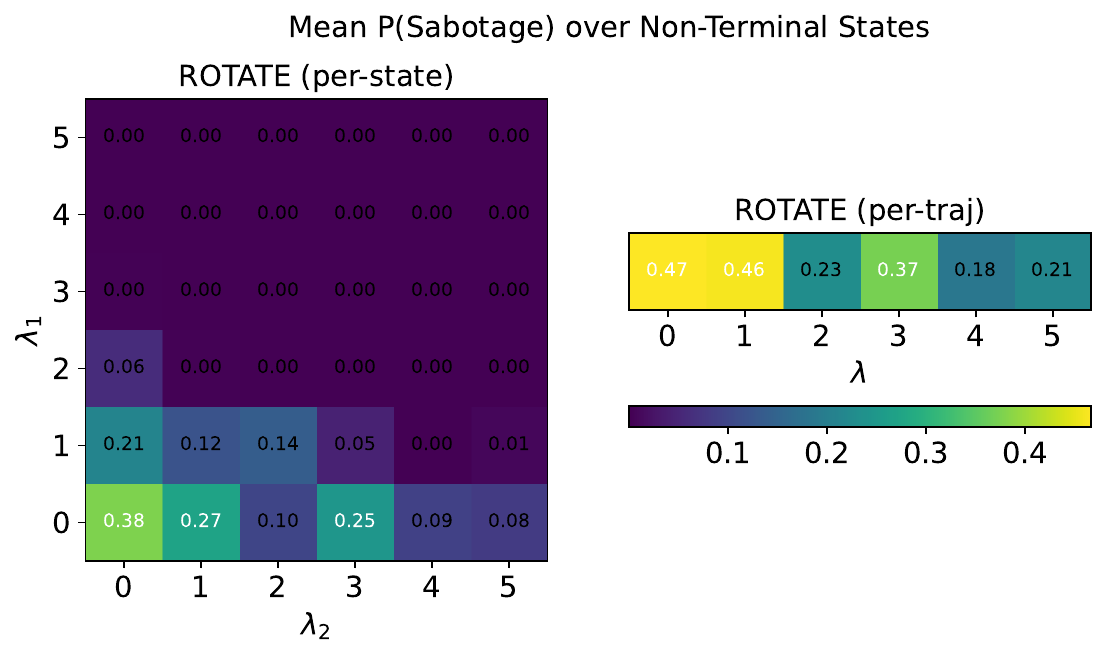}
    \caption{Mean probability of the sabotage action, aggregated across non-terminal states in the sabotage game, as the competence weights are varied. \textbf{(Left)} \ours{} with competence-bounded per-state regret. \textbf{(Right)} \ours{} with competence-bounded per-trajectory regret.}
    \label{fig:sabotage_heatmap_mean}
\end{figure}

Fig.~\ref{fig:sabotage_heatmap_mean} displays the mean probability of sabotage, aggregated across non-terminal states for per-trajectory regret (right) and per-state regret (left), as the competence weights vary from 0 to 5.
For \textit{per-state regret}, the mean probability of sabotage rapidly decreases to 0 as the competence weights increase. The effect of $\lambda_1$ is somewhat stronger than that of $\lambda_2$ at combating sabotage, with $\lambda_1=3, \lambda_2=0$ leading to a zero probability of sabotage, whereas $\lambda_1=0, \lambda_2=5$ still corresponds to a sabotage probability of $0.08$.
However, high enough values of either weight can lead to a zero probability of sabotage.
For \textit{per-trajectory regret}, increasing the competence weight similarly decreases the probability of sabotage, but the effect is much weaker than for per-state regret. 
In fact, even with $\lambda=5$, the probability of sabotage is still 0.21.

\subsection{Generalization Performance as Competence Weights are Varied}
\label{app:supp:comp_weights_gen}

\begin{figure}[h]
    \centering
    \includegraphics[width=\linewidth]{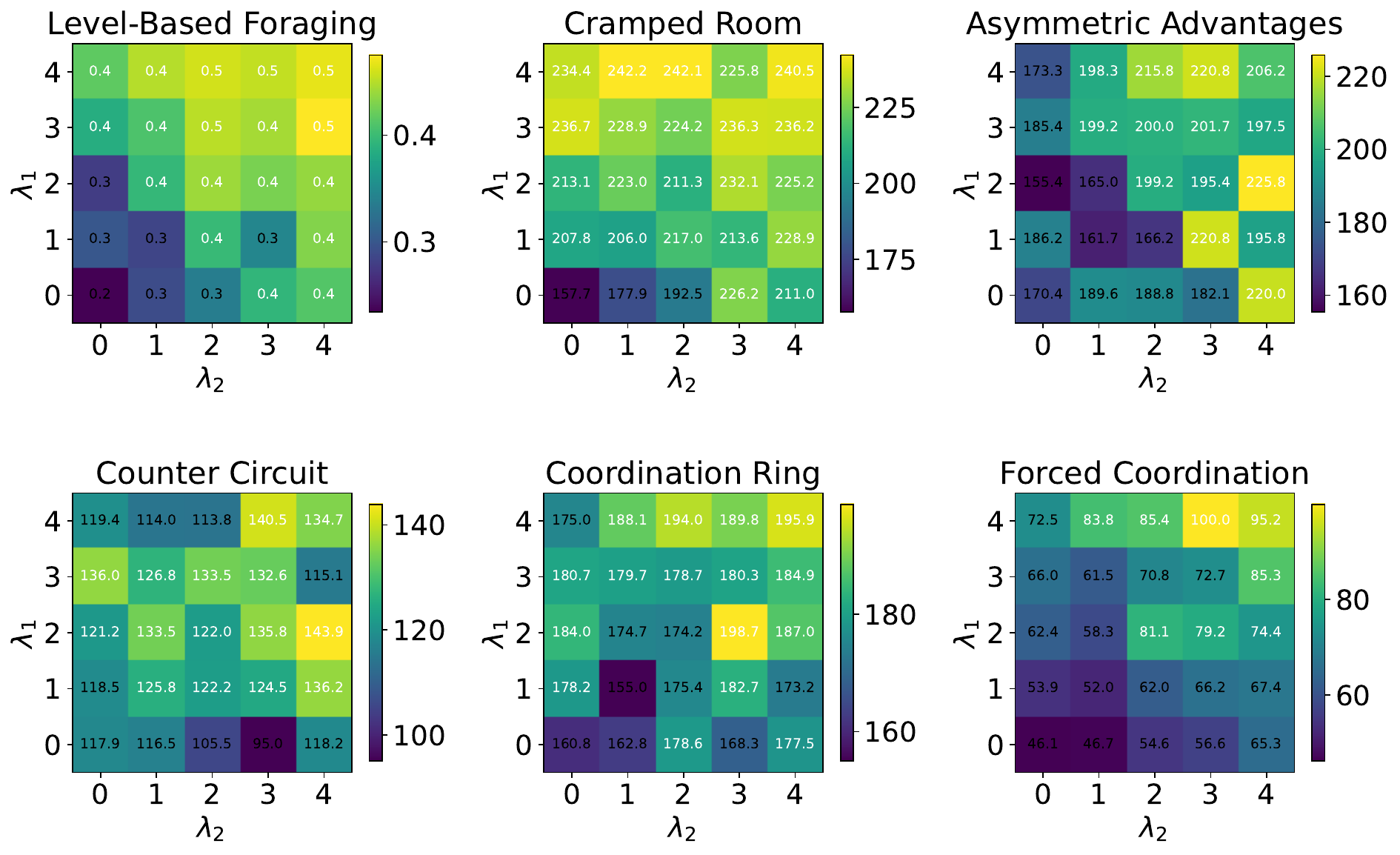}
    \caption{Train returns of \ours{} as the competence weights are varied.}
    \label{fig:comp_weights_train_heatmap}
\end{figure}

\begin{figure}[h]
    \centering
    \includegraphics[width=\linewidth]{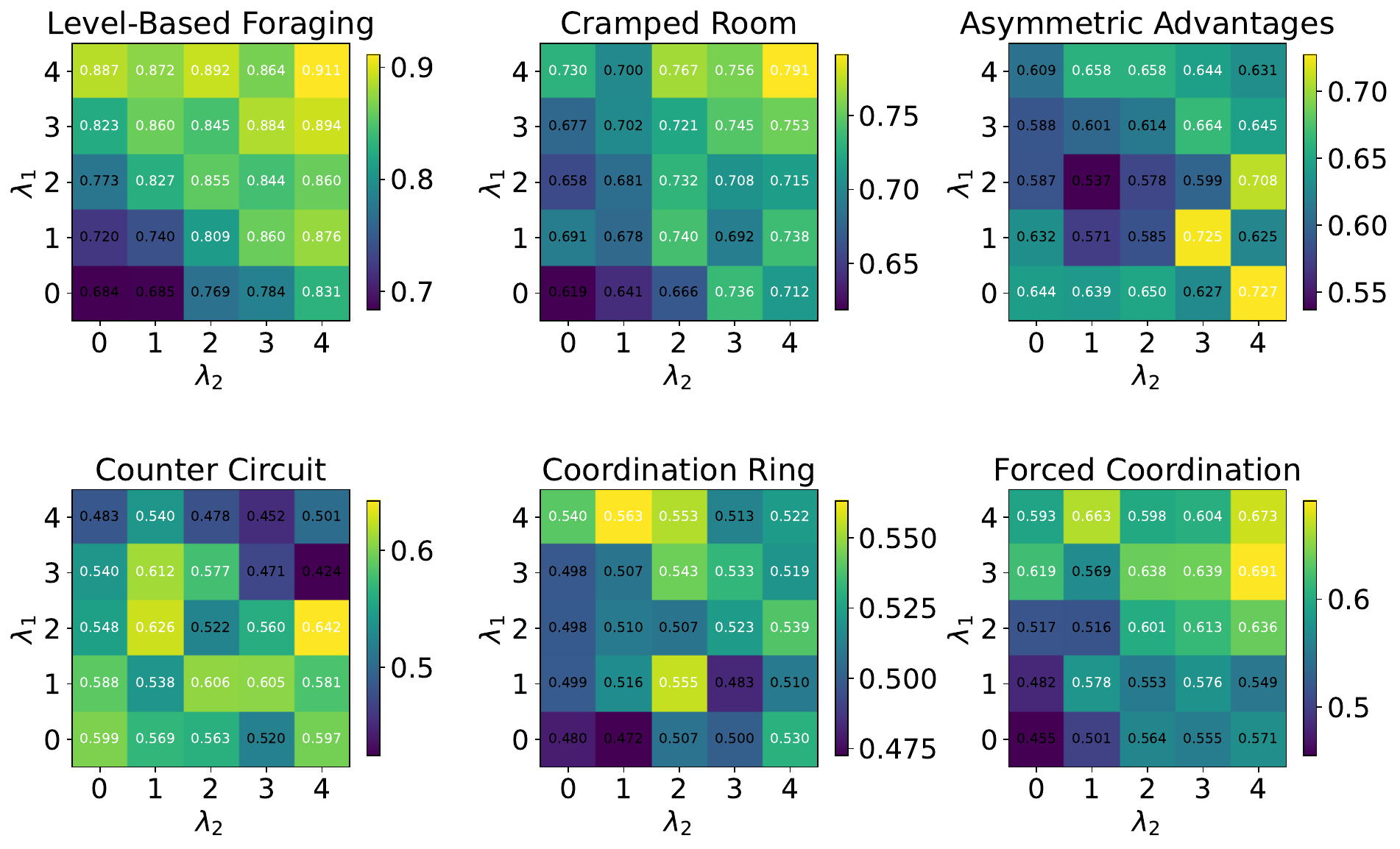}
    \caption{Normalized evaluation returns of \ours{} as the competence weights are varied.}
    \label{fig:comp_weights_eval_heatmap}
\end{figure}

Figures~\ref{fig:comp_weights_train_heatmap} and \ref{fig:comp_weights_eval_heatmap} show the training and evaluation returns of \ours{} as the competence weights $\lambda_1$, $\lambda_2$ are varied, on all tasks.
As the competence weights increase, the training returns tend to increase. 
For evaluation returns, intermediate competence weights perform best on CC, AA, and FC, while higher competence weights perform best on LBF, CR, and CoR.
Therefore, the optimal competence weights for improving cooperative generalization are task-dependent. 

\subsection{Ablations of Teammate Generation Objective}
\label{app:supp:teammate_gen_ablations}

We consider two alternatives to the teammate generation objective of \ours{} within the \ours{} framework.
First, we consider competence-bounded per-trajectory regret.
Second, we consider a version of \ours{} in which expected returns are maximized in states sampled from CoMeDi's mixed-play state distribution, $\pmp$ (defined in App.~\ref{app:supp_disc:comedi_mp_discussion}), rather than from SXP states, which resembles the mixed-play objective of CoMeDi~\citep{sarkar2023diverse} (as motivated in App.~\ref{app:supp_disc:comedi_mp_discussion}).

\begin{figure}[t]
    \centering
    \begin{subfigure}[t]{0.48\linewidth} %
        \includegraphics[width=\linewidth]{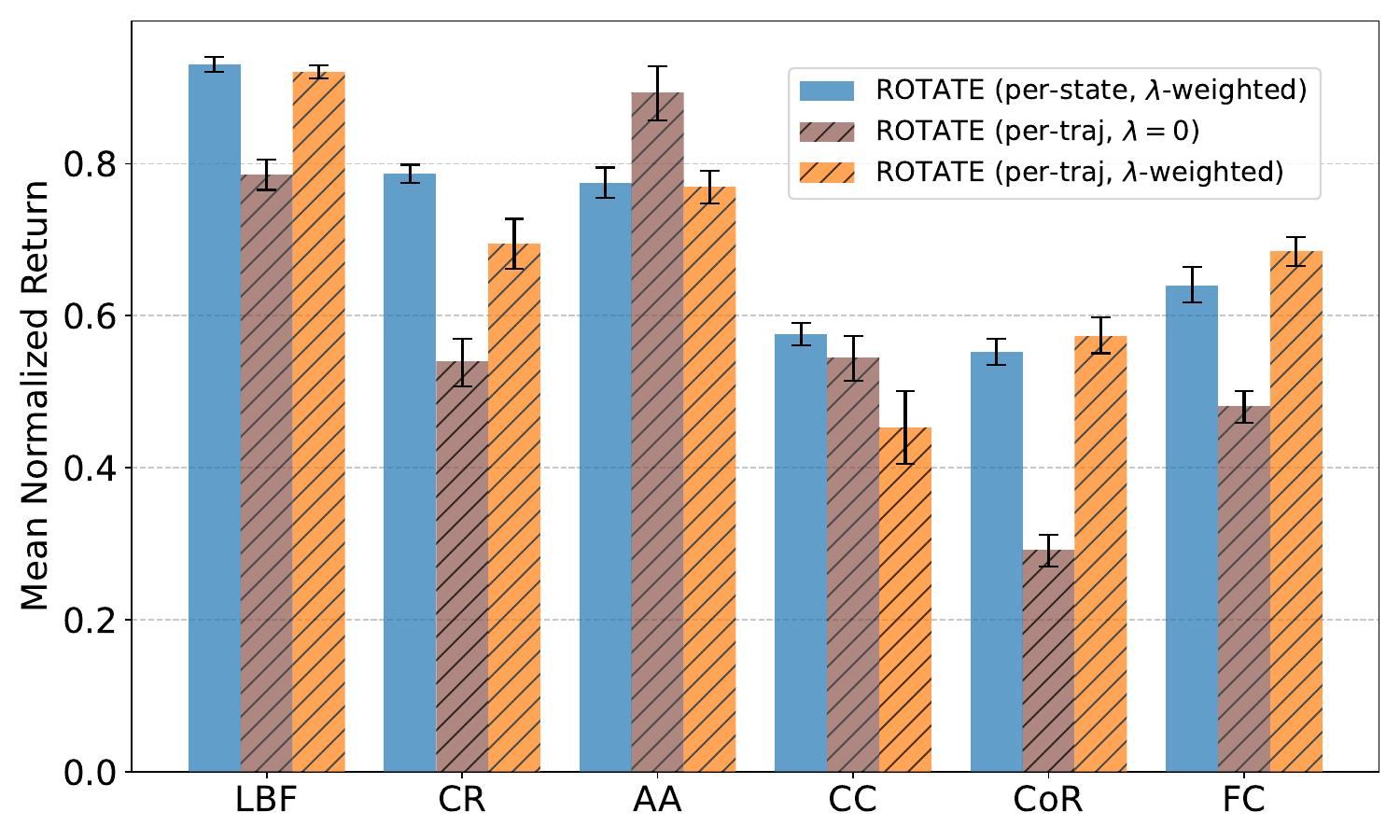}
        \caption{\ours{} with per-state vs per-trajectory regret, with and without $\lambda$-enforced competence bounds.}
        \label{fig:rotate_traj_regret_lambda}
    \end{subfigure}
    \hfill
    \begin{subfigure}[t]{0.48\linewidth} %
        \centering
        \includegraphics[width=\linewidth]{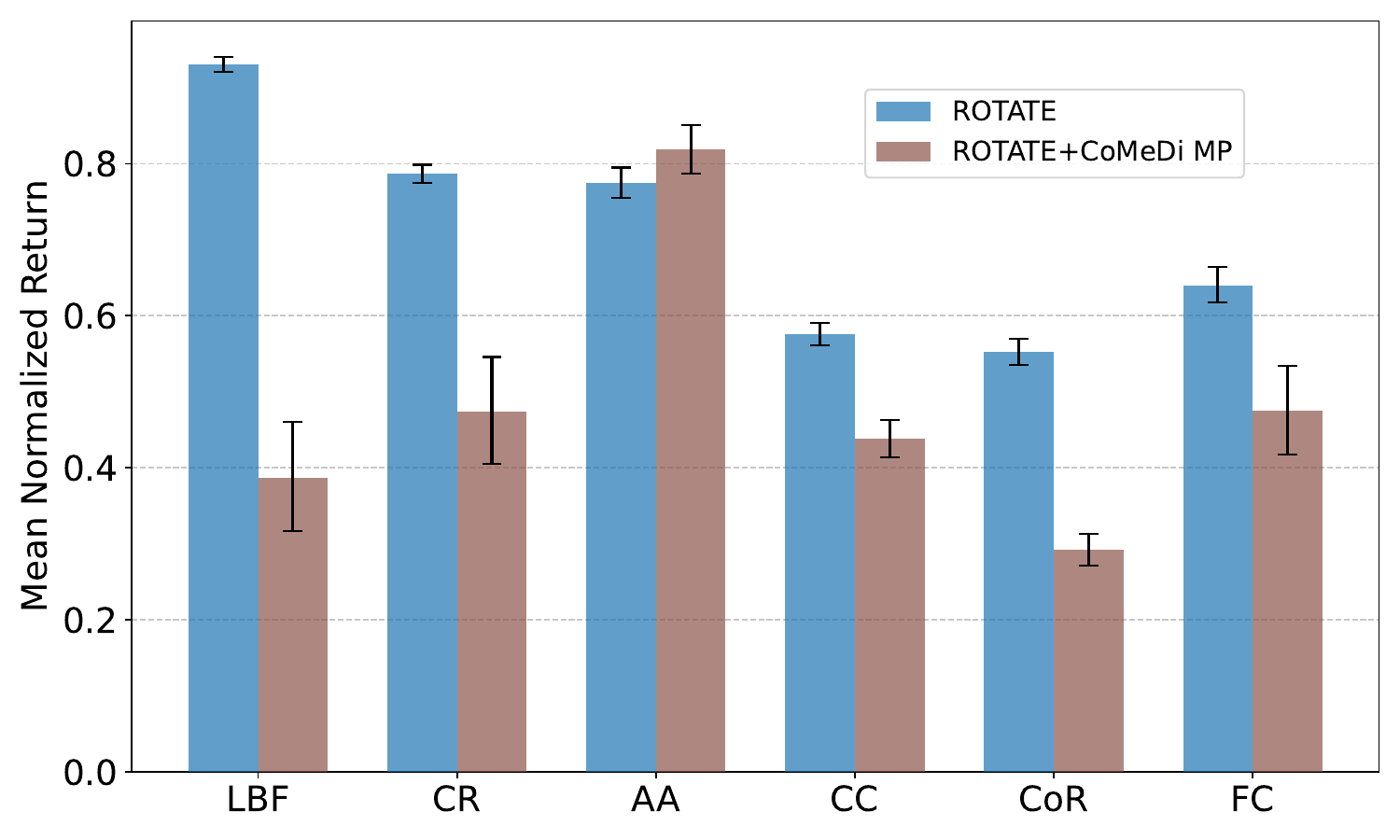}
        \caption{\ours{} vs \ours{} with CoMeDi's mixed-play (MP) objective.}
        \label{fig:ablations}
    \end{subfigure}
    \caption{\small The mean normalized returns of \ours{} and ablations designed to evaluate the effectiveness of \ours{}'s per-state regret-based teammate generation objective. 95\% stratified bootstrapped CI's over three trials are shown.}
    \label{fig:more_rotate_variations}
\end{figure}

\paragraph{Competence Bounded Trajectory Regret} 
The \ours{} teammate generation objective consists of competence-bounded per-state regret.
A natural question is how competence-bounded per-trajectory regret (Eq. \ref{eq:traj_regret_obj_lambda}) might compare in terms of training strong ego agents. 
As shown in Fig.~\ref{fig:rotate_traj_regret_lambda}, \ours{} outperforms per-trajectory regret ($\lambda=0$) on every task except AA. Because AA is also the task where adversarially robust behavior is optimal, this provides indirect evidence that per-trajectory regret induces adversarial behavior. 
Bounding competence by adding the same per-task $\lambda$ values used by \ours{} to the per-trajectory objective raises its return on most tasks relative to $\lambda=0$, and reduces the gap to \ours{} on 4/6 tasks. 
Therefore, for these benchmark tasks, placing a $p_0$ competence bound on per-trajectory regret is sufficient to reduce sabotaging behavior of plain per-trajectory regret. 
We still expect competence-bounded per-state regret (the \ours{} objective) to produce stronger cooperative generalization in broader tasks.
As discussed in Sec.~\ref{sec:per_state_regret}, enforcing competence on a per-state rather than per-trajectory basis is important for reducing the teammate's incentive to sabotage, which is demonstrated on the sabotage matrix game in App.~\ref{app:sabotage_game_extended}.
 
\paragraph{Mixed Play Ablation}
For fair comparison, the mixed-play variant uses the same hyperparameters as \ours{} and CoMeDi's tuned mixed-play weights. 
\ours{}'s teammate generation objective leads to higher normalized returns than maximizing CoMeDi's mixed-play objective in all environments except for Overcooked's Asymmetric Advantages (AA) setting (Fig.~\ref{fig:ablations}). 
Following the difference in starting states of trajectories for which these two maximize self-play returns, we conjecture that this is because \ours{} generates competent teammate policies in states from XP while the CoMeDi-like approach imposes the same thing in states from $p_{\text{MSTART}}$. Imposing good faith within policies in $\pxp$ is likely more important for training an ego agent that initially interacts with $\pi^{-i}$ during training by visiting states from $\pxp$. 

\subsection{Ablations of the \ours{} Population}
\label{app:supp:ind_ego_agent_pop}
\begin{figure}[t]
    \centering
    \includegraphics[width=0.48\linewidth]{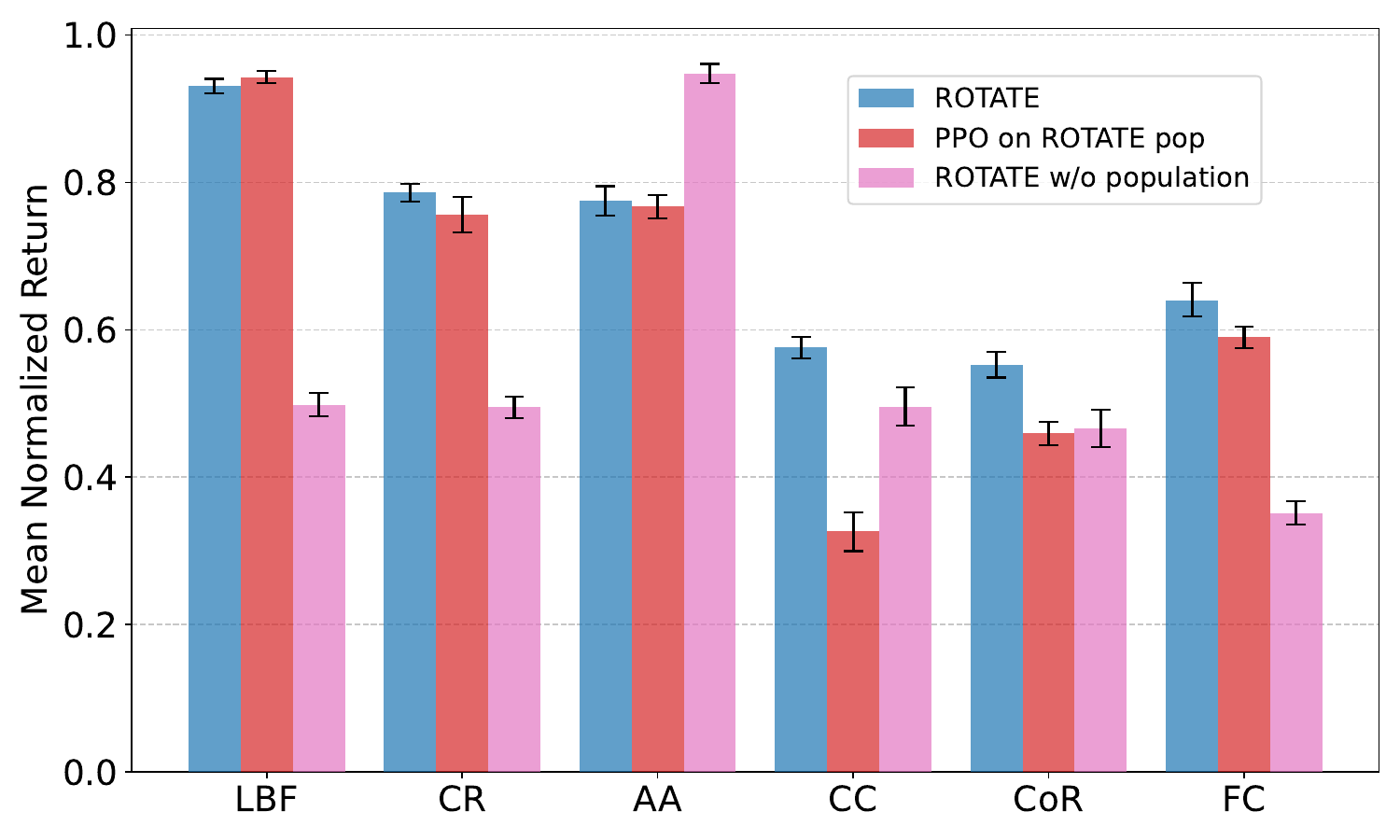}
    \caption{\small The mean normalized returns of \ours{} compared to an independently trained ego agent on \ours{}'s population and an ablation of \ours{} without the population. 95\% stratified bootstrapped CI's over three trials are shown.}
    \label{fig:ego-vs-rotate-pop}
\end{figure}

This subsection presents two ablations of the \ours{} population, both reported in Fig.~\ref{fig:ego-vs-rotate-pop}. The first removes the population buffer described in Sec.~\ref{sec:method}, and the second trains an independent ego agent on the final \ours{} population.

\paragraph{Is the population buffer necessary for \ours{} to learn well? (Yes.)}
We hypothesize that collecting all previously generated teammates in a population buffer helps the \ours{} agent improve in robustness against all previously discovered conventions. 
On the other hand, if there is no population buffer, then it becomes possible for the \ours{} ego agent to forget how to collaborate with teammates seen at earlier iterations of open-ended learning~\citep{kirkpatrick_overcoming_2017}, which creates the possibility that the ego agent and teammate generator oscillate between conventions.
As shown in Fig.~\ref{fig:ego-vs-rotate-pop}, \ours{} without the population buffer attains lower evaluation returns than the full \ours{} method on all tasks except AA, thus supporting the hypothesis that the population buffer improves ego agent learning.
As discussed in the RQ1 results (Sec.~\ref{sec:exp:results}), AA is a unique layout where agents can complete the task independently, even in the presence of an adversarial partner. 
As a corollary, there are few meaningful cooperative conventions that can be discovered, and no scenarios where convention mismatch leads to zero return (unlike LBF and FC). 

\paragraph{Is the \ours{} population useful for training an independent ego agent? (Yes.)}
Two-stage AHT algorithms first generate a population of teammates, and next train an ego agent against the population.
Although \ours{}'s teammate generation mechanism relies on the learning process of a particular ego agent, here, we investigate whether the population generated by \ours{} is useful for training independently generated ego agents.
Fig. \ref{fig:ego-vs-rotate-pop} compares the mean evaluation returns of the \ours{} ego agent against the mean evaluation returns of an independently trained PPO ego agent that was trained on the full, final \ours{} population, using the same configuration as \ours{}.
In 3/6 tasks (CC, CoR, and FC) the \ours{} ego agent outperforms the independently trained ego agent with non-overlapping confidence intervals, and on the remaining three tasks (LBF, CR, and AA) the two perform similarly.
This result suggests that the \ours{} population is a useful population of teammates even independent of the particular ego agent generated.

Note that because the population used to train the independent ego agent was itself produced by \ours{}'s coupled generation, this experiment isolates the ego agent's training regime---the curriculum induced by coupled teammate generation versus joint training on the full, final population---rather than the value of coupled generation itself. 
In continual learning, joint training on all tasks is often considered an oracle upper bound for sequential training~\citep{liu2023libero}. 
Viewed this way, the independently trained ego agent's strong performance is unsurprising: it faces a stationary training distribution and interacts with all teammates uniformly throughout training, whereas the \ours{} ego agent faces a nonstationary, growing population and trains against later teammates for fewer iterations. Closing the remaining gap to joint training (e.g., via curated buffer sampling) is left to future work.

\subsection{Inspection of \ours{} Generated Teammates}
\label{app:supp:teammate_inspection}

This section analyzes the \ours{} teammate populations on LBF and CC for the learned conventions, presence of handshake conventions, competence and diversity (as measured by XP matrices), and the regret induced on the ego agent during training. Results are in \textit{unnormalized} return units. 
For reference, the maximum return on LBF is 0.5 (corresponding to 100\% of food eaten in the percent-eaten results; see App.~\ref{sec:app_exp_tasks}), while CC has no well-defined maximum but awards 20 points per delivered dish.

\paragraph{Conventions.} 
On CC, the behavior of \ours{} teammates is characterized by using 63 features to create embedding vectors that characterize interactions between each teammate and its co-trained BR. 
The features track handoffs, region occupancy, resource usage, role specialization, and routing.
The resultant embedding vectors, z-scored and weighted by feature group and test-retest reliability, are clustered by cosine distance in feature space using agglomerative clustering~\citep{murtagh2012algorithms} with average linkage, where the number of clusters is chosen by silhouette score together with a bootstrap stability check over episodes.
Because competence and conventions could jointly influence cluster structure, we verified that similar-return teammates do not always share clusters. 

To analyze the conventions contained in the \ours{} teammate population, we visualize the teammate-BR corresponding to each cluster medoid, to find that the population includes the following 6 conventions:
\begin{itemize}
    \item \textit{3 onion specialists (pass to self, left path / pass to teammate / left path, clockwise)}: pass onions to themselves across the counter routing along the left side,  pass onions across the counter to the partner, or pass onions routing along the left side and traveling clockwise, respectively.
    \item \textit{Handoff receiver with delivery}: receives onions from a giver, pots them, then plates and delivers.
    \item \textit{Independent, left path}: completes the full task alone via the left path.
    \item \textit{Independent with occasional passing, clockwise}: mostly solo, but coordinates via occasional onion passes, moving clockwise with the partner.
\end{itemize}

\begin{figure}[t]
    \centering
    \includegraphics[width=0.7\linewidth]{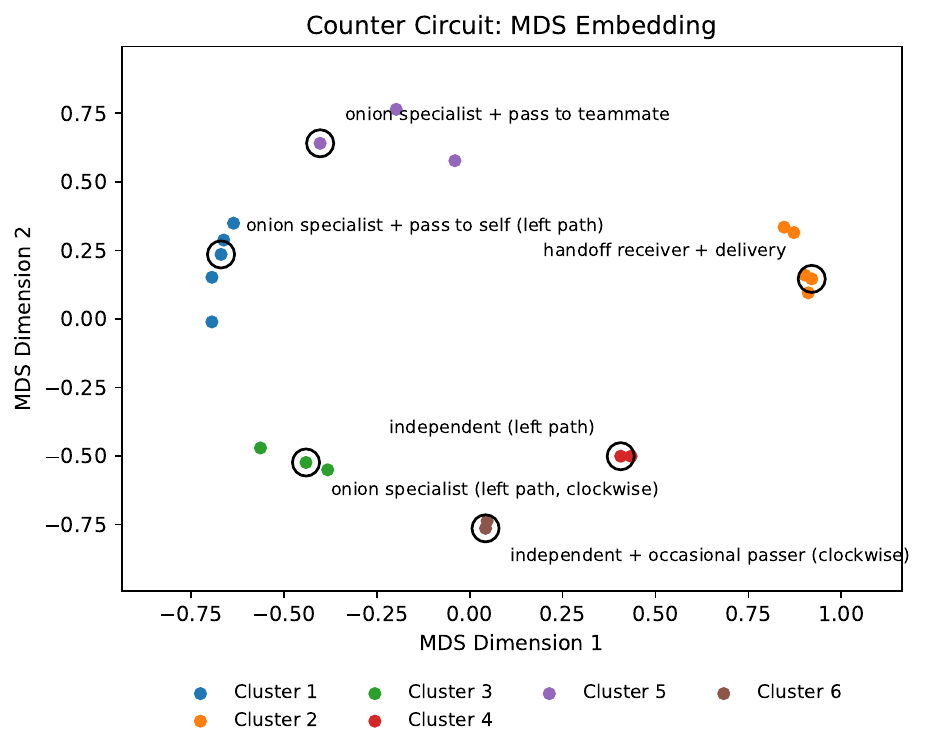}
    \caption{\small MDS embedding of the pairwise cosine distances between the behavioral feature vectors of Counter Circuit teammates, colored by agglomerative clustering. Cluster medoids are labeled with the convention identified by visually inspecting trajectory replays.}
    \label{fig:cc_convention_clusters}
\end{figure}

 We observe no sabotage-like behaviors such as blocking, resource hoarding, or delivery prevention.
 Fig.~\ref{fig:cc_convention_clusters} visualizes the cluster structure by applying multidimensional scaling (MDS)~\citep{torgerson1952multidimensional} to the pairwise cosine distance matrix.
 Each cluster medoid is labeled by the corresponding convention.

\paragraph{Perturbation analysis of handshake-style sabotage.} 
\ours{}'s teammate generation objective is designed to mitigate adversarial behaviors such as \textit{handshake conventions}~\citep{sarkar2023diverse} (policies that coordinate only in SP states and otherwise act in ways that harm coordination). 
\ours{} gathers XP interactions (teammate-ego) starting from states visited under SP (teammate-BR), and gathers SP interactions starting from states visited under XP, so no state is exclusive to SP or XP interaction, thereby discouraging handshakes. 

To empirically examine whether \ours{} teammates-BRs have learned handshake conventions, we measure each teammate $\pi$'s return when paired with its noise-perturbed BR, $(1-\epsilon)\cdot BR(\pi) + \epsilon\cdot\pi_{\text{rand}}$, sweeping $\epsilon$ from 0 to 1.
The returns are averaged over all teammates in the population (dropping two teammates that collapsed to zero return on LBF).
A teammate playing a handshake policy would show a sharp decline in team return at some level of $\epsilon$. 
Instead, Table~\ref{tbl:perturbation_returns} shows that as $\epsilon$ increases, the returns decline smoothly starting from $\epsilon=0.8$ on LBF and from $\epsilon=0.2$ on CC.

\begin{table}[t]
    \renewcommand{\arraystretch}{1.3}
    \captionsetup{skip=10pt}
    \centering
    \small
    \caption{\small Mean return of each \ours{} teammate paired with its noise-perturbed best response, $(1-\epsilon)\cdot BR(\pi) + \epsilon\cdot\pi_{\text{rand}}$, as the perturbation level $\epsilon$ is swept from 0 to 1 on LBF and Counter Circuit (CC).}
    \label{tbl:perturbation_returns}
    \resizebox{\linewidth}{!}{%
    \begin{tabular}{lccccccccccc}
    \hline
    Task & $\epsilon{=}0$ & 0.1 & 0.2 & 0.3 & 0.4 & 0.5 & 0.6 & 0.7 & 0.8 & 0.9 & 1.0 \\ \hline
    LBF & 0.498 & 0.499 & 0.499 & 0.499 & 0.498 & 0.497 & 0.488 & 0.455 & 0.368 & 0.252 & 0.131 \\
    CC & 140.6 & 140.4 & 136.0 & 128.4 & 116.8 & 102.3 & 83.5 & 63.0 & 43.2 & 27.0 & 16.3 \\ \hline
    \end{tabular}
    }
\end{table}

\begin{figure}[t]
    \centering
    \includegraphics[width=0.95\linewidth]{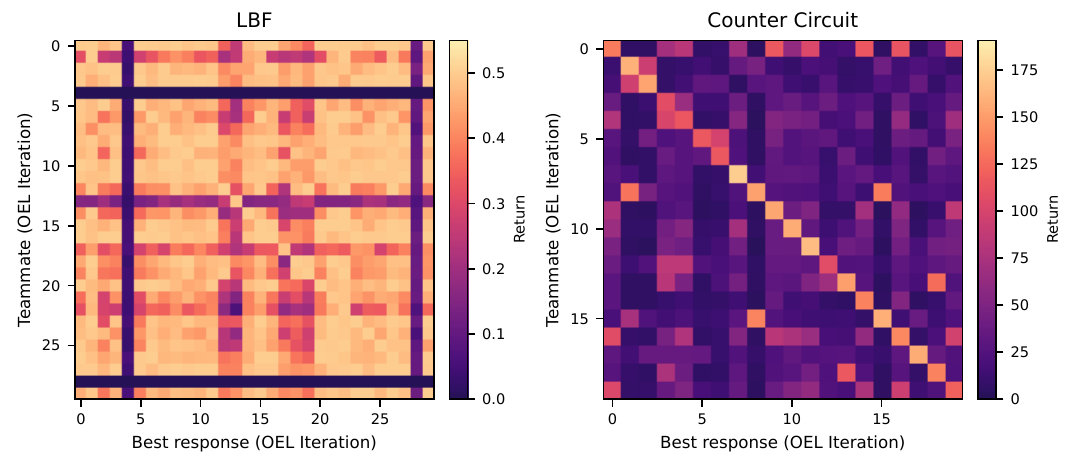}
    \caption{\small Cross-play matrices of \ours{} teammates (rows) evaluated with each teammate's co-trained best response (columns) on LBF and Counter Circuit, in raw return units.}
    \label{fig:xp_matrices}
\end{figure}

\paragraph{Teammate competence and diversity (XP matrix).} 
To analyze the competence and diversity of \ours{}'s teammate population, we compute the cross-play matrix of \ours{} teammates and co-trained BRs (Fig.~\ref{fig:xp_matrices}). 
The diagonal displays the estimated SP returns, and therefore estimates the teammates' competence.
The ratio between diagonal and off-diagonal returns provides evidence of convention diversity, since prior sections verified the absence of adversarial conventions.

On LBF, the XP matrix shows generally high values, but still shows zones of lower returns.
It also reveals that 2/30 teammates achieve uniformly zero returns; visual inspection confirmed that these teammates collapsed to idle behavior. 
The diagonal returns are near-optimal, and remain somewhat higher than the off-diagonals (mean of 0.499 vs 0.427, with the two idle teammates removed).
Taken together with the fact that \ours{} achieves high cooperative generalization with LBF evaluation teammates (Fig.~\ref{fig:rotate_main_results}~(left)), the XP matrix shows that there are few truly incompatible conventions on LBF, which is consistent with the XP matrix of the evaluation teammates (Fig.~\ref{fig:heldout_xp_matrices}).
On CC, the XP matrix shows a much higher gap between the diagonal and off-diagonal returns, consistent with our conventions analysis which showed that \ours{} teammates learned distinct, interpretable conventions on CC.

\begin{figure}[h]
    \centering
    \includegraphics[width=0.95\linewidth]{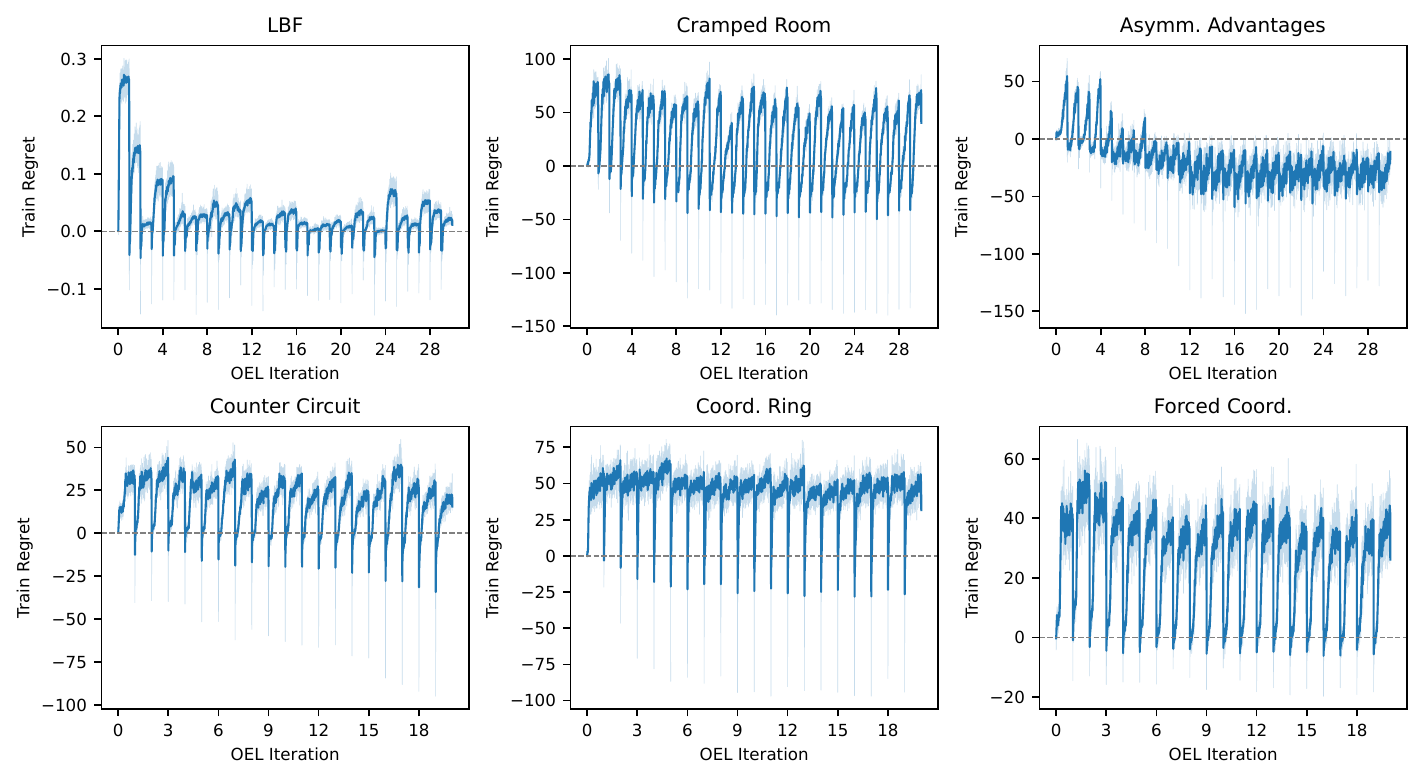}
    \caption{\small Regret induced on the ego agent during each teammate generation phase of \ours{} open-ended learning, in unnormalized return units.}
    \label{fig:train_regret_curves}
\end{figure}

\paragraph{Regret induced by teammates on the ego agent.} 
Fig.~\ref{fig:train_regret_curves} visualizes the (per-trajectory) regret induced by the teammate on the ego agent during each \textit{teammate generation phase} of open-ended learning  (i.e., $\mathbb{E}_{s_0 \sim p_0} [V(s_0 \mid \pi^{-i}, \BR(\pi^{-i})) - V(s_0 \mid \pi^{-i}, \piego)]$, in unnormalized return units). 
Note that the ego policy is frozen during the teammate generation phase, while the teammate and BR are initialized from scratch, which leads to the spikes and dips observed in the regret curves. 
At the start of each OEL iteration, the teammate and BR are learning from scratch against the current ego agent, so regret is initially negative.
Within each iteration, as the teammate/BR learn to cooperate, regret increases smoothly, rising above zero in all tasks except for AA, where the regret remains negative after the 8th OEL iteration.
We partially attribute the negative regret in AA to the increasing robustness of the ego agent over OEL iterations, and attribute the remainder to lack of sufficient training timesteps per-iteration. 
Indeed, in most tasks, the maximum per-iteration regret declines over iterations, with the effect being especially strong in LBF, CC, and FC. 
In AA, the teammates have minimal coordination opportunities with the ego agent, which may explain why the regret there remains negative.

Due to the \ours{} competence-bounded per-state regret objective, the teammate must still act cooperatively with the ego, so the per-iteration maximum regret does not become too high. On LBF, the maximum return is 0.5 (in raw return units; see App.~\ref{sec:app_exp_tasks}), but the per-iteration regret peaks are typically less than 0.1.
On CC, the regret peaks are between 25-50, which reflects a difference in SP and XP returns corresponding to delivering 2-4 dishes.

\subsection{Quality of \ours{} Co-Trained Best Responses}
\label{app:supp:br_quality}

\ours{}'s regret estimates rely on co-trained PPO approximations of the BR, which may fail to find the true BR. 
To assess the quality of these approximations, we independently train PPO BR policies (three seeds) for five \ours{} teammates on LBF and CC, with twice the training budget of the co-trained BR. 
Tables~\ref{tab:br_quality_lbf} and~\ref{tab:br_quality_cc} report the per-teammate returns.
The results show no systematic underestimation in either task. On LBF, the mean gap across 5 teammates (ind - co) is +0.0052, while on CC, the mean gap is +0.9.
While it may be surprising that the co-trained BRs can sometimes outperform the independently trained BRs, we attribute it to the fact that the co-trained BRs have a different training regime from the independently trained BRs. Specifically, the co-trained BRs are trained to maximize SP returns from XP states (i.e. SXP states) in order to remain a BR to the teammate everywhere that the two interact.

\begin{table}[h]
    \renewcommand{\arraystretch}{1.3}
    \captionsetup{skip=10pt}
    \centering
    \small
    \caption{\small Mean return of each \ours{} teammate's co-trained best response on LBF vs.\ an independently trained PPO best response, with 95\% CIs displayed.}
    \label{tab:br_quality_lbf}
    \begin{tabular}{lccc}
    \hline
        Teammate (iter.) & Co-trained BR & Independent BR & Gap (Ind - Co) \\ \hline
        0 & 0.493 [0.483, 0.500] & 0.497 [0.493, 0.500] & +0.004 \\
        5 & 0.488 [0.473, 0.500] & 0.499 [0.496, 0.500] & +0.010 \\
        10 & 0.492 [0.480, 0.500] & 0.495 [0.490, 0.499] & +0.003 \\
        13 & 0.487 [0.474, 0.497] & 0.496 [0.492, 0.499] & +0.009 \\
        25 & 0.499 [0.496, 0.500] & 0.499 [0.496, 0.500] & -0.000 \\ \hline
    \end{tabular}
\end{table}

\begin{table}[h]
    \renewcommand{\arraystretch}{1.3}
    \captionsetup{skip=10pt}
    \centering
    \small
    \caption{\small Return of each \ours{} teammate's co-trained best response on Counter Circuit versus an independently trained PPO best response with 95\% CIs displayed.}
    \label{tab:br_quality_cc}
    \begin{tabular}{lccc}
    \hline
        Teammate (iter.) & Co-trained BR & Independent BR & Gap (Ind - Co) \\ \hline
        3 & 107.2 [103.6, 110.5] & 109.0 [107.1, 110.8] & +1.8 \\
        6 & 113.4 [107.8, 118.6] & 136.8 [132.6, 140.8] & +23.4 \\
        7 & 178.4 [174.7, 182.0] & 149.3 [144.3, 154.2] & -29.1 \\
        12 & 116.1 [110.6, 121.1] & 117.5 [114.7, 120.1] & +1.4 \\
        18 & 130.3 [123.0, 137.0] & 137.3 [133.5, 140.9] & +7.0 \\ \hline
    \end{tabular}
\end{table}

\vspace{-5pt}
\subsection{Learning Curves}
\label{app:supp:learning_curves}
\begin{figure}[h]
    \centering
    \begin{subfigure}[h]{0.48\linewidth} %
        \centering
        \includegraphics[width=\linewidth]{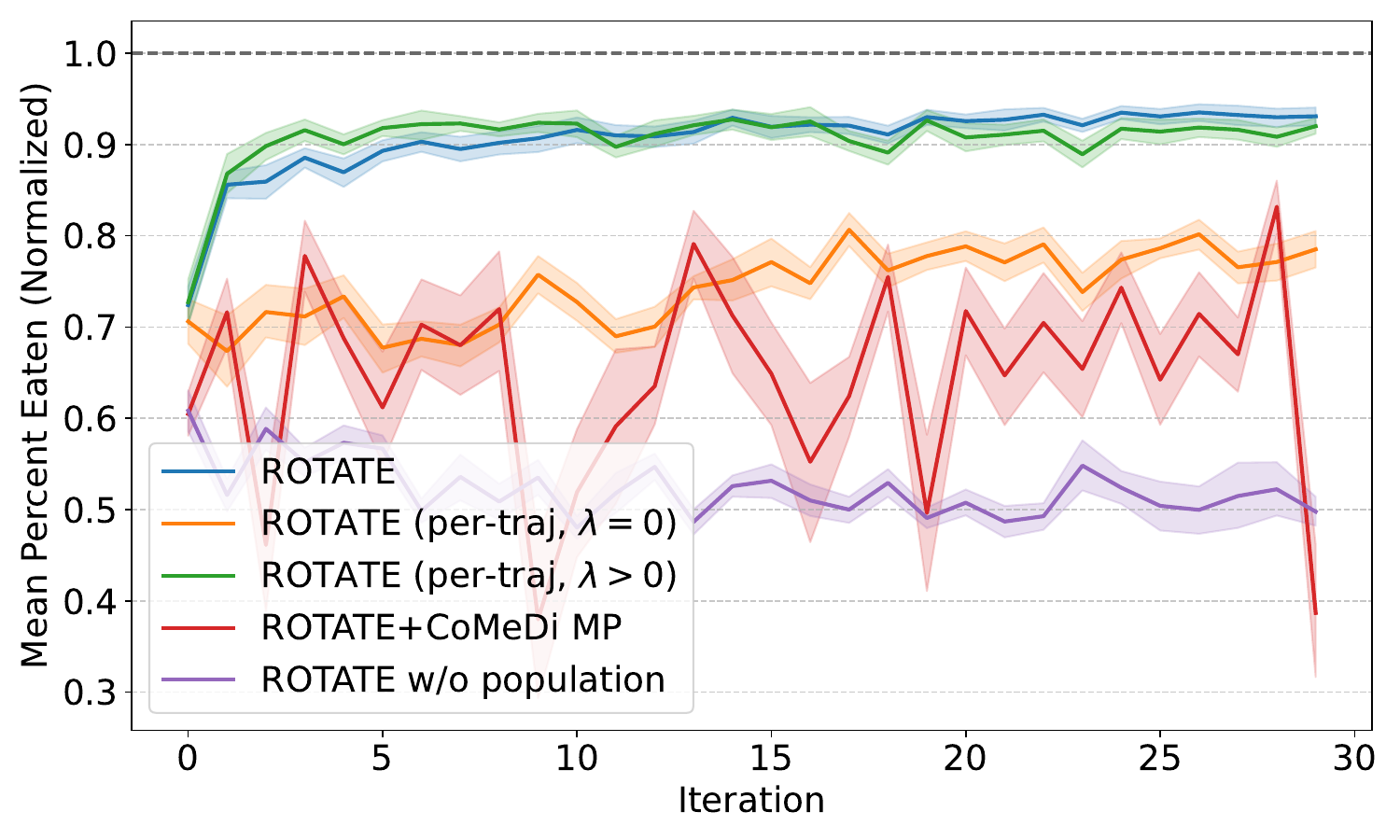}
        \caption{LBF.}
        \label{fig:lbf_curves_ablations}
    \end{subfigure}
    \begin{subfigure}[h]{0.48\linewidth} %
        \centering
        \includegraphics[width=\linewidth]{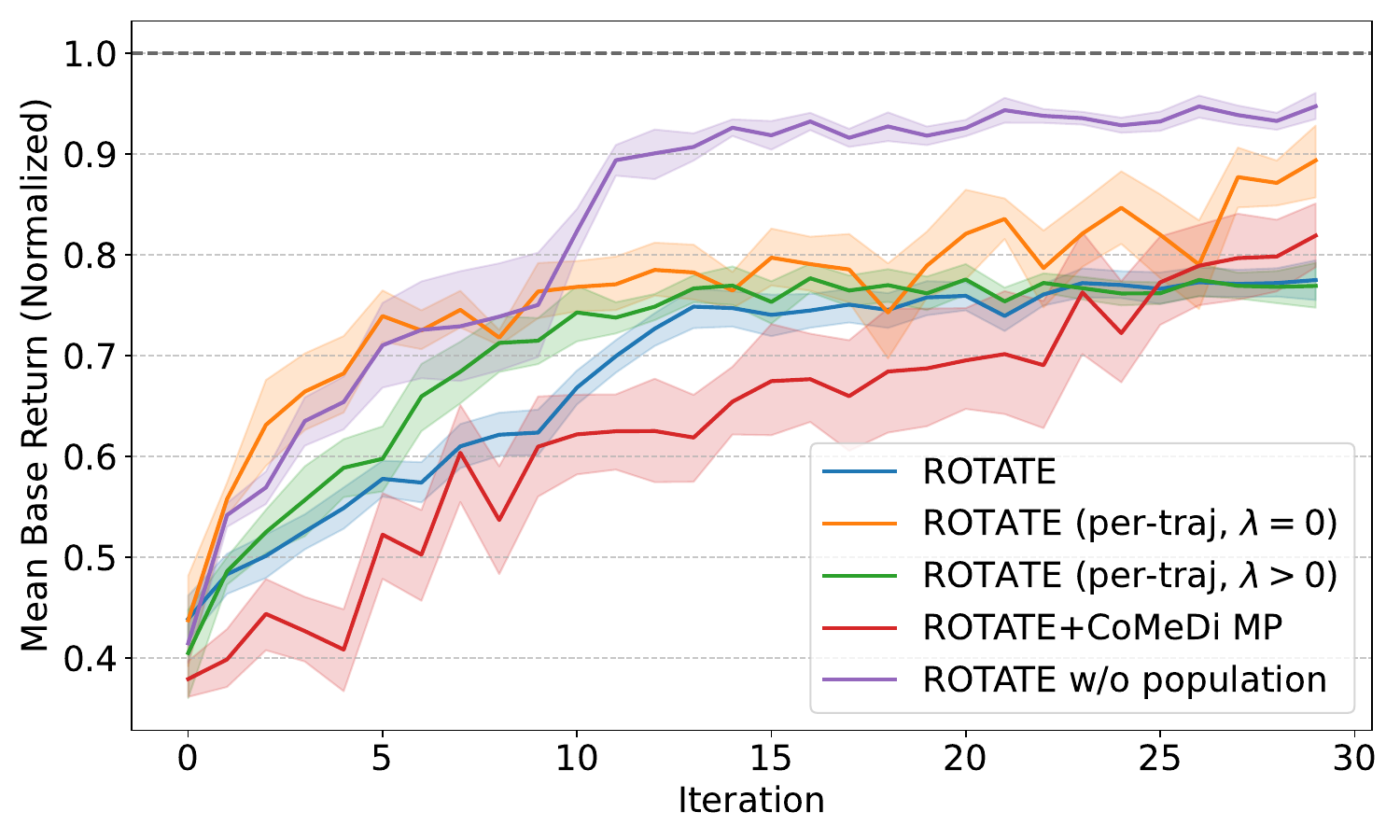}
        \caption{Asymmetric Advantages.}
        \label{fig:asymm_adv_curves_ablations}
    \end{subfigure}
    \begin{subfigure}[h]{0.48\linewidth} %
        \centering
        \includegraphics[width=\linewidth]{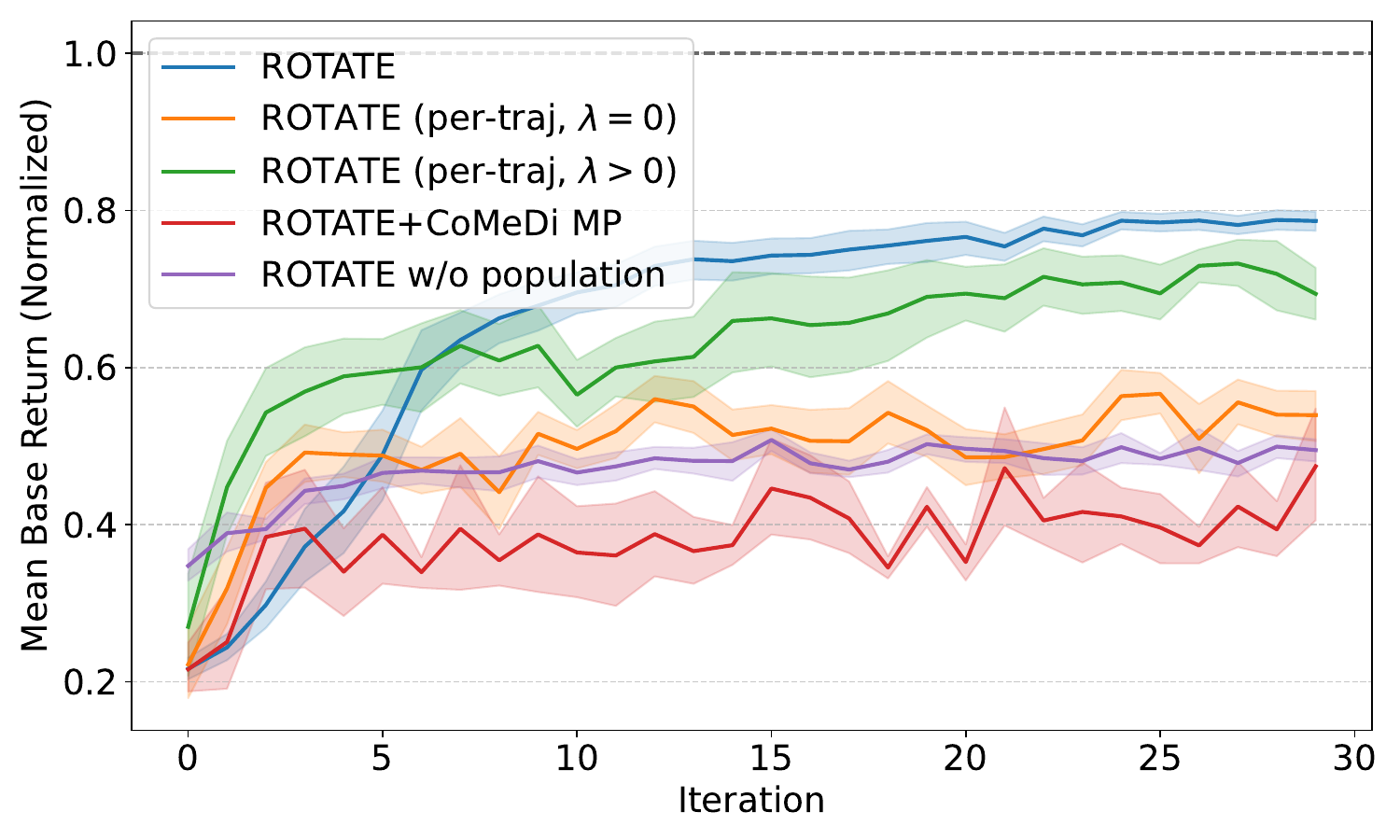}
        \caption{Cramped Room.}
        \label{fig:cramped_room_curves_ablations}
    \end{subfigure}
    \begin{subfigure}[h]{0.48\linewidth} %
        \centering
        \includegraphics[width=\linewidth]{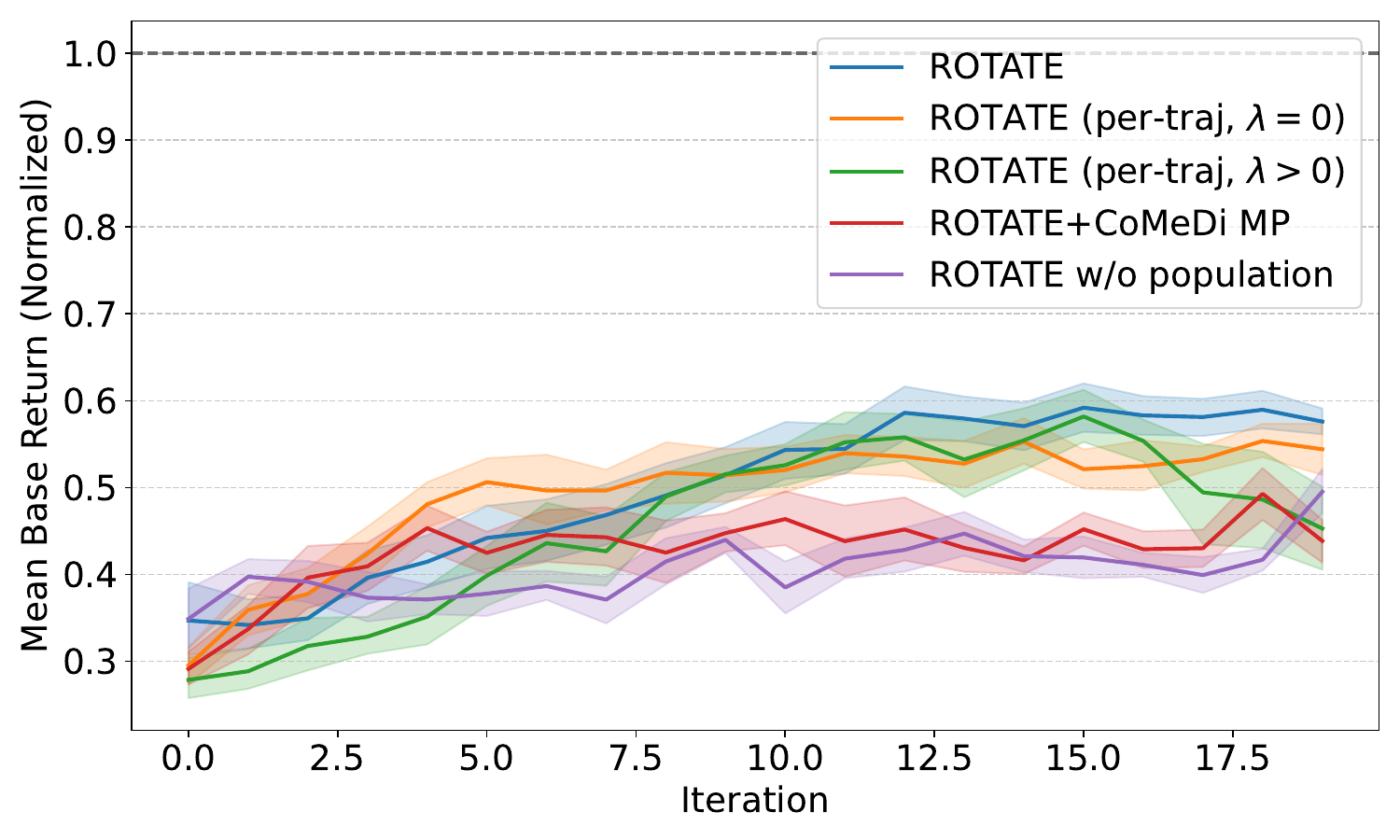}
        \caption{Counter Circuit.}
        \label{fig:counter_circuit_curves_ablations}
    \end{subfigure}
    \begin{subfigure}[h]{0.48\linewidth} %
        \centering
        \includegraphics[width=\linewidth]{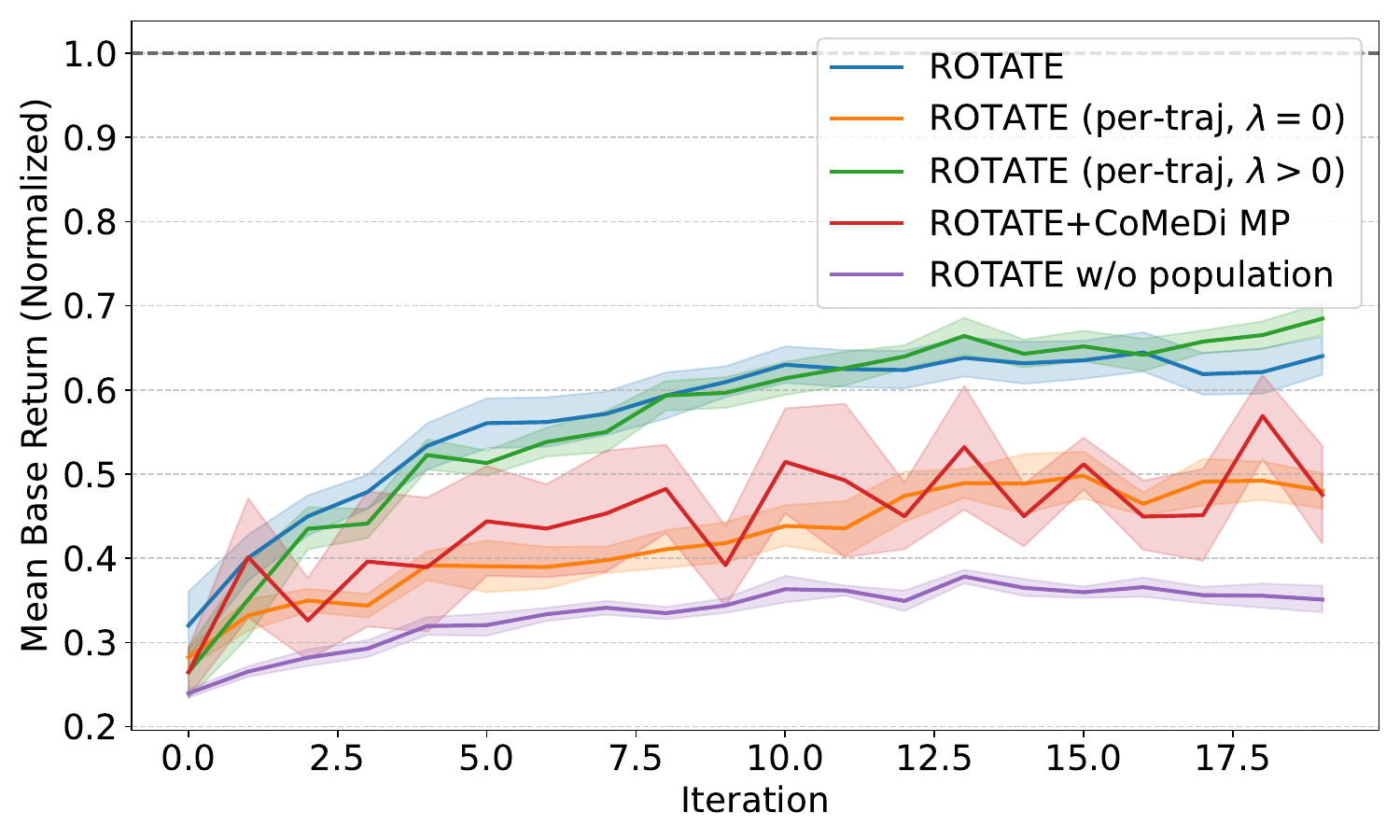}
        \caption{Forced Coordination.}
        \label{fig:forced_coord_curves_ablations}
    \end{subfigure}
    \begin{subfigure}[h]{0.48\linewidth} %
        \centering
        \includegraphics[width=\linewidth]{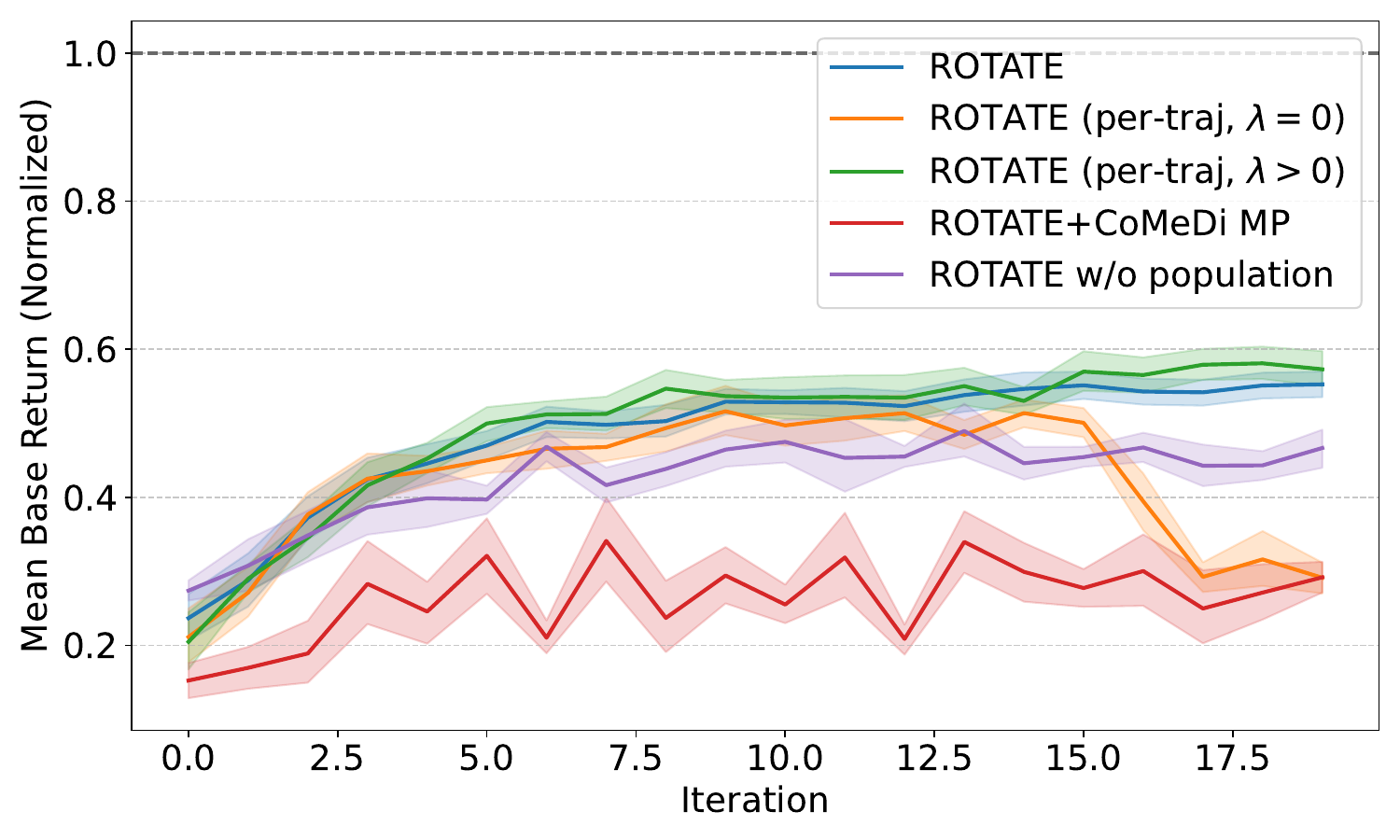}
        \caption{Coordination Ring.}
        \label{fig:coord_ring_curves_ablations}
    \end{subfigure}
    \caption{Learning curves of \ours{} and all variations of \ours{} considered in this paper. Normalized mean returns and stratified bootstrapped 95\% confidence intervals over 3 trials on $\Pieval$ are shown.}
    \label{fig:learning_curves_ablations}
\end{figure}

\vspace{-5pt}
Fig. \ref{fig:learning_curves_ablations} shows learning curves for \ours{} and all \ours{} variations tested in this paper, where the $x$-axis is the open-ended learning iteration, and the $y$-axis corresponds to the mean evaluation return.
On 4/6 tasks (LBF, CR, CC, and FC), \ours{} has better sample efficiency than the variants. On 3/6 tasks (LBF, CR, and FC), \ours{} dominates the variants at almost all points in learning.

\section{Supplemental Discussion}
\label{app:supp_discussion}

Here, we describe how \ours{} satisfies the definition of an open-ended system given by \citet{hughes_position_2024}. 
Next, we compare and contrast CoMeDi's mixed-play mechanism to \ours{}'s per-state regret. 
Finally, we discuss alternative estimators for \ours{} per-state regret.

\subsection{\ours{} as an Open-Ended System}
\label{app:supp_disc:formalize_oel}

\subsubsection{Definition of Open-ended Systems}

To formalize open-endedness, we employ the definition of open-ended systems provided by \citet{hughes_position_2024}. 
\citet{hughes_position_2024} consider a system $S$ that produces a sequence of artifacts $X_t$, indexed by time $t$. An observer $O$ attempts to predict a new artifact $X_T$ based on observations of past artifacts seen up until time $t$.

\begin{quote}
From the perspective of an observer, a system is \emph{open-ended} if and only if the sequence of artifacts it produces is both \underline{novel} and \underline{learnable}.
\end{quote}

A sequence of artifacts is \underline{novel} if it becomes increasingly unpredictable with respect to an observer's model at any fixed time $t$. Formally:
\begin{equation}
    \forall t, \forall T > t, \exists T' > T: \mathbb{E}[\ell(t, T')] > \mathbb{E}[\ell(t, T)].
\end{equation}
This is quantified by a loss metric, $\ell(t, T)$, where the first argument refers to the timestep that the observer has observed up until (i.e., artifacts $X_1, \dots, X_t$), while the second argument refers to the artifact that the observer must now predict (i.e., $X_{T}$).

Conversely, the system is \underline{learnable} whenever conditioning on a longer history makes artifacts more predictable:
\begin{equation}
    \forall T, \forall t < T, \forall T > t' > t: \mathbb{E}[\ell(t', T)] < \mathbb{E}[\ell(t, T)].
\end{equation}

\subsubsection{Open-endedness in ROTATE}

ROTATE displays both novelty and learnability according to the definition above, as we show below. Let the ego agent be the observer, the sequence of generated teammates be the artifacts, and $\ell(t, T)$ be the cooperative regret of the ego agent trained up to iteration $t$, evaluated on the teammate generated at iteration $T$.

\paragraph{Novelty.} At open-ended iteration $t+1$, ROTATE seeks to generate a teammate that \emph{challenges} the ego agent from iteration $t$ by directly maximizing its cooperative regret. By construction, for $T' = T+1$:
\begin{equation}
    \mathbb{E}[\ell(t, T')] > \mathbb{E}[\ell(t, T)].
\end{equation}

\paragraph{Learnability.} In the context of ROTATE, this condition states that for some \emph{future unseen teammate from time $T$}, the cooperative regret of ROTATE's ego agent from iteration $t'$ should be lower than the cooperative regret of ROTATE's ego agent from an earlier iteration $t$.

Intuitively, since all generated teammates are added to a population buffer that is uniformly sampled, ROTATE's ego agent at future iterations should always be more capable than the ego agent at previous iterations. 
Empirically, we observe this in the learning curves presented in Fig.~\ref{fig:learning_curves_ablations}. 
The normalized return against the test set of unseen teammates increases as the iterations increase. Importantly, the return is min-max normalized so that $1.0$ represents the return achieved by a teammate with its best response, while $0.0$ represents a known lower bound on returns. For a fixed teammate $\pi^{-i}$, the upper bound $V(\pi^{-i}, \text{BR}(\pi^{-i}))$ and the lower bound $\eta$ are constant with respect to the ego agent, so the min-max normalized return is an affine, strictly decreasing function of the cooperative regret, $\text{NRet}(t) = 1 - \text{CR}(t)/c$ with $c = V(\pi^{-i}, \text{BR}(\pi^{-i})) - \eta > 0$. An increase in normalized return from iteration $t$ to $t'$ is therefore equivalent to a decrease in cooperative regret over the same interval.

\subsection{Relationship To CoMeDi and Mixed Play}
\label{app:supp_disc:comedi_mp_discussion}

CoMeDi~\citep{sarkar2023diverse} is a two-stage teammate generation AHT algorithm, whose teammate generation process trains one teammate per iteration, with an objective that encourages the new teammates to be distinct from previously discovered teammates. 

CoMeDi adds trained teammate policies to a teammate policy buffer, $\Pi^{\text{train}}$. Each iteration begins by identifying the teammate policy that is most \textit{compatible} with the currently trained teammate $\pi^{-i}$, out of all previously generated policies:
\begin{equation}
    \pi^{\text{comp}} = \underset{\pi^{-j}\in\Pi^{\text{train}}}{\text{argmax }} \mathbb{E}_{s_0 \sim p_{0}} [V(s_0 | \pi^{-i}, \pi^{-j})].
\end{equation}

The new teammate policy $\pi^{-i}$ is trained with an objective that improves the per-trajectory regret objective (Eq.~\ref{eq:traj_regret_obj}) by adding a term that maximizes the returns from states gathered in \textit{mixed-play}, which we describe below. 

Let mixed-play starting states be sampled from states visited when $\pi^{-i}$ interacts with the \textit{mixed} policy, which uniformly samples actions from $\pi^{\text{comp}}$ and $\text{BR}(\pi^{-i})$ at each timestep:
\begin{equation}
    \label{eq:mixed_play_start}
    p_{\text{MSTART}} := d\left(\pi^{-i}, \dfrac{1}{2} \pi^{\text{comp}} + \dfrac{1}{2} \text{BR}(\pi^{-i}); p_0\right).
\end{equation}
From these starting states, CoMeDi then gathers mixed-play interaction data, where $\pi^{-i}$ interacts with $\text{BR}(\pi^{-i})$. The resulting mixed-play state visitation is then expressed as:
\begin{equation}
    \label{eq:mixed_play_data}
    p_{\text{MP}} := d\left(\pi^{-i}, \text{BR}(\pi^{-i}); p_{\text{MSTART}}\right).
\end{equation}
The complete objective that~\citet{sarkar2023diverse} optimizes to train a collection of diverse teammates is then defined as:
\begin{equation}
    \label{eq:mp_regret_obj}
    \underset{\pi}{\max\ } (
    \mathbb{E}_{s_0 \sim p_{0}} \left[\text{CR} (\pi^{\text{comp}}, \pi^{-i}, s_{0})\right] + \underbrace{\mathbb{E}_{s_0 \sim p_{\text{MP}}} [V(s_0 | \pi, \BR(\pi))]}_{\text{mixed-play return maximization}}
    ).
\end{equation}

CoMeDi~\citep{sarkar2023diverse} optimizes this objective to discourage $\pi^{-i}$ from learning poor actions for collaborations outside of $p_{\text{SP}}$. This is because $\pi^{-i}$ is now also trained to maximize returns in states visited during mixed-play, which resembles some states encountered while cooperating with $\pi^{\text{comp}}$. Discerning whether a state is likely encountered while interacting with $\pi^{\text{comp}}$ and consequently choosing to diminish team returns will no longer be an optimal policy to maximize Eq.~\ref{eq:mp_regret_obj}.

Although the importance of using $p_{\text{MSTART}}$ as a starting state for data collection is questionable, we take inspiration from CoMeDi's maximization of $V(s_0 | \pi, \BR(\pi))$ outside of states from $p_{\text{SP}}$. We argue that maximizing $V(s_0 | \pi^{-i}, \BR(\pi^{-i}))$ is a key component towards making $\pi^{-i}$ act in good faith by always choosing actions yielding optimal collective returns assuming $\BR(\pi^{-i})$ is substituted as the partner policy. Unlike CoMeDi, \ours{} maximizes $V(s_0 | \pi^{-i}, \BR(\pi^{-i}))$ on trajectories gathered from a starting state from $\pxp$ (i.e., SXP states) instead of $p_{\text{MSTART}}$, which results in the $\lambda$-weighted return terms of Eq.~\ref{eq:constrained_state_based_regret_obj}. We formulate this objective to encourage $\pi^{-i}$ to act in good faith in states sampled from $\pxp$, which is visited while $\pi^{-i}$ interacts with $\pi^{\text{ego}}$. Since $\pi^{-i}$ is not sabotaging $\piego$ by selecting actions that make collaboration impossible in $\pxp$, the ego policy learning process becomes less challenging. We conjecture that this leads to $\piego$ with better performance, as indicated in Fig.~\ref{fig:rotate_main_results}.

\begin{figure}[t]
    \centering
    \includegraphics[width=0.40\linewidth]{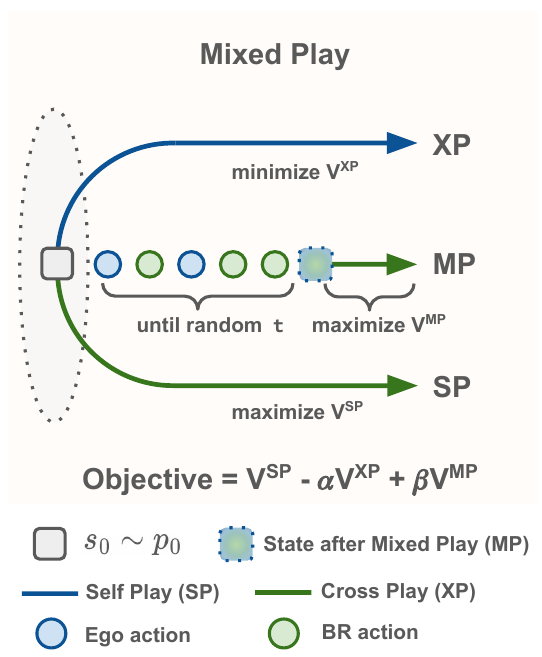}
    \caption{\small CoMeDi-style mixed-play objective for teammate generation, in the context of open-ended AHT.}
    \label{fig:comedi_mp_obj}
\end{figure}

While Fig.~\ref{fig:rotate_main_results} compares ROTATE with CoMeDi, Fig.~\ref{fig:ablations} compares ROTATE with a modified CoMeDi approach that now follows the open-ended training framework described in Algorithm~\ref{alg:OpenEndedFramework}. In this modified version of CoMeDi, we train a newly generated teammate policy to maximize Eq.~\ref{eq:mp_regret_obj} while substituting $\pi^{\text{comp}}$ with the trained $\pi^{\text{ego}}$. Rather than promoting meaningful differences with previously generated teammate policies, this creates a teammate policy that maximizes the ego agent policy's per-trajectory regret while discouraging teammates from diminishing team returns in certain states. This version of CoMeDi's teammate generation objective within the ROTATE open-ended framework is visualized in Fig.~\ref{fig:comedi_mp_obj}.

\section{The Policy Gradient of Per-State Regret}
\label{app:rotate_regret_est}
This section supports Section~\ref{sec:method_teamgen} of the main paper, showing how a teammate policy is trained to maximize the \ours{} per-state regret objective using the standard policy gradient theorem~\citep{sutton1999policy, williams1992simple}. First, the gradient of the cooperative regret objective is expressed as a difference of two standard policy gradients.
Next, the section describes how optimizing this objective differs from standard return maximization.
The section ends by describing gradient estimators that trade off bias and variance.

The full \ours{} teammate generation objective (Eq.~\ref{eq:constrained_state_based_regret_obj}) augments the cooperative regret (CR) term with $\lambda$-weighted teammate return terms that enforce competence. Because those return terms are maximized by a standard policy gradient, the analysis below focuses on the policy gradient for the per-state regret term, and drops the $\lambda$-weighted competence term.

\paragraph{Reduction to standard policy gradients.} Let $p_{\text{opt}}$ denote the state distribution under which cooperative regret is maximized, which corresponds to the mixture $\mathcal{U}(\pxp, \psp)$ used in Eq.~\ref{eq:constrained_state_based_regret_obj}. 
The expected cooperative regret from starting states $s_0\sim p_{\text{opt}}$ is defined as:
\begin{equation}
    \label{eq:coop_regret_obj_apdx}
    \mathbb{E}_{s_0 \sim p_{\text{opt}}} \left[\text{CR} (\piego, \pi^{-i}, s_0)\right] =
    \mathbb{E}_{s_0 \sim p_{\text{opt}}} \left[V(s_0|\pi^{-i}, BR(\pi^{-i})) - V(s_0|\pi^{-i}, \piego)\right],
\end{equation}
where the value functions on the right-hand side are respectively defined as:
\begin{align}
    \label{eq:vals_expanded}
    V(s_0|\pi^{-i}, BR(\pi^{-i})) &=
    \mathbb{E}_{\substack{ a^{-i}_{t}\sim \pi^{-i}, \\ a^{\text{partner}}_{t} \sim \pi^{\text{BR}}, P}}\Bigg[\sum_{t=0}^{\infty} \gamma^{t}R(s_{t}, a_{t})\Bigg| s_{0}\Bigg], \nonumber \\
    V(s_0|\pi^{-i}, \piego) &=
    \mathbb{E}_{\substack{ a^{-i}_{t}\sim \pi^{-i}, \\ a^{\text{partner}}_{t} \sim \pi^{\text{ego}}, P}}\Bigg[\sum_{t=0}^{\infty} \gamma^{t}R(s_{t}, a_{t})\Bigg| s_{0}\Bigg],
\end{align}
where $a^{\text{partner}}_{t}$ denotes the action of the teammate's partner ($BR(\pi^{-i})$ in the first value function, and $\piego$ in the second), and $a_t = (a^{-i}_t, a^{\text{partner}}_t)$ denotes the joint action.

The teammate policy $\pi^{-i}$ is a neural network parameterized by $\theta^{-i}$.
Note that during teammate policy learning, the ego agent policy $\piego$ is fixed. 
We assume that the distribution $p_{\text{opt}}$ and the best-response policy $\pi^{\text{BR}}$ are fixed with respect to $\theta^{-i}$ (justified below). By linearity of the gradient, $\nabla_{\theta^{-i}}$ applied to Eq.~\ref{eq:coop_regret_obj_apdx} splits into a difference of the gradients of two expected returns. Each expected return is an instance of the standard policy gradient setting, in which $\pi^{-i}_{\theta^{-i}}$ is the only learnable policy. Concretely, because the partner policy in each term ($\pi^{\text{BR}}$ in the first term, and $\pi^{\text{ego}}$ in the second) is fixed, the partner can be absorbed into the environment dynamics from the perspective of $\pi^{-i}$, so each term can be treated as the expected return in a single-agent Markov decision process. Applying the policy gradient theorem~\citep{sutton1999policy, williams1992simple} to each term therefore yields:
\begin{align}
    \label{eq:coop_regret_obj_apdx_grad_final}
    \nabla_{\theta^{-i}} \text{ }\mathbb{E}_{s_0 \sim p_{\text{opt}}} \left[\text{CR} (\piego, \pi^{-i}, s_0)\right] = \mathbb{E}_{s_0 \sim p_{\text{opt}}}\Bigg[&\mathbb{E}_{\tau \sim \mathbb{P}_{\text{SP}}(\cdot|s_0)}\Bigg[\sum_{j=0}^{\infty} \nabla_{\theta^{-i}}\text{log}\left(\pi^{-i}_{\theta^{-i}}(a^{-i}_{j}|s_j)\right)F^\gamma_{j:\infty}(\tau)\Bigg]\nonumber \\ -&\mathbb{E}_{\tau \sim \mathbb{P}_{\text{XP}}(\cdot|s_0)}\Bigg[\sum_{j=0}^{\infty} \nabla_{\theta^{-i}}\text{log}\left(\pi^{-i}_{\theta^{-i}}(a^{-i}_{j}|s_j)\right)F^\gamma_{j:\infty}(\tau)\Bigg]\Bigg],
\end{align}
where $\tau=\langle(s_t,a^{-i}_{t},a^{\text{partner}}_{t})\rangle_{t=0}^{\infty}$ denotes an interaction trajectory between the teammate and a partner, and $F^\gamma_{a:b}(\tau) = \sum_{t=a}^{b} \gamma^{t}R(s_{t}, a_{t})$ denotes the discounted return accumulated between timesteps $a$ and $b$ along trajectory $\tau$. The distributions $\mathbb{P}_{\text{SP}}(\cdot|s_0)$ and $\mathbb{P}_{\text{XP}}(\cdot|s_0)$ are distributions over trajectories that begin at the given starting state $s_0$, generated by rolling out $\pi^{-i}$ alongside $\pi^{\text{BR}}$ (self-play) and $\pi^{\text{ego}}$ (cross-play), respectively:
\begin{align}
    \label{eq:traj_dists}
   \mathbb{P}_{\text{SP}}(\tau|s_0) &= \prod_{t=0}^{\infty}\left(\pi^{-i}(a^{-i}_{t}|s_t)\,\pi^{\text{BR}}(a^{\text{BR}}_{t}|s_t)\, P(s_{t+1}|s_{t}, a^{-i}_{t}, a^{\text{BR}}_{t})\right), \nonumber\\
   \mathbb{P}_{\text{XP}}(\tau|s_0) &= \prod_{t=0}^{\infty}\left(\pi^{-i}(a^{-i}_{t}|s_t)\,\pi^{\text{ego}}(a^{\text{ego}}_{t}|s_t)\, P(s_{t+1}|s_{t}, a^{-i}_{t}, a^{\text{ego}}_{t})\right).
\end{align}
Note that $\mathbb{P}_{\text{SP}}$ and $\mathbb{P}_{\text{XP}}$ are distributions over trajectories, distinct from the state-visitation distributions $\psp$ and $\pxp$ defined in Section~\ref{sec:method_teamgen}.

\paragraph{Estimating the policy gradient of per-state regret.} Exactly evaluating Eq.~\ref{eq:coop_regret_obj_apdx_grad_final} is intractable in most applications, because it is often not possible to evaluate $\mathbb{P}_{\text{SP}}(\tau|s_0)$ and $\mathbb{P}_{\text{XP}}(\tau|s_0)$ for all plausible trajectories. 
The gradient is therefore estimated from samples using the following three steps. 
First, sample starting states from the distribution under which regret is maximized, $s_0\sim p_{\text{opt}}$. Second, sample trajectories to estimate the two expectation terms on the right-hand side of Eq.~\ref{eq:coop_regret_obj_apdx_grad_final}. 
The first term is estimated by rolling out $\pi^{-i}$ with $\text{BR}(\pi^{-i})$ from $s_0$ to sample from $\mathbb{P}_{\text{SP}}(\cdot|s_0)$, while the second is estimated by rolling out $\pi^{-i}$ with $\pi^{\text{ego}}$ from $s_0$ to sample from $\mathbb{P}_{\text{XP}}(\cdot|s_0)$. 
Third, the log-likelihood and return terms of Eq.~\ref{eq:coop_regret_obj_apdx_grad_final} are evaluated on the sampled trajectories.

The error of this estimator, relative to the exact gradient, is governed by its bias and variance, which are controlled by the choice of return estimator for $F^\gamma_{j:\infty}(\tau)$. 
Standard choices span the bias-variance spectrum, from the unbiased but high-variance Monte Carlo return-to-go, to the high-bias but low-variance value or advantage network estimators, to the generalized advantage estimator (GAE)~\citep{Schulman2015HighDimensionalCC}, whose $\lambda$ parameter interpolates between these extremes. 
\ours{} relies on the GAE to estimate the per-state regret, as described in App.~\ref{app:rotate_algorithm_description}.

\paragraph{Assumptions.}
The derivation of Eq.~\ref{eq:coop_regret_obj_apdx_grad_final} treats two quantities as fixed with respect to $\theta^{-i}$ in order to apply the standard policy gradient theorem, although neither is truly independent of it: the best-response policy $\pi^\text{BR}$ and the starting-state distribution $p_{\text{opt}}$.
Each assumption is justified below.

Regarding treating $\pi^{\text{BR}}$ as fixed with respect to $\theta^{-i}$: the exact best response depends on $\theta^{-i}$ through the best-response operator, but is treated as fixed at each update because it is approximated by a separately trained network (as described in App.~\ref{app:rotate_algorithm_description}). 
Consequently, Eq.~\ref{eq:coop_regret_obj_apdx_grad_final} omits the gradient flowing through the best response, introducing bias relative to the gradient of the true regret. 
Nonetheless, treating the other agents' policies as fixed during each update is the standard assumption underlying independent learning algorithms in multi-agent reinforcement learning, such as independent PPO (IPPO), which achieves strong empirical performance in cooperative tasks~\citep{yu_surprising_2022}. 
\citet{sun2023trustregion} provide theoretical support for this practice, proving that despite the non-stationarity each agent faces when the other agents' policies update concurrently, a monotonic improvement guarantee is retained as long as every agent's policy update remains within a trust region of the joint policy, a constraint that PPO's ratio clipping enforces. 
Because \ours{} trains the teammate and best-response policies with PPO losses (Algorithm~\ref{alg:rotate_teamgen_alg}), the same mechanism is expected to mitigate the non-stationarity introduced by the concurrently trained $\pi^{\text{BR}}$, although the formal guarantee of \citet{sun2023trustregion} is established for a shared-return objective rather than for regret.

Regarding treating $p_{\text{opt}}$ as fixed with respect to $\theta^{-i}$: the starting-state distribution $p_{\text{opt}}$ is the key difference between per-state regret maximization and standard return maximization. In the standard setting, trajectories begin at the environment's initial state distribution $p_0$, which does not depend on the learned policy. By contrast, $p_{\text{opt}} = \mathcal{U}(\pxp, \psp)$ is a mixture of the XP and SP state-visitation distributions, both of which are induced by $\pi^{-i}_{\theta^{-i}}$ and therefore shift whenever $\theta^{-i}$ is updated. The objective is thus self-referential, because $\theta^{-i}$ appears both in the regret being averaged and in the distribution the average is taken over. Compared to Eq.~\ref{eq:coop_regret_obj_apdx_grad_final}, the exact gradient of this objective contains an additional term capturing the sensitivity of $p_{\text{opt}}$ to $\theta^{-i}$.

Although Eq.~\ref{eq:coop_regret_obj_apdx_grad_final} is not the exact gradient of the per-state regret objective, it may be understood as the exact gradient of a surrogate objective in which $p_{\text{opt}}$ is frozen at the distribution induced by $\theta^{-i}$ at the most recent data-collection step. 
Computing the gradient with respect to a frozen distribution is a common technique in reinforcement learning to simplify the computation of gradients and sometimes stabilize the optimization process.
In particular, this idea appears in semi-gradient temporal-difference methods~\citep{sutton2018reinforcement}, target networks in deep reinforcement learning~\citep{mnih2015dqn}, and the PPO objective~\citep{Schulman2017ProximalPO}. Among these examples, PPO is the closest analogue, because it likewise evaluates its loss on states and actions drawn from a policy-induced distribution frozen at the parameters from the previous data-collection step.
Consequently, the trust-region reasoning applied above to $\pi^{\text{BR}}$ is expected to extend to \ours{}'s treatment of $p_{\text{opt}}$. Because each update changes $\theta^{-i}$ by a bounded amount, and fresh XP and SP data are collected before every round of updates, the frozen distribution is expected to remain close to the $p_{\text{opt}}$ distribution that the current $\theta^{-i}$ induces.

\section{ROTATE Pseudocode}
\label{app:algorithm_pseudocode}

This section presents the pseudocode for \ours{}, which follows the open-ended framework specified by \cref{alg:OpenEndedFramework}.
Section~\ref{sec:oeaht} described an open-ended training framework for training an ego agent that can effectively collaborate with previously unseen teammates, which is detailed in \cref{alg:OpenEndedFramework}. 
In the open-ended training framework, 
a \textbf{TeammateGenerator} function determines a buffer of teammate policy parameters, $\text{B}^{\text{new}}_{\pi}$, that acts as support for the distribution from which teammate policy parameters are sampled.
The teammate generator function considers the ego agent's current policy parameters, $\theta^{\text{ego}}$, and the previous buffer of teammate policy parameters, $\text{B}_{\pi}$.
Ideally, the teammate generation function generates and samples teammates that induce learning challenges for $\pi^{\text{ego}}$. 
An \textbf{EgoUpdate} function specifies a procedure that updates the ego agent's policy parameters based on the $\text{B}^{\text{new}}_{\pi}$ designed by the teammate generator.

\subsection{ROTATE Teammate Generation Algorithm}
\label{app:rotate_algorithm_description}
\ours{}'s teammate generation algorithm is detailed in Algorithm~\ref{alg:rotate_teamgen_alg}.
As described in Section~\ref{sec:method_teamgen}, this teammate generation algorithm jointly trains the parameters of a teammate policy and an estimate of its best response (BR) policy, based on a provided ego agent policy.
The two policies optimize distinct objectives.
The BR policy performs standard return maximization, while the teammate policy follows the sampled regret policy gradient derived in App.~\ref{app:rotate_regret_est}, augmented with the $\lambda$-weighted competence terms of Eq.~\ref{eq:constrained_state_based_regret_obj}.
Accordingly, this subsection first describes the initialization and data collection steps shared by both policies, then defines the three PPO loss functions from which the updates are constructed, and finally presents the BR policy, teammate policy, and critic updates separately.

\begin{algorithm}[htbp]\setstretch{1.15}
\caption{\ours{} TeammateGenerator Function}
\label{alg:rotate_teamgen_alg}
\begin{algorithmic}[1]
\STATE {\bfseries Input:} Environment $\text{Env}$, ego agent policy $\pi_{\theta^{\text{ego}}}$, teammate policy parameter buffer $\text{B}_{\pi}$,
number of updates $N_{\text{updates}}$, PPO clipping parameter $\epsilon$, PPO update epochs $N_{\text{epochs}}$, competence weights $\lambda_1, \lambda_2$.
\STATE {\bfseries Output:} Updated teammate buffer $\text{B}_{\pi}$.

\STATE $\theta^{-i}, \theta^{\text{BR}} \leftarrow \textbf{RandomInit}(\pi),\ \textbf{RandomInit}(\pi)$ \label{alg:rotate-tg:init-policy}
\STATE $\sigma^{\text{BR}} \leftarrow \textbf{RandomInit}(V)$ \label{alg:rotate-tg:init-critic-br}
\STATE $\sigma^{-i,\text{BR}}, \sigma^{-i,\text{ego}} \leftarrow \textbf{RandomInit}(V),\ \textbf{RandomInit}(V)$
\algorithmiccomment{Init teammate and BR critic parameters} \label{alg:rotate-tg:init-critic-tm}

\FOR{$t_{\text{update}}=1$ {\bfseries to} $N_{\text{updates}}$}
    \STATE $D_{\text{SP}} \leftarrow \textbf{Interact}(\pi_{\theta^{\text{BR}}}, \pi_{\theta^{-i}}, p_{0}^{\text{Env}})$
    \STATE $D_{\text{XP}} \leftarrow \textbf{Interact}(\pi_{\theta^{\text{ego}}}, \pi_{\theta^{-i}}, p_{0}^{\text{Env}})$
    \label{alg:rotate-tg:collect-sp-xp}
    \STATE $s_{\text{XP}} \leftarrow \textbf{SampleStates}(D_{\text{XP}})$
    \STATE $s_{\text{SP}} \leftarrow \textbf{SampleStates}(D_{\text{SP}})$
    \algorithmiccomment{Sample XP/SP states} \label{alg:rotate-tg:sample-xp-states}
    \STATE $D_{\text{SXP}} \leftarrow \textbf{Interact}(\pi_{\theta^{\text{BR}}}, \pi_{\theta^{-i}}, \mathcal{U}(s_{\text{XP}}))$ \label{alg:rotate-tg:collect-sxp}
    \STATE $D_{\text{XSP}} \leftarrow \textbf{Interact}(\pi_{\theta^{\text{ego}}}, \pi_{\theta^{-i}}, \mathcal{U}(s_{\text{SP}}))$
    \algorithmiccomment{Gather SP, XP, SXP, and XSP data} \label{alg:rotate-tg:collect-xsp}

    \STATE $\theta^{\text{BR}}_{\text{old}}, \theta^{-i}_{\text{old}} \leftarrow \theta^{\text{BR}}, \theta^{-i}$
    \STATE $\sigma^{\text{BR}}_{\text{old}}, \sigma^{-i,\text{BR}}_{\text{old}}, \sigma^{-i,\text{ego}}_{\text{old}} \leftarrow
    \sigma^{\text{BR}}, \sigma^{-i,\text{BR}}, \sigma^{-i,\text{ego}}$
    \algorithmiccomment{Store old model parameters.}

    \FOR{$k_{\text{update}}=1$ {\bfseries to} $N_{\text{epochs}}$}
        \STATE $\mathcal{L}_{\text{BR}}(\theta^{\text{BR}}) \leftarrow \text{Eq.}~\ref{eq:br_loss}$
        \algorithmiccomment{BR policy loss: standard PPO loss on $D_{\text{SP}} \cup D_{\text{SXP}}$}
        \label{alg:rotate-tg:pol-loss-br}
        \STATE $\mathcal{L}_{\text{REG}}(\theta^{-i}) \leftarrow \text{Eq.}~\ref{eq:regret_pol_loss}$
        \algorithmiccomment{Teammate policy loss: per-state regret, competence, and entropy}
        \label{alg:rotate-tg:regret-policy-loss}

        \STATE $L_{V}(\sigma^{\text{BR}}) \leftarrow \mathcal{L}_{\text{VAL}}\left(
        \sigma^{\text{BR}}, \sigma^{\text{BR}}_{\text{old}}, D_{\text{SP}} \cup D_{\text{SXP}}\right)$
        \label{alg:rotate-tg:val-loss-br}
        \STATE $L_{V}(\sigma^{-i,\text{BR}}) \leftarrow \mathcal{L}_{\text{VAL}}\left(
        \sigma^{-i,\text{BR}}, \sigma^{-i,\text{BR}}_{\text{old}}, D_{\text{SP}} \cup D_{\text{SXP}}\right)$
        \label{alg:rotate-tg:val-loss-teammate}
        \STATE $L_{V}(\sigma^{-i,\text{ego}}) \leftarrow \mathcal{L}_{\text{VAL}}\left(
        \sigma^{-i,\text{ego}}, \sigma^{-i,\text{ego}}_{\text{old}}, D_{\text{XP}} \cup D_{\text{XSP}}\right)$
        \label{alg:rotate-tg:val-loss-ego}

        \STATE $\theta^{\text{BR}} \leftarrow \textbf{GradDesc}\left(\theta^{\text{BR}}, \nabla_{\theta^{\text{BR}}} \mathcal{L}_{\text{BR}}(\theta^{\text{BR}})\right)$
        \label{alg:rotate-tg:br-pol-gd}
        \STATE $\theta^{-i} \leftarrow \textbf{GradDesc}\left(\theta^{-i}, \nabla_{\theta^{-i}} \mathcal{L}_{\text{REG}}(\theta^{-i})\right)$
        \algorithmiccomment{Update policies} \label{alg:rotate-tg:tm-pol-gd}

        \STATE $\sigma^{\text{BR}} \leftarrow \textbf{GradDesc}\left(\sigma^{\text{BR}}, \nabla_{\sigma^{\text{BR}}} L_{V}(\sigma^{\text{BR}})\right)$
        \STATE $\sigma^{-i,\text{BR}} \leftarrow \textbf{GradDesc}\left(\sigma^{-i,\text{BR}}, \nabla_{\sigma^{-i,\text{BR}}} L_{V}(\sigma^{-i,\text{BR}})\right)$
        \STATE $\sigma^{-i,\text{ego}} \leftarrow \textbf{GradDesc}\left(\sigma^{-i,\text{ego}}, \nabla_{\sigma^{-i,\text{ego}}} L_{V}(\sigma^{-i,\text{ego}})\right)$
        \algorithmiccomment{Update critics.} \label{alg:rotate-tg:critic-loss-ego}
    \ENDFOR
\ENDFOR

\STATE $\text{B}_{\pi} \leftarrow \text{B}_{\pi} \cup \langle \theta^{-i} \rangle$
\algorithmiccomment{Add generated teammate policy parameter} \label{alg:rotate-tg:add_teammate_to_buffer}
\STATE {\bfseries Return:} $\text{B}_{\pi}$ \label{alg:rotate-tg:return-statement}
\end{algorithmic}
\end{algorithm}

\paragraph{Initialization of policy and value function networks.}
The parameters of the teammate and BR policies, $\theta^{-i}$ and $\theta^\text{BR}$, are initialized in \cref{alg:rotate-tg:init-policy}. The parameters of the BR critic network, $\sigma^\text{BR}$,  are initialized in \cref{alg:rotate-tg:init-critic-br}, while those for the teammate, $\sigma^{-i, \text{BR}}$ and $\sigma^{-i, \text{ego}}$, are initialized in \cref{alg:rotate-tg:init-critic-tm}. 
Note that the teammate maintains two critics, for separately estimating returns when interacting with the BR and ego agent policies. 

\paragraph{Data collection.}
The training of the teammate and BR policies is based on the SP, XP, XSP, and SXP interaction data gathered in \cref{alg:rotate-tg:collect-sp-xp,alg:rotate-tg:sample-xp-states,alg:rotate-tg:collect-sxp,alg:rotate-tg:collect-xsp}, which Section~\ref{sec:method_teamgen} motivates and describes.
Recall that an SXP interaction requires resetting an environment to start from an available XP state, and an XSP interaction analogously requires resetting to an SP state.
Since resetting from all available XP states for SXP interaction is impractical, \ours{} \textit{samples} from XP states to obtain start states for SXP interactions (and similarly for XSP).
Experiences from SP, XP, SXP, and XSP interaction are stored in buffers $D_{\text{SP}}, D_{\text{XP}}, D_{\text{SXP}}, D_{\text{XSP}}$ in the form of a collection of tuples, $D = \langle(s_{k}, a_{k}, r_k, s'_k)\rangle_{k=1}^{|D|}$.

Note that the data collection lines realize the first two steps of the regret policy gradient estimation procedure described in App.~\ref{app:rotate_regret_est}.
Specifically, \cref{alg:rotate-tg:sample-xp-states} realizes the first step by sampling start states from the recently collected buffers, thereby freezing the start-state distribution $p_{\text{opt}}$ at the distribution induced by the current policies, as discussed in the assumptions paragraph of App.~\ref{app:rotate_regret_est}.
\cref{alg:rotate-tg:collect-sxp,alg:rotate-tg:collect-xsp} realize the second step by rolling out the teammate policy alongside the BR and ego agent policies from the sampled states.
Finally, Lines \ref{alg:rotate-tg:pol-loss-br} to \ref{alg:rotate-tg:critic-loss-ego} realize the third step by evaluating the policy and critic losses on the stored experiences, as detailed in the remainder of this subsection.

\paragraph{Generalized PPO loss functions.}
The policy and critic updates of Algorithm~\ref{alg:rotate_teamgen_alg} are constructed from three loss functions: a generalized PPO clip loss, $\mathcal{L}_{\text{CLIP}}$, an entropy loss, $\mathcal{L}_{\text{ENT}}$, and a value function loss, $\mathcal{L}_{\text{VAL}}$.
The clip loss receives $\left(\theta, \theta_{\text{old}},\sigma_{\text{old}}, D, \epsilon, w, \varsigma \right)$ as input and evaluates
\begin{equation}
\label{eq:ppo_clip_loss}
\begin{split}
    &\mathcal{L}_{\text{CLIP}}\left(\theta, \theta_{\text{old}},\sigma_{\text{old}}, D, \epsilon, w, \varsigma \right) = \\
    &\quad\underset{\left(s_t, a_t, r_t, s_{t+1}\right)\in D}{\mathbb{E}}  \left[-\text{min}\left(\dfrac{\pi_{\theta}(a_t|s_t)}{\pi_{\theta_{\text{old}}}(a_t|s_t)}\varsigma w_t A_t, \text{clip}\left(\dfrac{\pi_{\theta}(a_t|s_t)}{\pi_{\theta_{\text{old}}}(a_t|s_t)}, 1 - \epsilon, 1 + \epsilon \right)\varsigma w_t A_t\right)\right],
\end{split}
\end{equation}
where $A_t$ denotes an estimate of the \textit{advantage} function computed from the critic parameterized by $\sigma_{\text{old}}$ via the Generalized Advantage Estimation (GAE) algorithm~\citep{Schulman2015HighDimensionalCC}.
The final two arguments generalize the standard PPO clip loss and are introduced solely to express the teammate update: $w_t$ is a per-transition weight applied to the advantage, through which the teammate update implements the per-state regret estimator, and $\varsigma \in \{+1, -1\}$ is a sign applied to the advantage, through which the teammate update implements the subtraction within the regret policy gradient.
The entropy loss receives $\left(\theta, D\right)$ as input and evaluates
\begin{equation}
\label{eq:ent_loss}
\mathcal{L}_{\text{ENT}}\left(\theta, D\right) = \underset{\left(s_t, a_t, r_t, s_{t+1}\right)\in D}{\mathbb{E}} \left[\pi_{\theta}(a_t|s_t) \text{log}\left(\pi_{\theta}(a_t|s_t)\right)\right],
\end{equation}
whose minimization increases policy entropy.
With unit weights ($w = \mathbf{1}$) and a positive sign ($\varsigma = +1$), the sum $\mathcal{L}_{\text{CLIP}} + \mathcal{L}_{\text{ENT}}$ is exactly the standard PPO loss~\citep{Schulman2017ProximalPO}, whose minimization maximizes returns while encouraging sufficient exploration.
Finally, the value function loss trains a critic network parameterized by $\sigma$.
The function receives $(\sigma, \sigma_{\text{old}}, D)$ as input and computes the standard mean squared Bellman error (MSBE) loss:
\begin{equation}
\label{eq:val_loss}
     \mathcal{L}_{\text{VAL}}\left(\sigma, \sigma_{\text{old}}, D\right) = \underset{\left(s, a, r', s'\right)\in D}{\mathbb{E}}\left[\left(V_{\sigma}(s) - V^\text{targ}_{\sigma_\text{old}}(s)
     \right)^{2}\right],
\end{equation}
where $V^\text{targ}_{\sigma_\text{old}}(s) := A^\text{GAE}_{\sigma_\text{old}} + V_{\sigma_\text{old}}(s)$ is the target value estimate.

\paragraph{Best response policy update.}
The BR policy is trained by standard return maximization.
Accordingly, \cref{alg:rotate-tg:pol-loss-br} computes the BR loss by instantiating the standard PPO loss with unit weights and a positive sign:
\begin{equation}
\label{eq:br_loss}
\mathcal{L}_{\text{BR}}(\theta^{\text{BR}}) = \mathcal{L}_{\text{CLIP}}\left(\theta^{\text{BR}}, \theta^{\text{BR}}_{\text{old}}, \sigma^{\text{BR}}_{\text{old}}, D_{\text{SP}} \cup D_{\text{SXP}}, \epsilon, \mathbf{1}, +1\right) + \mathcal{L}_{\text{ENT}}\left(\theta^{\text{BR}}, D_{\text{SP}} \cup D_{\text{SXP}}\right).
\end{equation}
The loss is evaluated on the merged dataset $D_{\text{SP}} \cup D_{\text{SXP}}$, with advantages estimated by the BR critic $\sigma^{\text{BR}}$.
Including the SXP data ensures that the BR policy learns to maximize returns not only from initially reachable states, but also from states visited by the ego agent, so that the BR policy remains a meaningful best response on the start-state distribution over which regret is estimated.

\paragraph{Teammate policy update.}
The teammate policy update instantiates the sampled estimator of the regret policy gradient (Eq.~\ref{eq:coop_regret_obj_apdx_grad_final}), augmented with the competence terms of Eq.~\ref{eq:constrained_state_based_regret_obj}.
\cref{alg:rotate-tg:regret-policy-loss} computes the teammate loss $\mathcal{L}_{\text{REG}}$, which combines six instances of the clip loss $\mathcal{L}_{\text{CLIP}}$ with a single entropy loss:
\begin{equation}
\label{eq:regret_pol_loss}
\begin{split}
&\mathcal{L}_{\text{REG}}(\theta^{-i}) = \\
&\quad\underbrace{\ell\left(D_{\text{SP}}, \sigma^{-i,\text{BR}}_{\text{old}}, w_{\text{ts}}, +1\right) + \ell\left(D_{\text{SXP}}, \sigma^{-i,\text{BR}}_{\text{old}}, \mathbf{1}, +1\right)}_{\text{SP side of the regret gradient (Eq.~\ref{eq:coop_regret_obj_apdx_grad_final})}} \\
&\quad+ \underbrace{\ell\left(D_{\text{XP}}, \sigma^{-i,\text{ego}}_{\text{old}}, w_{\text{ts}}, -1\right) + \ell\left(D_{\text{XSP}}, \sigma^{-i,\text{ego}}_{\text{old}}, \mathbf{1}, -1\right)}_{\text{XP side of the regret gradient (Eq.~\ref{eq:coop_regret_obj_apdx_grad_final})}} \\
&\quad+ \underbrace{\lambda_2\, \ell\left(D_{\text{SP}}, \sigma^{-i,\text{BR}}_{\text{old}}, w_{\text{ts}}, +1\right) + \lambda_1\, \ell\left(D_{\text{SXP}}, \sigma^{-i,\text{BR}}_{\text{old}}, \mathbf{1}, +1\right)}_{\text{competence terms of Eq.~\ref{eq:constrained_state_based_regret_obj}}} \\
&\quad+ \mathcal{L}_{\text{ENT}}\left(\theta^{-i}, D_{\text{SP}} \cup D_{\text{XP}} \cup D_{\text{SXP}} \cup D_{\text{XSP}}\right),
\end{split}
\end{equation}
where $\ell\left(D, \sigma, w, \varsigma\right)$ abbreviates $\mathcal{L}_{\text{CLIP}}\left(\theta^{-i}, \theta^{-i}_{\text{old}}, \sigma, D, \epsilon, w, \varsigma\right)$, and the timestep weight $w_{\text{ts}}$ is defined below.
The first four terms of Eq.~\ref{eq:regret_pol_loss} form a surrogate loss whose gradient is the sampled regret policy gradient of Eq.~\ref{eq:coop_regret_obj_apdx_grad_final}, with two standard substitutions.
First, the GAE replaces the return estimate $F^\gamma_{j:\infty}(\tau)$, trading variance for bias as discussed in App.~\ref{app:rotate_regret_est}.
Second, the PPO clipped surrogate replaces the vanilla score-function estimator, encouraging each update to remain within a trust region~\citep{Schulman2017ProximalPO}.

The structure of the four regret terms mirrors the structure of Eq.~\ref{eq:coop_regret_obj_apdx_grad_final}, which expresses the regret gradient as the difference of two standard policy gradients, namely an SP side, in which the teammate is partnered with $\pi^{\text{BR}}$, and an XP side, in which the teammate is partnered with $\piego$.
The sign argument $\varsigma$ carries the minus sign of this difference.
Because the same teammate log-probability ratio multiplies the return estimates of both terms of Eq.~\ref{eq:coop_regret_obj_apdx_grad_final}, negating the advantages of the XP-side terms ($\varsigma = -1$) implements the subtraction of the XP-side policy gradient.
Consistent with this decomposition, the XP-side terms use advantages estimated by the ego-interaction critic $\sigma^{-i,\text{ego}}$, while the SP-side terms use advantages estimated by the BR-interaction critic $\sigma^{-i,\text{BR}}$.
Furthermore, each side of the regret gradient is evaluated over start states drawn from the mixture $p_{\text{opt}} = \mathcal{U}(\pxp, \psp)$, so each side contributes one loss term per start-state distribution, yielding the four regret terms of Eq.~\ref{eq:regret_pol_loss}.
Table~\ref{tab:regret_loss_terms} summarizes this correspondence.

\begin{table}[htbp]
\centering
\small
\caption{\small Correspondence between the four regret terms of Eq.~\ref{eq:regret_pol_loss} and the two sides of the regret policy gradient (Eq.~\ref{eq:coop_regret_obj_apdx_grad_final}), evaluated over the two start-state distributions composing $p_{\text{opt}}$.}
\label{tab:regret_loss_terms}
\begin{tabular}{lcc}
\toprule
 & Start states $\sim \pxp$ & Start states $\sim \psp$ \\
\midrule
SP side ($\varsigma = +1$, partner $\pi^{\text{BR}}$, critic $\sigma^{-i,\text{BR}}$) & $D_{\text{SXP}}$, weight $\mathbf{1}$ & $D_{\text{SP}}$, weight $w_{\text{ts}}$ \\
XP side ($\varsigma = -1$, partner $\piego$, critic $\sigma^{-i,\text{ego}}$) & $D_{\text{XP}}$, weight $w_{\text{ts}}$ & $D_{\text{XSP}}$, weight $\mathbf{1}$ \\
\bottomrule
\end{tabular}
\end{table}

The dataset and weight in each cell of Table~\ref{tab:regret_loss_terms} follow from how the corresponding expectation is estimated.
We use two different strategies for estimating the diagonal and off-diagonal cells of the table; each is described below. 

Consider first the two off-diagonal cells, in which the rollout partner matches the origin of the start-state distribution: the XP side evaluated over start states drawn from $\pxp$, and the SP side evaluated over start states drawn from $\psp$.
Focusing on the XP cell, the estimated quantity corresponds to the XP side of Eq.~\ref{eq:coop_regret_obj_apdx_grad_final}, restricted to the $\pxp$ half of the mixture $p_{\text{opt}}$:
\begin{equation}
\label{eq:xp_side_pxp}
g_{\text{XP}} := -\mathbb{E}_{s_0 \sim \pxp}\, \mathbb{E}_{\tau \sim \mathbb{P}_{\text{XP}}(\cdot \mid s_0)}\Bigg[\sum_{j=0}^{\infty} \nabla_{\theta^{-i}}\text{log}\left(\pi^{-i}_{\theta^{-i}}(a^{-i}_{j} \mid s_j)\right)F^\gamma_{j:\infty}(\tau)\Bigg].
\end{equation}
Estimating $g_{\text{XP}}$ requires start states sampled from $\pxp$ and, for each sampled start state, an XP rollout continuing from that state.
The dataset $D_{\text{XP}}$ supplies both requirements at once.
First, by the definition of the state-visitation distribution, the state $s_t$ visited at the $t$-th timestep of an XP episode is a sample from $\pxp$.
Second, because the episode continues from $s_t$ under the same partner $\piego$, the remainder of the episode serves as the rollout for $s_t$, so no additional rollouts are needed.
Treating each visited state $s_t$ of an episode as a start-state sample may therefore contribute one copy of the sum inside Eq.~\ref{eq:xp_side_pxp}, taken over the transitions at timesteps $j \geq t$ of that episode.
Consequently, averaging these copies over the start-state samples replaces the two expectations of Eq.~\ref{eq:xp_side_pxp} with a double sum of terms, $\sum_{t} \sum_{j \geq t}$, in which the outer sum ranges over the start-state samples $s_t$ and the inner sum ranges over the transitions of the rollout beginning at $s_t$. Replacing the return estimate $F^\gamma_{j:\infty}(\tau)$ with the GAE, the resulting estimator of $g_{\text{XP}}$ is:

\begin{equation}
\label{eq:xp_side_pxp_estimator}
\hat{g}_{\text{XP}} = - \sum_{t=0}^\infty \sum_{j = t}^\infty \Bigg[ \nabla_{\theta^{-i}}\text{log}\left(\pi^{-i}_{\theta^{-i}}(a^{-i}_{j} \mid s_j)\right)A^{\text{GAE}}_{\sigma^{-i,\text{ego}}_{\text{old}}}(s_j, a_j)\Bigg].
\end{equation}

Based on the observation that the $t$th transition in an XP trajectory appears $t+1$ times in Eq.~\ref{eq:xp_side_pxp_estimator}, and leveraging the finite-horizon nature of the experimental tasks, we can collect terms to express the estimator as a single sum over timesteps, in which each transition is weighted by its timestep:
\begin{equation}
    \label{eq:tsw_grad}
    \hat{g}_{\text{XP}} \approx -\underset{\tau \sim D_{\text{XP}}}{\mathbb{E}} \left[\sum_{t=0}^\infty (t+1) \cdot A^{\text{GAE}}_{\sigma^{-i,\text{ego}}_{\text{old}}}(s_t, a_t) \, \nabla_{\theta^{-i}} \log \pi_{\theta^{-i}}(a_t \mid s_t)\right].
\end{equation}
Thus, for each zero-indexed transition, the weight $w_{\text{ts}}$ is set to $t+1$.
The SP-side cell on $D_{\text{SP}}$ follows analogously, with $\pi^{\text{BR}}$ as the rollout partner and a positive sign.

In contrast, estimating a side of the regret gradient on start states drawn from the \textit{other} side's visitation distribution requires fresh rollouts with the corresponding partner, which is precisely the role of the SXP and XSP data.
Each such branch rollout estimates the value of only its single start state, so the $D_{\text{SXP}}$ and $D_{\text{XSP}}$ terms receive unit weight $\mathbf{1}$.

The two $\lambda$-weighted terms of Eq.~\ref{eq:regret_pol_loss} maximize the $\lambda$-weighted returns of the teammate when partnered with the BR policy, enforcing the competence constraint of Eq.~\ref{eq:constrained_state_based_regret_obj}.
Because these competence terms coincide with the SP-side regret terms up to their coefficients, collecting terms shows that the $D_{\text{SP}}$ and $D_{\text{SXP}}$ losses enter Eq.~\ref{eq:regret_pol_loss} with total coefficients $(1 + \lambda_2)$ and $(1 + \lambda_1)$, in which the unit portion belongs to the regret surrogate and the $\lambda$ portion enforces competence.
Finally, the entropy loss enters $\mathcal{L}_{\text{REG}}$ exactly once, evaluated over all four interaction datasets, rather than once per $\mathcal{L}_{\text{CLIP}}$ instance.
\cref{alg:rotate-tg:tm-pol-gd} then updates the teammate policy parameters via applying gradient descent on Eq.~\ref{eq:regret_pol_loss}.

\paragraph{Value function updates.}
\cref{alg:rotate-tg:val-loss-br,alg:rotate-tg:val-loss-teammate,alg:rotate-tg:val-loss-ego} train the three critic networks by evaluating the value function loss $\mathcal{L}_{\text{VAL}}$ of Eq.~\ref{eq:val_loss}, which measures returns from the interaction between the generated teammate policy and its best response or ego agent policy.
Each critic is trained on the same interaction data for which it supplies advantage estimates in the policy losses.
Specifically, the BR critic $\sigma^{\text{BR}}$ and the teammate's BR-interaction critic $\sigma^{-i,\text{BR}}$ are trained on $D_{\text{SP}} \cup D_{\text{SXP}}$, while the teammate's ego-interaction critic $\sigma^{-i,\text{ego}}$ is trained on $D_{\text{XP}} \cup D_{\text{XSP}}$.

The previously defined loss functions can be minimized using any gradient descent-based optimization technique, as we indicate in Lines \ref{alg:rotate-tg:br-pol-gd} to \ref{alg:rotate-tg:critic-loss-ego}. 
In practice, our implementation uses the Adam optimization technique~\citep{kingma_adam_2017}. 
At the end of this teammate generation process, \cref{alg:rotate-tg:return-statement,alg:rotate-tg:add_teammate_to_buffer} indicate how the generated teammate policy parameter is added to a storage buffer, which is subsequently uniformly sampled to provide teammate policies for ego agent training.

\subsection{ROTATE Ego Agent Update Algorithm}
\label{app:rotate_ego_algorithm_description}

\begin{algorithm}[t]\setstretch{1.1}
\caption{\ours{} EgoUpdate Function}
\label{alg:rotate_ego_update}
\begin{algorithmic}[1]
\STATE {\bfseries Input:} Environment $\text{Env}$, ego policy parameters $\theta^{\text{ego}}$, teammate policy buffer $\text{B}_{\pi}$,
number of updates $N_{\text{updates}}$, PPO clipping parameter $\epsilon$, PPO update epochs $N_{\text{epochs}}$.
\STATE {\bfseries Output:} Updated ego policy $\theta^{\text{ego}}$.

\STATE $\sigma^{\text{ego}} \leftarrow \textbf{Init}(V)$
\algorithmiccomment{Initialize critic for ego policy}

\FOR{$t_{\text{update}}=1$ {\bfseries to} $N_{\text{updates}}$}
    \STATE $\theta^{-i} \sim \mathcal{U}(\text{B}_{\pi})$
    \algorithmiccomment{Sample teammate parameters uniformly} \label{alg:rotate-ego:sample_teammate}

    \STATE $D \leftarrow \textbf{Interact}(\pi_{\theta^{-i}}, \pi_{\theta^{\text{ego}}}, p_{0}^{\text{Env}})$
    \label{alg:rotate-ego:interact}

    \STATE $\theta^{\text{ego}}_{\text{old}}, \sigma^{\text{ego}}_{\text{old}} \leftarrow \theta^{\text{ego}}, \sigma^{\text{ego}}$

    \FOR{$k_{\text{update}}=1$ {\bfseries to} $N_{\text{epochs}}$}
        \STATE $\mathcal{L}_{\text{EGO}}(\theta^{\text{ego}}) \leftarrow \mathcal{L}_{\text{CLIP}}\left(
        \theta^{\text{ego}}, \theta^{\text{ego}}_{\text{old}}, \sigma^{\text{ego}}_{\text{old}}, D, \epsilon, \mathbf{1}, +1\right) + \mathcal{L}_{\text{ENT}}\left(\theta^{\text{ego}}, D\right)$
        \algorithmiccomment{Ego policy loss} \label{alg:rotate-ego:ppo-loss}

        \STATE $L_{V}(\sigma^{\text{ego}}) \leftarrow \mathcal{L}_{\text{VAL}}\left(
        \sigma^{\text{ego}}, \sigma^{\text{ego}}_{\text{old}}, D\right)$
        \algorithmiccomment{Ego critic loss} \label{alg:rotate-ego:critic-loss}

        \STATE $\theta^{\text{ego}} \leftarrow \textbf{GradDesc}\left(
        \theta^{\text{ego}}, \nabla_{\theta^{\text{ego}}} \mathcal{L}_{\text{EGO}}(\theta^{\text{ego}})\right)$
        \algorithmiccomment{Update policy}

        \STATE $\sigma^{\text{ego}} \leftarrow \textbf{GradDesc}\left(
        \sigma^{\text{ego}}, \nabla_{\sigma^{\text{ego}}} L_{V}(\sigma^{\text{ego}})\right)$
        \algorithmiccomment{Update critic}
    \ENDFOR
\ENDFOR

\STATE {\bfseries Return:} $\theta^{\text{ego}}$
\end{algorithmic}
\end{algorithm}

The ego agent policy's training process proceeds according to Algorithm~\ref{alg:rotate_ego_update}. 
\cref{alg:rotate-ego:sample_teammate} illustrates how \ours{} creates different teammate policies by uniformly sampling model parameters from the $\text{B}_{\pi}$ resulting from the teammate generation process. 
Using the experience collaborating with the sampled policies outlined in \cref{alg:rotate-ego:interact}, the ego agent's policy parameters are updated to maximize its returns via PPO in \cref{alg:rotate-ego:ppo-loss}. 
The ego policy loss $\mathcal{L}_{\text{EGO}}$ reuses the $\mathcal{L}_{\text{CLIP}}$ and $\mathcal{L}_{\text{ENT}}$ functions of App.~\ref{app:rotate_algorithm_description} with unit weights and a positive sign, since the ego agent performs standard return maximization from the environment's initial state distribution.
The single difference from the teammate generation setting is that the input dataset, $D$, stores the historical sequence of observed states and executed actions, $h$, rather than states.
Likewise, the critic loss $\mathcal{L}_{\text{VAL}}$ in \cref{alg:rotate-ego:critic-loss} is assumed to be computed on the observation-action history rather than states.
Like recent AHT learning algorithms~\citep{zintgraf2021deep, rahman_towards_2021, papoudakis2020liam}, $\pi^{\text{ego}}$ and $V^\text{ego}$ are conditioned on the ego agent's observation-action history to facilitate an adaptive $\pi^{\text{ego}}$ through an improved characterization of teammates' policies. 
The history-conditioned ego architecture and other practical implementation details are described in App.~\ref{app:implementation}.
Because the population buffer grows over the course of training, the ego agent faces a nonstationary training distribution. 
Experimentally, the ego agent learns effectively against this nonstationary distribution only with a lower entropy coefficient and learning rate than those used to train the teammate and best response agents (typically on the order of \num{1e-4} for the entropy coefficient and \num{1e-5} for the learning rate).
Finally, the ego agent update function returns the updated ego agent policy parameters, which are provided as part of the inputs for the next call to \ours{}'s teammate generation function.

\section{Baselines Overview}
\label{app:baselines}

The main paper compares \ours{} to six baselines: PAIRED, Minimax Return, FCP, BRDiv, CoMeDi, and COLE.
Each baseline is briefly described below, followed by a discussion of the computational complexity of teammate generation baselines compared to \ours{}, and a discussion of the relationship of Mixed Play (MP) with per-state and per-trajectory regret. 
Implementation details are described in App. \ref{app:implementation}.

\paragraph{PAIRED~\citep{dennis_emergent_2020}:} A UED algorithm in which a regret-maximizing ``adversary" agent proposes environment variations that an allied antagonist achieves high returns on, but a protagonist agent receives low returns on. The algorithm is directly applicable to AHT by defining a teammate generator for the role of the adversary, a best response agent to the generated teammate for the role of the antagonist, and an ego agent for the role of the protagonist. %
In contrast to \ours{}, our PAIRED extension jointly trains the ego agent, teammate, and BR. The teammate is optimized to maximize the per-trajectory regret of the ego agent. Following the original PAIRED, which does not maintain a population of environments, our extension does not use a population buffer, so the ego agent trains only against the most recently generated teammate.

\paragraph{Minimax Return~\citep{morimoto_robust_2005}:}
The Minimax Return baseline is based on the minimax return baseline of the PAIRED paper~\citep{dennis_emergent_2020}, with the original idea due to \citet{morimoto_robust_2005}.
It uses the same open-ended framework as \ours{}, but generates teammates that minimize the ego agent's \textit{return} rather than maximizing its regret. Note that the Minimax Bayes algorithm of \citet{villin2025minimaxapproachadhoc} is not our Minimax Return baseline. While conceptually related, it instead addresses how to \textit{curate} a pre-existing population of teammates to best train an ego learner.

\paragraph{Fictitious Co-Play~\citep{strouse_fcp_2022}:} 
A two-stage AHT algorithm in which a pool of teammates is generated by running IPPO~\citep{yu_surprising_2022} with varying seeds, and saving multiple checkpoints to the pool. The ego agent is an IPPO agent that is trained against the pool.

\paragraph{BRDiv~\citep{rahman2023generating}:} 
A two-stage AHT algorithm in which a population of ``confederate" and best-response agent pairs is generated, and an ego agent is trained against the confederates. BRDiv maintains a cross-play matrix containing the returns for all confederate and best-response pairs. The diagonal returns (self-play) are maximized, while the off-diagonal returns (cross-play) are minimized. BRDiv and LIPO~\citep{charakorn2023generating} share a similar objective, with three main differences: (1) if \texttt{xp\_weight} denotes the weight on the XP return, then BRDiv requires that the coefficient on the SP return is always $1 + 2*\texttt{xp\_weight}$, (2) LIPO introduces a secondary diversity metric based on mutual information, and (3) LIPO assumes that agents within a team (i.e., a confederate-BR pair) share parameters.

\paragraph{CoMeDi~\citep{sarkar2023diverse}:} 
CoMeDi is a two-stage AHT algorithm. In the first stage, a population of teammates is generated, and in the second stage, an ego agent is trained against the teammate population.
The teammate generation stage trains teammate policies one at a time, in which the $n$th teammate policy is trained to maximize its SP return, minimize its XP return with the previously generated teammate (i.e., from among teammates $1, \cdots, n-1$) that it best collaborates with, and maximize its ``mixed-play" (MP) return. The relationship between the regret objectives described in Section \ref{sec:method} and MP is further discussed in App. \ref{app:supp_disc:comedi_mp_discussion}.

\paragraph{COLE~\citep{li_cole_2023}:}
Cooperative Open-Ended Learning (COLE) is a population-based framework for AHT that generates teammates iteratively, similar to CoMeDi. 
At each iteration, a new policy is trained to maximize returns against a distribution over the current population of strategies (policies), and then added to the population.
The distribution is based on a cooperative incompatibility score, which measures how difficult each policy is to coordinate with.
The cooperative incompatibility of each policy is quantified using a Shapley value metric to determine the contribution of each policy to the overall incompatibility of the population. 
By training each policy against the least compatible policies in the population, COLE encourages the generation of teammates that are more compatible with a wider range of partners.
At the end of the teammate generation process, the teammate with the highest average cross-play return across the population is selected and returned as the final ego agent. 

\subsection{Computational Complexity of \ours{} versus Teammate Generation Baselines}
\label{app:computation_complexity_discussion}

The computational complexity of \ours{} is compared with that of the teammate generation baselines, in terms of the population size and the number of objective updates. 
In the following, $n$ denotes the population size, while $T$ indicates the number of updates needed to train an individual population member.
The precise meaning of $n$ and $T$ might vary with the algorithm, but is made clear in each description. 

\paragraph{FCP:}
Let $T$ denote the number of RL updates needed to train each IPPO team and let $n$ denote the number of teams trained by FCP. Then, the computational complexity of FCP is $O(nT)$.

\paragraph{BRDiv/LIPO:}
Both BRDiv~\citep{rahman2023generating} and LIPO~\citep{charakorn2023generating} require sampling trajectories from each pair of agents in the population, for each update. Thus, if the total number of updates is $T$ and the population size is $n$, then the algorithm's time complexity is $O(n^2T)$.
Due to the quadratic complexity in $n$, BRDiv and LIPO are typically run with smaller population sizes, with $n$ < 10 for all non-matrix game tasks in both original papers.

\paragraph{CoMeDi:}
Recall that CoMeDi trains population members one at a time, such that each agent is distinct from the previously discovered teammates in the population. This necessitates performing evaluation rollouts of the currently trained agent against all previously generated teammates at each RL update step. 
Let $T$ be the number of RL updates required to train the $i$th agent to convergence, and let $n$ denote the population size. Then CoMeDi's time complexity is $O(n^2T)$---making it scale quadratically in $n$, similar to BRDiv and LIPO.

\paragraph{COLE:}
COLE trains a new teammate policy to convergence at each iteration of open-ended learning. The cooperative incompatibility score is based on the cross-play returns of the current population, which requires evaluating each new teammate against all previously generated teammates. Thus, similar to CoMeDi, the time complexity of COLE is $O(n^2T)$, where $n$ is the population size and $T$ is the number of RL updates required to train each teammate to convergence.

\paragraph{\ours{}:}  
In \ours{}, a new teammate is trained to convergence for each iteration of open-ended learning. Thus, the number of open-ended learning iterations is equal to the population size $n$, where within each iteration, there are $O(T)$ RL updates performed. Therefore, the complexity of \ours{} is $O(nT)$, meaning that our method scales linearly in the population size $n$.

\section{Experimental Tasks}
\label{sec:app_exp_tasks}

Experiments in the main paper are conducted on JAX re-implementations of Level-Based Foraging (LBF)~\citep{albrecht_game-theoretic_2013, bonnet_jumanji_2023}, five tasks from the Overcooked suite (Cramped Room (CR), Asymmetric Advantages (AA), Counter Circuit (CC), Coordination Ring (CoR), and Forced Coordination (FC))~\citep{carroll_utility_2019,rutherford_JAXmarl_2024} and a sabotage matrix game. 
The JAX implementations of LBF and Overcooked are provided by Jumanji and JaxMARL, respectively, both released under the Apache-2.0 license. The human gameplay data used for the human proxy evaluation (App.~\ref{app:supp:human_proxy_eval}) is distributed in the MIT-licensed Overcooked-AI repository~\citep{carroll_utility_2019}.
The sabotage matrix game is already fully described in the main paper. The other tasks are described below. 

\paragraph{Level-Based Foraging (LBF)}
Originally introduced by \citet{albrecht_game-theoretic_2013}, Level-Based Foraging is a cooperative logistics problem in which $N$ players interact within a rectangular grid world to obtain $k$ foods. 
All players and foods have a positive integer \textit{level}, where groups of one to four players may only \textit{load} (collect) a food if the sum of player levels is at least the food's level. 
A food's level is configured so that it is always possible to load it.
\begin{wrapfigure}{r}{0.3\textwidth}
    \centering
    \includegraphics[width=\linewidth]{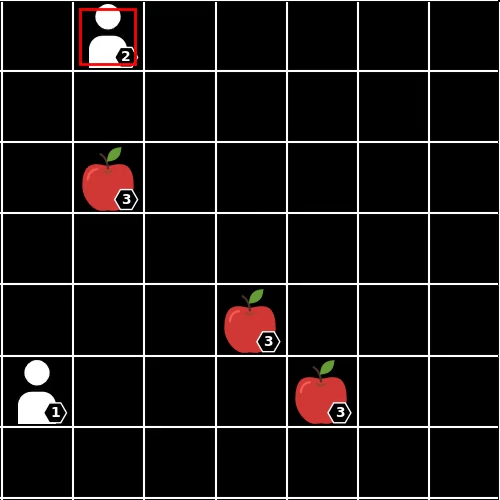}
    \caption{\small Level-based foraging environment. The apple icons denote food. The number on the icon indicates each player's and food's level. The AHT player is indicated by the red box.}
    \label{fig:lbf_env}
\end{wrapfigure}

We use the JAX re-implementation of LBF by \citet{bonnet_jumanji_2023}, which was based on the implementation by~\citet{christianos2020shared}.
The implementation permits the user to specify the number of players, number of foods, grid world size, level of observability, and whether to set the food level equal to the sum of player levels in order to force players to coordinate to load each food.

The experiments in this paper configured the LBF environment to a $7\times7$ grid, in which two players interact to collect three foods. Our LBF configuration is shown in Fig. \ref{fig:lbf_env}.  
Each player observes the full environment state, allowing each player to observe the locations of other agents and all foods and the number of time steps elapsed in the current episode. 
Each player has six discrete actions: up, down, left, right, no-op, and \textit{load}, the last of which is the special food collection action. 
A food may only be collected if the sum of player levels is at least the level of the food. 
Since this paper focuses on fully cooperative scenarios, we set the food level equal to the sum of the levels of both players, so all foods require cooperation in order to be collected. 
When a food is collected, both players receive an identical reward, which is normalized such that the maximum return in an episode is 0.5. Raw LBF returns are converted to the percent of food eaten, reported elsewhere in this paper, by multiplying by 200, so a return of 0.5 corresponds to 100\% of food eaten.
An episode terminates if an invalid action is taken, if players collide, or when 100 time steps have passed. Player and food locations are randomized for each episode.

\paragraph{Overcooked}

Introduced by~\citet{carroll_utility_2019}, the Overcooked suite is a set of two-player collaborative cooking tasks, based on the commercially successful Overcooked video game.
Designed to study human-AI collaboration, the original Overcooked suite consists of five simple environment \textit{layouts}, in which two agents collaborate within a grid world kitchen to cook and deliver onion soups.
While \citet{carroll_utility_2019} introduced Overcooked to study human-AI coordination, Overcooked has become popularized for AHT research as well~\citep{charakorn2023generating,sarkar2023diverse,erlebach2024raccoon}.

We use the JAX re-implementation of the Overcooked suite by~\citet{rutherford_JAXmarl_2024}, which is based on the original implementation by~\citet{carroll_utility_2019}.
Later versions of Overcooked include features such as multiple dish types, order lists, and alternative layouts, but this paper considers only the five original Overcooked layouts:
Cramped Room (CR), Asymmetric Advantages (AA), Counter Circuit (CC), Coordination Ring (CoR), and Forced Coordination (FC).

\begin{figure}[h]
    \centering
    \begin{subfigure}[t]{0.15\linewidth} %
        \includegraphics[trim= 30 15 30 15,clip,width=\linewidth]{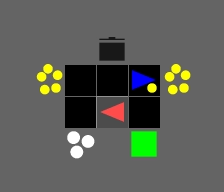}
        \captionsetup{format=hang, font=small}
        \caption{Cramped Room.}
        \label{fig:cr_env}
    \end{subfigure}
    \hfill
    \begin{subfigure}[t]{0.15\linewidth} %
        \centering
        \includegraphics[trim= 30 30 30 30,clip,width=\linewidth]{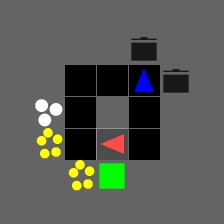}
        \captionsetup{format=hang, font=small}
        \caption{Coord.\newline Ring.}
        \label{fig:cor_env}
    \end{subfigure}
    \hfill
    \begin{subfigure}[t]{0.15\linewidth} %
        \centering
        \includegraphics[trim= 30 30 30 30,clip,width=\linewidth]{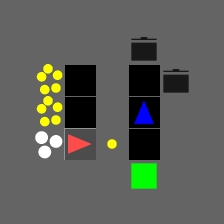}
        \captionsetup{format=hang, font=small}
        \caption{Forced Coord.}
        \label{fig:fc_env}
    \end{subfigure}
    \begin{subfigure}[t]{0.27\linewidth} %
        \centering
        \includegraphics[trim= 30 30 30 30,clip,width=\linewidth]{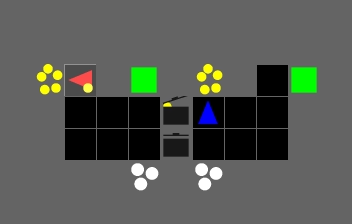}
        \captionsetup{format=hang, font=small}
        \caption{Asymmetric Advantages.\newline}
        \label{fig:aa_env}
    \end{subfigure}
    \hfill
    \begin{subfigure}[t]{0.24\linewidth} %
        \centering
        \includegraphics[trim= 30 30 30 30,clip,width=\linewidth]{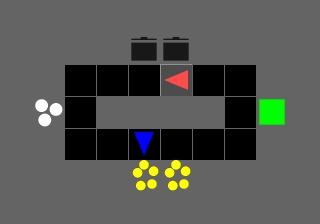}
        \captionsetup{format=hang, font=small}
        \caption{Counter Circuit.\newline}
        \label{fig:cc_env}
    \end{subfigure}
    \caption{\small The five classic Overcooked layouts. Each yellow circle is an onion, while white circles are plates. Grid spaces with multiple yellow (resp. white) circles are onion (resp. plate) piles, which agents must visit to pick up an onion (or plate). The green square is the delivery location, where finished dishes must be sent to receive a reward. Black squares denote free space, while adjacent gray spaces are empty counters. A black pot icon indicates pots, while agents are shown as red and blue pointers. The AHT agent is highlighted.}
    \label{fig:overcooked_env}
\end{figure}

The objective for all five tasks is to deliver as many onion soups as possible, where the only difference between the tasks is the environment layout, as shown in Fig. \ref{fig:overcooked_env}.
To deliver an onion soup, players must place three onions in a pot to cook, use a plate to pick up the cooked soup, and send the plated soup to the delivery location.
Each player observes the state and location of all environment features (counters, pots, delivery, onions, and plates), the position and orientation of both players, and an urgency indicator, which is 1 if there are 40 or fewer remaining time steps, and 0 otherwise.
Each player has six discrete actions, consisting of the four movement actions, interact, and no-op. 
The reward function awards both agents $+20$ upon successfully delivering a dish, which is the return reported in the experimental results. 
To improve sample efficiency, all algorithms are trained using a shaped reward function that provides each agent an additional reward of $0.1$ for picking up an onion, $0.5$ for placing an onion in the pot, $0.1$ for picking up a plate, and $1.0$ for picking up a soup from the pot with a plate. 
An episode terminates after 400 time steps. 
Player locations are randomized in each episode. In divided layouts such as AA and FC, we ensure that a player is spawned on each half of the layout.

\subsection{Optimal Behavior in Asymmetric Advantages}
\label{app:comment_on_aa}
Optimal behavior in AA still requires effective coordination due to layout asymmetry. 
In the ``left" kitchen, the delivery zone is adjacent to the pots while the onions are farther, while in the ``right" kitchen, the opposite is true. 
Thus, an optimal team consists of the ``left" agent delivering finished soup, and the ``right" agent placing onions in the pots---and indeed, we observe that teams of IPPO agents converge to this behavior.
In the main paper, we observed that Minimax Return performed surprisingly well on this task.
Visualizing the trained Minimax Return policies on AA reveals that the adversarial teammate trained by Minimax Return cannot drive the ego agent's return to zero, and does not prevent the ego agent from learning how to perform the task independently.

\section{Implementation Details}
\label{app:implementation}

As implementations of prior methods use PyTorch while this paper uses JAX, we re-implemented all methods in this paper, using PPO~\citep{Schulman2017ProximalPO} with Generalized Advantage Estimation (GAE)~\citep{Schulman2015HighDimensionalCC} as a base RL algorithm and Adam~\citep{kingma_adam_2017} as the default optimizer.
The code for this paper is released for reproducibility at \url{https://github.com/carolinewang01/rotate}.
A detailed discussion of \ours{}'s losses, update procedure, and pseudocode is provided in App. \ref{app:algorithm_pseudocode}.
This section discusses implementation details such as training time choices, agent architectures, and key hyperparameters for \ours{} and all baselines.

\subsection{Training Compute}
\label{app:training_compute}

For fair comparison, every method was trained with the same total environment-interaction budget on each task, counting both the teammate generation stage and ego agent training: 360M steps on LBF, 240M steps on CR and AA, and 440M steps on CoR, CC, and FC. This budget applies to all methods except for FCP, whose population consists of independently trained IPPO agents. Rather than scaling its population to consume the shared budget, we used a similar population size to the original paper (ours is ${\sim}110$ agents per task, while the original used 96 agents per task).
Population sizes for the remaining two-stage baselines were chosen as follows.
BRDiv was trained with a population size of 3-4 agents.
While we attempted training BRDiv with a larger population size, the algorithm was prone to discovering degenerate solutions in which only 2-3 agents in the population could discover solutions with high SP returns, and all other agents in the population would have zero returns.
CoMeDi was trained with a population size of 10 agents.
We attempted to train CoMeDi with a larger population size, but due to the algorithm's quadratic complexity in the population size, its runtime surpassed the available time budget.
Nevertheless, the population size of 10 forms a reasonable comparison to \ours{} because (1) the original paper used a population size of 8 for all Overcooked tasks, and (2) the configuration of CoMeDi in this paper runs for a similar wall-clock time as \ours{}.

\subsection{Compute infrastructure}

Experiments were performed on two servers, each with the following specifications:
\begin{itemize}[leftmargin=*]
    \item \textbf{CPUs}: two Intel(R) Xeon(R) Gold 6342 CPUs, each with 24 cores and two threads per core.
    \item \textbf{GPUs}: four NVIDIA A100 GPUs, each with 81920 MiB VRAM.
\end{itemize}

The experiments in this paper were implemented in JAX and parallelized across seeds.
On the servers above, each method took approximately 4-6 hours of wall-clock time to run.
\subsection{Agent Architectures}

For all methods considered in this paper, agents are implemented using neural networks and an actor-critic architecture, as is standard for PPO-based RL algorithms. 
All AHT methods implement policies without parameter sharing~\citep{christianos2020shared}, to enable greater behavioral diversity. 
Specifics for ego agents, teammates, and best response agents are described below.

As mentioned in the main paper, ego agents are history-conditioned. 
Thus, ego agents are implemented with the S5 actor-critic architecture, a recently introduced recurrent architecture shown to have stronger long-term memory than prior types of recurrent architectures. Another advantage of the S5 architecture over typical recurrent architectures (e.g. LSTMs) is that it is parallelizable during training, allowing significant speedups in JAX~\citep{lu_s5_2023}. 

On the other hand, teammates and best response agents are state-based. 
Best response agents are implemented with fully connected neural networks.
Teammates are also based on fully connected neural networks, but the precise architecture varies based on the algorithm. 
For methods in which the teammate only interacts with itself (FCP) or with the ego agent (Minimax Return), a standard actor-critic architecture is used. 
However, for open-ended learning methods that optimize regret (\ours{} and PAIRED), or for teammate generation methods that optimize adversarial diversity (CoMeDi and BRDiv), teammates must estimate returns when interacting with multiple agents. 
Thus, for these methods, the teammate architecture includes a critic for each type of interaction. 

For \ours{} and PAIRED, the teammate must estimate returns when interacting with the ego agent and its best response, and so it maintains a critic network for each partner type. 
For CoMeDi and BRDiv, given a population with $n$ agents, each teammate must estimate the return when interacting with the other $n-1$ agents in the population. As it would be impractical to maintain $n-1$ critics for each teammate, the teammate instead uses a critic that conditions on the agent ID of a candidate partner agent---in effect, implementing the $n-1$ critics via \textit{parameter sharing}~\citep{christianos2020shared}. 

\begin{table}[h]
\centering
\small
\caption{Hyperparameters for IPPO.}
\label{tab:ippo_hyperparams}
\setlength{\tabcolsep}{3pt}
\renewcommand{\arraystretch}{1.3} %
\captionsetup{skip=10pt} %
\begin{tabular}{ >{\raggedright\arraybackslash}p{1.9cm} *{6}{>{\raggedright\arraybackslash}p{1.75cm}} }
\hline
Task & LBF & CR & AA & CC & CoR & FC \\ \hline
Timesteps & \textbf{5e5}, 1e6 & 1e6 & 1e6 & 1e6, \textbf{3e6} & \textbf{3e6} & 1e6, \textbf{3e6}, 1e7 \\
Number envs & \textbf{8}, 16 & \textbf{8}, 16 & 8 & \textbf{8}, 16 & 8 & 8 \\
Epochs & 7, \textbf{15} & 15 & 15 & \textbf{15}, 30 & 15 & 15 \\
Minibatches & \textbf{4}, 8 & 4, 8, \textbf{16}, 32 & 16 & 16 & 16 & 16 \\
Clip-Eps & \textbf{0.03}, 0.05 & 0.03, 0.05, 0.10, 0.15, \textbf{0.2}, 0.3 & 0.2, \textbf{0.3} & \textbf{0.1}, 0.2 & \textbf{0.1}, 0.2, 0.3 & \textbf{0.1}, 0.2 \\
Ent-Coef & 5e-3, \textbf{0.01}, 0.03, 0.05 & 5e-3, \textbf{0.01}, 0.03, 0.05 & \textbf{0.01}, 0.02 & 0.01, 0.03, \textbf{0.05} & 0.001, 0.01, \textbf{0.05} & 0.01, \textbf{0.05} \\
LR & 1e-4 & 1e-4 & \textbf{1e-4}, 1e-3 & 1e-4, \textbf{1e-3} & 1e-4, 5e-4, \textbf{1e-3} & 1e-4, 5e-4, \textbf{1e-3} \\
Anneal LR & \textbf{true}, false & \textbf{true}, false & true & \textbf{true}, false & true & true \\
\hline
\end{tabular}
\end{table}

\begin{table}[htbp]
\centering
\small
\caption{Hyperparameters for the teammate generation stage of BRDiv.}
\label{tab:brdiv_hyperparams}
\setlength{\tabcolsep}{3pt}
\renewcommand{\arraystretch}{1.3} %
\captionsetup{skip=10pt} %
\begin{tabular}{ >{\raggedright\arraybackslash}p{2.0cm} >{\raggedright\arraybackslash}p{2.7cm} *{6}{>{\raggedright\arraybackslash}p{0.85cm}} }
\hline
 & Search Grid & LBF & CR & AA & CC & CoR & FC \\ \hline
Timesteps & --- & 3.3e8 & 2.1e8 & 2.1e8 & 3.8e8 & 3.8e8 & 3.8e8 \\
XP Coefficient & 0.001, 0.005, 0.01, 0.05, 0.1, 0.5, 1 & \textbf{0.1} & \textbf{0.5} & \textbf{0.5} & \textbf{0.01} & \textbf{0.005} & \textbf{0.01} \\
Population size & --- & \textbf{3}, 4, 5, 10 & 2, 3, \textbf{4}, 5 & \textbf{3}, 4 & \textbf{3}, 4 & \textbf{3}, 4 & \textbf{3}, 4 \\
Num Envs & --- & 64 & 64 & 64 & 128 & 128 & 128 \\
LR & 1e-5, 5e-5, 1e-4, 5e-4, 1e-3 & \textbf{5e-5} & \textbf{1e-5} & \textbf{1e-3} & \textbf{1e-5} & \textbf{1e-5} & \textbf{5e-4} \\
Ent-Coef & 0.005, 0.01, 0.03, 0.05, 0.1 & \textbf{0.005} & \textbf{0.05} & \textbf{0.05} & \textbf{0.01} & \textbf{0.05} & \textbf{0.01} \\
Clip-Eps & 0.05, 0.1, 0.2, 0.3, 0.5 & \textbf{0.2} & \textbf{0.3} & \textbf{0.05} & \textbf{0.01} & \textbf{0.1} & \textbf{0.05} \\
\hline
\end{tabular}
\end{table}

\begin{table}[htbp]
\centering
\small
\caption{Hyperparameters for \ours{}. Hyperparameters specific to the teammate training process are marked by "(T)".}
\label{tab:rotate_hyperparams}
\setlength{\tabcolsep}{3pt}
\renewcommand{\arraystretch}{1.3} %
\captionsetup{skip=10pt} %
\begin{tabular}{ >{\raggedright\arraybackslash}p{2.05cm} >{\raggedright\arraybackslash}p{2.3cm} *{6}{>{\raggedright\arraybackslash}p{1.3cm}} }
\hline
 & Search Grid & LBF & CR & AA & CC & CoR & FC \\ \hline
OEL Iterations & --- & 30 & 30 & 30 & 20 & 20 & 20 \\
Num Envs & --- & 16 & 16 & 16 & 16 & 16 & 16 \\
$\lambda_1$ & 0, 1, 2, 3, 4, 5, 6, 7, 8, 9, 10 & \textbf{4} & \textbf{4} & \textbf{9} & \textbf{2} & \textbf{10} & \textbf{3} \\
$\lambda_2$ & 0, 1, 2, 3, 4, 5, 6, 7, 8, 9, 10 & \textbf{4} & \textbf{4} & \textbf{0} & \textbf{4} & \textbf{10} & \textbf{4} \\
Minibatches & --- & \textbf{4}, 8 & 8 & 8 & 8 & 8 & 8 \\
Timesteps per Iter (Ego) & --- & 2e6 & 2e6 & 2e6 & 6e6 & 6e6 & 6e6 \\
Epochs (Ego) & --- & 5, \textbf{10}, 20 & \textbf{10}, 15 & 10 & 10 & 10 & \textbf{5}, 10 \\
Ent-Coef (Ego) & --- & \textbf{1e-4}, 1e-3, 0.01, 0.05 & 1e-4, \textbf{1e-3}, 1e-2 & \textbf{1e-3}, 0.01 & \textbf{1e-3}, 0.05 & \textbf{1e-3}, 0.05 & \textbf{1e-4}, 1e-3, 1e-2 \\
LR (Ego) & --- & \textbf{5e-5}, 1e-4, 1e-3 & 1e-5, 3e-5, \textbf{5e-5}, 1e-4 & 1e-5, 3e-5, \textbf{5e-5,} 1e-4 & 3e-5, \textbf{5e-5,} 1e-3 & 1e-5, \textbf{3e-5}, 5e-5, 1e-3 & 8e-6, \textbf{1e-5}, 3e-5, 5e-5, 1e-4 \\
Eps-Clip (Ego) & --- & 0.05, \textbf{0.1} & \textbf{0.1}, 0.2 & \textbf{0.1}, 0.3 & 0.1 & 0.1 & 0.1 \\
Anneal LR (Ego) & --- & true, \textbf{false} & true, \textbf{false} & true, \textbf{false} & true, \textbf{false} & true, \textbf{false} & true, \textbf{false} \\
Timesteps per Iter (T) & --- & 1e7 & 6e6 & 6e6 & 1.6e7 & 1.6e7 & 1.6e7 \\
Epochs (T) & --- & 20 & 20 & 20 & 20 & 20 & 20 \\
Ent-Coef (T) & 0.005, 0.01, 0.03, 0.05, 0.1 & \textbf{0.01} & \textbf{0.01} & \textbf{0.05} & \textbf{0.05} & \textbf{0.01} & \textbf{0.05} \\
LR (T) & 1e-5, 5e-5, 1e-4, 5e-4, 1e-3 & \textbf{1e-4} & \textbf{1e-4} & \textbf{5e-5} & \textbf{1e-3} & \textbf{1e-3} & \textbf{1e-3} \\
Clip-Eps (T) & 0.05, 0.1, 0.2, 0.3, 0.5 & \textbf{0.1} & \textbf{0.2} & \textbf{0.5} & \textbf{0.1} & \textbf{0.5} & \textbf{0.1} \\
Anneal LR (T) & --- & \textbf{true}, false & \textbf{true}, false & false & false & false & true, \textbf{false} \\ \hline
\end{tabular} \\
\end{table}

\begin{table}[htbp]
\centering
\small
\caption{Hyperparameters for PPO ego agent for all teammate generation methods.}
\label{tab:ppo_ego_tm_hyperparams}
\setlength{\tabcolsep}{3pt}
\renewcommand{\arraystretch}{1.3} %
\captionsetup{skip=10pt} %
\begin{tabular}{l l l l l l l }
\hline
 & LBF & CR & AA & CC & CoR & FC \\ \hline
Total Timesteps & 6e7 & 6e7 & 6e7 & 1.2e8 & 1.2e8 & 1.2e8 \\ %
Num Envs & 8 & 8 & 8 & 8 & 8 & 8 \\ %
LR & 5e-5 & 5e-5 & 5e-5 & 5e-5 & 3e-5 & 1e-5 \\ %
Epochs & 10 & 10 & 10 & 10 & 10 & 5 \\ %
Minibatches & 4 & 8 & 8 & 8 & 8 & 8 \\ %
Ent-Coef & 1e-4 & 1e-3 & 1e-3 & 1e-3 & 1e-3 & 1e-4 \\ %
Clip-Eps & 0.1 & 0.1 & 0.1 & 0.1 & 0.1 & 0.1 \\ %
Anneal LR & false & false & false & false & false & false \\ \hline
\end{tabular}
\end{table}

\begin{table}[htbp]
\centering
\small
\caption{Hyperparameters for teammate generation stage of FCP.}
\label{tab:fcp-tg-hyperparams}
\setlength{\tabcolsep}{3pt}
\renewcommand{\arraystretch}{1.3} %
\captionsetup{skip=10pt} %
\begin{tabular}{ >{\raggedright\arraybackslash}p{2.0cm} >{\raggedright\arraybackslash}p{2.55cm} *{6}{>{\raggedright\arraybackslash}p{0.85cm}} }
\hline
 & Search Grid & LBF & CR & AA & CC & CoR & FC \\ \hline
Timesteps Per Agent & --- & 2e6 & 2e6 & 2e6 & 4e6 & 4e6 & 4e6 \\
Num Seeds & --- & 23 & 23 & 23 & 22 & 22 & 22 \\ %
Num Checkpoints & --- & 5 & 5 & 5 & 5 & 5 & 5 \\ %
Num Envs & --- & 8 & 8 & 8 & 8 & 8 & 8 \\ %
LR & 1e-5, 5e-5, 1e-4, 5e-4, 1e-3 & \textbf{1e-5} & \textbf{1e-4} & \textbf{1e-5} & \textbf{1e-3} & \textbf{1e-3} & \textbf{5e-4} \\ %
Epochs & --- & 15 & 15 & 15 & 15 & 15 & 15 \\ %
Minibatches & --- & 4 & 16 & 16 & 16 & 16 & 16 \\ %
Ent-Coef & 0.005, 0.01, 0.03, 0.05, 0.1 & \textbf{0.005} & \textbf{0.01} & \textbf{0.1} & \textbf{0.03} & \textbf{0.01} & \textbf{0.005} \\ %
Eps-Clip & 0.05, 0.1, 0.2, 0.3, 0.5 & \textbf{0.5} & \textbf{0.2} & \textbf{0.2} & \textbf{0.05} & \textbf{0.3} & \textbf{0.2} \\ %
Anneal LR & --- & true & true & true & true & true & true \\ \hline
\end{tabular}
\end{table}

\begin{table}[htbp]
\centering
\small
\caption{Hyperparameters for the teammate generation stage of CoMeDi.}
\label{tab:comedi-tg-hyperparams}
\setlength{\tabcolsep}{3pt}
\renewcommand{\arraystretch}{1.3} %
\captionsetup{skip=10pt} %
\begin{tabular}{ >{\raggedright\arraybackslash}p{2.0cm} >{\raggedright\arraybackslash}p{3.1cm} *{6}{>{\raggedright\arraybackslash}p{0.85cm}} } \hline
 & Search Grid & LBF & CR & AA & CC & CoR & FC \\ \hline
Timesteps Per Iteration & --- & 3.3e7 & 2.1e7 & 2.1e7 & 3.8e7 & 3.8e7 & 3.8e7 \\
Population Size & --- & 10 & 10 & 10 & 10 & 10 & 10 \\
Num Envs & --- & 48 & 48 & 48 & 48 & 48 & 48 \\
LR & 1e-5, 5e-5, 1e-4, 5e-4, 1e-3 & \textbf{5e-4} & \textbf{5e-4} & \textbf{5e-5} & \textbf{1e-3} & \textbf{1e-3} & \textbf{5e-4} \\
Epochs & --- & 15 & 15 & 15 & 15 & 15 & 15 \\
Minibatches & --- & 8 & 8 & 8 & 8 & 8 & 8 \\
Ent-Coef & 0.005, 0.01, 0.03, 0.05, 0.1 & \textbf{0.03} & \textbf{0.005} & \textbf{0.1} & \textbf{0.03} & \textbf{0.005} & \textbf{0.01} \\
Eps-Clip & 0.05, 0.1, 0.2, 0.3, 0.5 & \textbf{0.3} & \textbf{0.3} & \textbf{0.1} & \textbf{0.5} & \textbf{0.3} & \textbf{0.5} \\
Anneal LR & --- & false & false & false & false & false & false \\
$\alpha$ & 0.1, 0.2, 0.3, 0.4, 0.5, 0.6, 0.7, 0.8, 0.9, 1.0 & \textbf{0.6} & \textbf{0.4} & \textbf{0.4} & \textbf{0.1} & \textbf{0.7} & \textbf{0.3} \\
$\beta$ & 0.1, 0.2, 0.3, 0.4, 0.5, 0.6, 0.7, 0.8, 0.9, 1.0 & \textbf{1.0} & \textbf{0.7} & \textbf{0.4} & \textbf{0.1} & \textbf{0.4} & \textbf{0.4} \\
\hline
\end{tabular}
\end{table}

\begin{table}[htbp]
\centering
\small
\caption{Hyperparameters for PAIRED.}
\label{tab:paired_hyperparams}
\setlength{\tabcolsep}{3pt}
\renewcommand{\arraystretch}{1.3} %
\captionsetup{skip=10pt} %
\begin{tabular}{ >{\raggedright\arraybackslash}p{2.0cm} >{\raggedright\arraybackslash}p{2.55cm} *{6}{>{\raggedright\arraybackslash}p{0.85cm}} } \hline
 & Search Grid & LBF & CR & AA & CC & CoR & FC \\ \hline
Timesteps & --- & 3.6e8 & 2.4e8 & 2.4e8 & 4.4e8 & 4.4e8 & 4.4e8 \\
Num Seeds & --- & 3 & 3 & 3 & 3 & 3 & 3 \\
Num Checkpoints & --- & 10 & 10 & 10 & 10 & 10 & 10 \\
Num Envs & --- & 16 & 16 & 16 & 16 & 16 & 16 \\
LR & 1e-5, 5e-5, 1e-4, 5e-4, 1e-3 & \textbf{1e-4} & \textbf{5e-4} & \textbf{1e-5} & \textbf{5e-5} & \textbf{1e-4} & \textbf{5e-5} \\
Epochs & --- & 15 & 15 & 15 & 15 & 15 & 15 \\
Minibatches & --- & 4 & 8 & 8 & 8 & 8 & 8 \\
Ent-Coef & 0.005, 0.01, 0.03, 0.05, 0.1 & \textbf{0.03} & \textbf{0.005} & \textbf{0.05} & \textbf{0.01} & \textbf{0.005} & \textbf{0.1} \\
Eps-Clip & 0.05, 0.1, 0.2, 0.3, 0.5 & \textbf{0.2} & \textbf{0.05} & \textbf{0.1} & \textbf{0.05} & \textbf{0.05} & \textbf{0.3} \\
Anneal LR & --- & false & false & false & false & false & false \\
\hline
\end{tabular}
\end{table}

\begin{table}[htbp]
\centering
\small
\caption{Hyperparameters for Minimax Return. Hyperparameters specific to the teammate training process are marked by "(T)".}
\label{tab:minimax_hyperparams}
\setlength{\tabcolsep}{3pt}
\renewcommand{\arraystretch}{1.3} %
\captionsetup{skip=10pt} %
\begin{tabular}{ >{\raggedright\arraybackslash}p{2.1cm} >{\raggedright\arraybackslash}p{2.55cm} *{6}{>{\raggedright\arraybackslash}p{0.85cm}} } \hline
 & Search Grid & LBF & CR & AA & CC & CoR & FC \\ \hline
OEL Iterations & --- & 30 & 30 & 30 & 20 & 20 & 20 \\
Num Envs & --- & 16 & 16 & 16 & 16 & 16 & 16 \\
Timesteps Per Iter (Ego) & --- & 2e6 & 2e6 & 2e6 & 6e6 & 6e6 & 6e6 \\
Timesteps Per Iter (T) & --- & 1e7 & 6e6 & 6e6 & 1.6e7 & 1.6e7 & 1.6e7 \\
LR & 1e-5, 5e-5, 1e-4, 5e-4, 1e-3 & \textbf{1e-3} & \textbf{1e-4} & \textbf{1e-5} & \textbf{1e-5} & \textbf{1e-4} & \textbf{5e-5} \\
Epochs & --- & 15 & 15 & 15 & 15 & 15 & 15 \\
Minibatches & --- & 4 & 8 & 8 & 8 & 8 & 8 \\
Ent-Coef & 0.005, 0.01, 0.03, 0.05, 0.1 & \textbf{0.03} & \textbf{0.1} & \textbf{0.05} & \textbf{0.005} & \textbf{0.005} & \textbf{0.01} \\
Eps-Clip & 0.05, 0.1, 0.2, 0.3, 0.5 & \textbf{0.03} & \textbf{0.1} & \textbf{0.2} & \textbf{0.05} & \textbf{0.1} & \textbf{0.05} \\
Anneal LR & --- & false & true & false & false & false & false \\ \hline
\end{tabular}
\end{table}

\begin{table}[htbp]
\centering
\small
\caption{Hyperparameters for COLE.}
\label{tab:cole_hyperparams}
\setlength{\tabcolsep}{3pt}
\renewcommand{\arraystretch}{1.3} %
\captionsetup{skip=10pt} %
\begin{tabular}{ >{\raggedright\arraybackslash}p{2.0cm} >{\raggedright\arraybackslash}p{3.2cm} *{6}{>{\raggedright\arraybackslash}p{1.1cm}} }
\hline
 & Search Grid & LBF & CR & AA & CC & CoR & FC \\ \hline
Timesteps Per Iteration & --- & 1.8e7 & 1.85e7 & 1.85e7 & 3.38e7 & 3.38e7 & 3.38e7 \\
Population Size & 12, 15, 18, 20 (LBF); 13, 15, 17 (CR, CoR) & \textbf{20} & \textbf{13} & 13 & 13 & \textbf{13} & 13 \\
Num Envs & --- & 64 & 64 & 64 & 64 & 64 & 64 \\
LR & 1e-5, 5e-5, 1e-4, 5e-4, 1e-3 & \textbf{1e-3} & \textbf{1e-3} & 1e-3 & 1e-3 & \textbf{1e-3} & 1e-3 \\
Epochs & --- & 15 & 15 & 15 & 15 & 15 & 15 \\
Minibatches & --- & 8 & 8 & 8 & 8 & 8 & 8 \\
Ent-Coef & 0.1, 0.05, 0.03, 0.01, 0.005 & \textbf{0.05} & \textbf{0.05} & 0.01 & 0.01 & \textbf{0.01} & 0.01 \\
Eps-Clip & --- & 0.05 & 0.05 & 0.3 & 0.01 & 0.1 & 0.05 \\
Anneal LR & --- & false & false & false & false & false & false \\
$\alpha$ & 0.1, 0.5, 1, 1.5 & \textbf{1.5} & \textbf{0.1} & 0.1 & 0.1 & \textbf{0.1} & 0.1 \\
$c_{\text{UCT}}$ & 0.01, 0.1, 0.2, 0.3, 0.4, 0.5 & \textbf{0.4} & \textbf{0.1} & 0.3 & 0.3 & \textbf{0.3} & 0.3 \\
\hline
\end{tabular}
\end{table}

\subsection{Hyperparameters}
\label{app:hyperparams}

This section presents the hyperparameters for \ours{} (\cref{tab:rotate_hyperparams}), baseline methods (\cref{tab:brdiv_hyperparams,tab:ppo_ego_tm_hyperparams,tab:fcp-tg-hyperparams,tab:comedi-tg-hyperparams,tab:paired_hyperparams,tab:minimax_hyperparams,tab:cole_hyperparams}), and IPPO (\cref{tab:ippo_hyperparams}), which served as the base algorithm for all methods and was used to train evaluation teammates.
For all methods, hyperparameters were first tuned manually, and then tuned by automated hyperparameter search for the most important hyperparameters. The procedure is described below.

 Manual hyperparameter tuning began with the base IPPO hyperparameters for each task (\cref{tab:ippo_hyperparams}), typically varying one parameter over a range while holding the others fixed, and varying parameters jointly only when varying one at a time did not yield the desired result. These manually-tuned IPPO settings served as the starting point for the IPPO training hyperparameters used by all algorithms. Next, hyperparameters for the shared ego agent training portion of each teammate generation algorithm (\cref{tab:ppo_ego_tm_hyperparams}) were tuned on a \textit{per-task} basis, and held fixed for fair comparison. Finally, the hyperparameters were tuned for each algorithm on a per-task basis, focusing on the teammate generation portion. 

The automated hyperparameter search consisted of random search over predefined hyperparameter ranges. For each algorithm-task combination, 15 configurations were sampled uniformly at random, where each was evaluated with 2 seeds at a reduced training budget. The final hyperparameter configuration was selected from the searched configurations and the configuration resulting from manual tuning.
The hyperparameters tuned by the random search are the learning rate, entropy coefficient, and PPO clip epsilon common to all methods, together with each method's own diversity or regret coefficients ($\lambda_1, \lambda_2$ for \ours{}, $\alpha, \beta$ for CoMeDi, and the XP-return coefficient for BRDiv). 

For each hyperparameter, the searched values are listed in the tables, and the selected values are bolded. The hyperparameters for the two-stage teammate generation methods are presented in separate tables, where those corresponding to the shared ego agent training stage are presented in \cref{tab:ppo_ego_tm_hyperparams}. 
All experiments in the paper were performed with a discount factor of $\gamma=0.99$ and $\lambda^{\text{GAE}}=0.95$.

\section{Evaluation Teammates}
\label{app:pieval_details}

\begin{figure}[t]
    \centering
    \includegraphics[width=\linewidth]{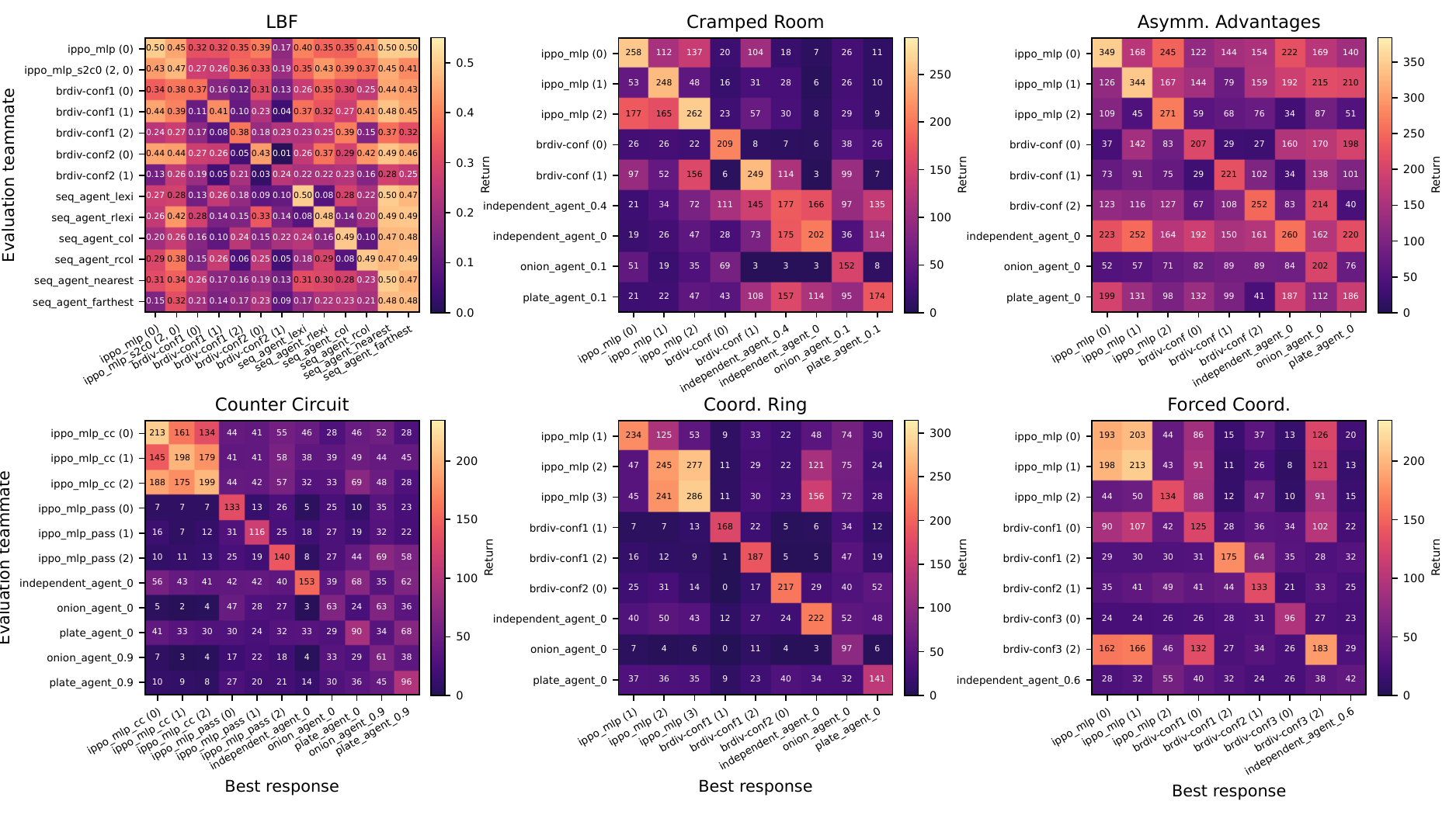}
    \caption{\small Cross-play matrices among the evaluation teammates and their best responses, per task, in unnormalized return units. Each diagonal entry is the raw empirical return of the teammate paired with its own RL-trained best response; the diagonal/off-diagonal gap evidences the diversity of the evaluation set. The per-teammate upper bounds used for return normalization are reported separately in \cref{tab:normalization_bounds}.}
    \label{fig:heldout_xp_matrices}
\end{figure}

\begin{table}[htbp]
\centering
\caption{Summary of evaluation teammate types.}
\label{tab:pieval_types}
\renewcommand{\arraystretch}{1.3}
\captionsetup{skip=10pt}
\begin{tabular}{ l c p{7.5cm} } \hline
Type & Tasks & Description \\ \hline
ippo\_mlp $(k)$ & LBF and Overcooked & Teammate trained by IPPO in self-play to maximize return; $k$ indexes the training seed. On CC, ippo\_mlp\_cc agents navigate counterclockwise around the counter. \\
ippo\_mlp $(k, c)$ & LBF & Intermediate checkpoint of a teammate trained by IPPO, where $k$ indexes the training seed and $c$ the checkpoint. \\
ippo\_mlp\_pass $(k)$ & Overcooked & Teammate trained by IPPO with reward shaping to pass onions across the counter (CC); $k$ as above. \\
brdiv-conf$j$ $(k)$ & LBF and Overcooked & Confederate teammate trained by BRDiv, where $j$ indexes the BRDiv training run and $k$ the member of its confederate population. \\
independent\_agent\_$p$ & Overcooked & Programmed agent that cooks and delivers soups on its own. The parameter $p$ is the probability that the agent places a held item on a nearby counter. \\
onion\_agent\_$p$ & Overcooked & Programmed agent that places onions in non-full pots; $p$ as above. \\
plate\_agent\_$p$ & Overcooked & Programmed agent that plates finished soups and delivers them; $p$ as above. \\
seq\_agent\_$\ast$ & LBF & Planning agents that collect food in a fixed order: lexicographic (lexi), reverse lexicographic (rlexi), column-major (col), reverse column-major (rcol), or nearest/farthest-first by Manhattan distance from the agent's initial position (nearest, farthest). \\
\hline
\end{tabular}

\end{table}

Evaluation teammates were constructed using three strategies: training IPPO teammates in self-play using varied seeds and reward shaping, training teammates with BRDiv, and manually programming heuristic agents.
The evaluation teammates trained using IPPO and BRDiv were trained using different seeds than those used for training \ours{} and baseline methods.
The teammate construction procedure results in distinct teammate archetypes, which are summarized in \cref{tab:pieval_types} and detailed below. The precise set of evaluation teammates used for each task may be read from the axes of the XP matrices shown in \cref{fig:heldout_xp_matrices}. 

Generally, IPPO agents execute straightforward, return-maximizing strategies. 
On the other hand, since BRDiv agents are trained to maximize self-play returns with their best response partner and to reduce cross-play returns with all other best response policies in the population, the generated teammates display more adversarial behavior compared to IPPO and heuristics. 
Coefficients on the SP and XP returns were carefully tuned to ensure that the behavior was not too adversarial, which we operationalized as teammates whose SP returns were high but whose XP returns were near zero. 
Finally, the manually programmed heuristic agents have a large range of skills and levels of determinism. 
The LBF heuristics are planning-based agents that deterministically attempt to collect the apples in a specific order.
Given a best response partner, the LBF heuristics can achieve the optimal task return in LBF.
The Overcooked heuristics execute pre-programmed roles that are agnostic to the layout, along with some basic collision-avoidance logic.
The ``onion" heuristic collects onions and places them in non-full pots.
The ``plate" heuristic plates soups that are ready, and delivers them.
The ``independent" heuristic attempts to fulfill both roles by itself.
All three heuristic types have a user-specified parameter that defines the probability that the agent places whatever it is holding on a nearby counter. 
The feature serves two purposes: first, it creates a larger space of behaviors, and second, it allows the heuristics to work for the FC task, where the agent in the left half of the kitchen must pass onions and plates to the right, while the agent in the right half must pick up resources from the dividing counter, cook soup, and deliver.

To quantify the diversity of $\Pieval$, we compute the cross-play (XP) matrix among evaluation teammates for each task, visualized in \cref{fig:heldout_xp_matrices}. Averaging each matrix, the mean diagonal return (where each teammate is paired with its own best response) substantially exceeds the mean off-diagonal return (where teammates are paired with another teammate's best response). The large gap between diagonal and off-diagonal returns provides evidence that the evaluation teammates employ diverse conventions, so the normalized evaluation return does capture robustness to conventions. Candidate teammates whose diagonal and off-diagonal returns were similar added no diversity and were pruned, which is why the per-source mix of evaluation teammates is uneven across tasks.

\subsection{Evaluation Return Normalization Procedure}
\label{app:lower_upper_return_bounds}

Normalizing evaluation returns is necessary in AHT because AHT asks whether the ego agent is a good partner \textit{for a given teammate}, relative to what is achievable with that teammate.
Because evaluation teammates vary in skill and therefore have different maximum team returns, unnormalized returns conflate ego agent quality with teammate skill.
Normalization is also necessary to report returns aggregated across evaluation teammates, and is 
in line with best-practice recommendations for both RL~\citep{agarwal_deep_2021} and AHT~\citep{wang_zsc-eval_2024}.

Concretely, each method is evaluated by the min-max normalized team return with each evaluation teammate.
The lower return bound is set to zero since a poor teammate could always cause a zero return in all tasks considered.
Ideally, the upper return bound for each evaluation teammate would be the return achieved with its theoretically optimal best response partner, but this bound is computationally intractable to obtain.
Instead, initial estimates are obtained by training BR policies via RL, and each estimate is then revised upward whenever \textit{any} partner policy exceeds it, including evaluated methods.
This correction procedure yields tighter bounds than using the RL-trained best responses alone or the compared methods alone.
The resulting per-teammate upper bounds are reported in \cref{tab:normalization_bounds}.

\begin{table}[htbp]
\centering
\small
\begin{tabular}[t]{lr}
\toprule
\multicolumn{2}{c}{\textbf{LBF}} \\
Teammate & Bound \\
\midrule
ippo\_mlp (0) & 100.0 \\
ippo\_mlp\_s2c0 (2, 0) & 96.4 \\
brdiv-conf1 (0) & 97.4 \\
brdiv-conf1 (1) & 100.0 \\
brdiv-conf1 (2) & 89.6 \\
brdiv-conf2 (0) & 100.0 \\
brdiv-conf2 (1) & 62.5 \\
seq\_agent\_lexi & 100.0 \\
seq\_agent\_rlexi & 100.0 \\
seq\_agent\_col & 100.0 \\
seq\_agent\_rcol & 100.0 \\
seq\_agent\_nearest & 100.0 \\
seq\_agent\_farthest & 100.0 \\
\bottomrule
\end{tabular}
\hfill
\begin{tabular}[t]{lr}
\toprule
\multicolumn{2}{c}{\textbf{Cramped Room}} \\
Teammate & Bound \\
\midrule
ippo\_mlp (0) & 258.4 \\
ippo\_mlp (1) & 253.8 \\
ippo\_mlp (2) & 262.2 \\
brdiv-conf (0) & 214.1 \\
brdiv-conf (1) & 248.8 \\
independent\_agent\_0.4 & 197.2 \\
independent\_agent\_0 & 202.2 \\
onion\_agent\_0.1 & 151.9 \\
plate\_agent\_0.1 & 191.2 \\
\bottomrule
\end{tabular}
\hfill
\begin{tabular}[t]{lr}
\toprule
\multicolumn{2}{c}{\textbf{Asymm. Advantages}} \\
Teammate & Bound \\
\midrule
ippo\_mlp (0) & 382.5 \\
ippo\_mlp (1) & 369.4 \\
ippo\_mlp (2) & 299.7 \\
brdiv-conf (0) & 283.4 \\
brdiv-conf (1) & 345.3 \\
brdiv-conf (2) & 333.1 \\
independent\_agent\_0 & 308.1 \\
onion\_agent\_0 & 299.4 \\
plate\_agent\_0 & 284.7 \\
\bottomrule
\end{tabular}
\\[1.5em]
\begin{tabular}[t]{lr}
\toprule
\multicolumn{2}{c}{\textbf{Counter Circuit}} \\
Teammate & Bound \\
\midrule
ippo\_mlp\_cc (0) & 213.1 \\
ippo\_mlp\_cc (1) & 200.9 \\
ippo\_mlp\_cc (2) & 199.4 \\
ippo\_mlp\_pass (0) & 136.6 \\
ippo\_mlp\_pass (1) & 115.9 \\
ippo\_mlp\_pass (2) & 170.0 \\
independent\_agent\_0 & 153.4 \\
onion\_agent\_0 & 95.0 \\
plate\_agent\_0 & 90.0 \\
onion\_agent\_0.9 & 91.9 \\
plate\_agent\_0.9 & 105.6 \\
\bottomrule
\end{tabular}
\hfill
\begin{tabular}[t]{lr}
\toprule
\multicolumn{2}{c}{\textbf{Coord. Ring}} \\
Teammate & Bound \\
\midrule
ippo\_mlp (1) & 249.7 \\
ippo\_mlp (2) & 246.6 \\
ippo\_mlp (3) & 285.9 \\
brdiv-conf1 (1) & 168.4 \\
brdiv-conf1 (2) & 187.2 \\
brdiv-conf2 (0) & 216.9 \\
independent\_agent\_0 & 221.6 \\
onion\_agent\_0 & 96.9 \\
plate\_agent\_0 & 141.2 \\
\bottomrule
\end{tabular}
\hfill
\begin{tabular}[t]{lr}
\toprule
\multicolumn{2}{c}{\textbf{Forced Coord.}} \\
Teammate & Bound \\
\midrule
ippo\_mlp (0) & 220.0 \\
ippo\_mlp (1) & 214.4 \\
ippo\_mlp (2) & 225.6 \\
brdiv-conf1 (0) & 131.6 \\
brdiv-conf1 (2) & 184.7 \\
brdiv-conf2 (1) & 143.8 \\
brdiv-conf3 (0) & 95.6 \\
brdiv-conf3 (2) & 183.1 \\
independent\_agent\_0.6 & 81.2 \\
\bottomrule
\end{tabular}
\caption{Per-teammate upper bounds used for min-max normalization of evaluation returns (the lower bound is always 0). Each bound is obtained from the RL-trained best-response return estimate, revised upward whenever any evaluated policy exceeded it. Units are percent of food eaten for LBF and the unshaped base return for the Overcooked tasks.}
\label{tab:normalization_bounds}
\end{table}

As described in Section \ref{sec:exps}, our normalized return metric is a member of the class of BRProx metrics recommended by~\citet{wang_zsc-eval_2024}, with the \textit{mean} as the main aggregation metric. The main comparison between \ours{} and baselines is also reported with IQM and median as aggregation functions (Fig. \ref{fig:per_task_interval_estimates}).

\end{document}